\documentclass{article}

\PassOptionsToPackage{numbers, compress}{natbib}

    \usepackage[preprint]{neurips_2025}

\usepackage[utf8]{inputenc} 
\usepackage[T1]{fontenc}    
\usepackage{bbding}

\usepackage{adjustbox}
\usepackage{nicefrac}       
\usepackage{microtype}      
\usepackage{bm}
\usepackage{amssymb}
\usepackage{arydshln}
\usepackage{makecell}

\usepackage{amsmath,amsfonts}
\usepackage{algorithmic}
\usepackage{algorithm}
\usepackage{array}
\usepackage[caption=false,font=normalsize,labelfont=sf,textfont=sf]{subfig}
\usepackage{textcomp}
\usepackage{stfloats}
\usepackage{url}
\usepackage{verbatim}
\usepackage{graphicx}
\usepackage{booktabs}
\usepackage{xcolor, colortbl}
\usepackage{tabularx}
\usepackage{multirow}
\usepackage{enumitem}
\usepackage{bbm}
\usepackage{pifont}
\usepackage{capt-of}
\usepackage{pifont}
\usepackage{arydshln}
\usepackage[pagebackref=false,breaklinks=true,colorlinks, bookmarks=false]{hyperref}
\usepackage{wrapfig}

\def\ie{\emph{i.e.}}

\newcommand{\figref}[1]{Fig.~\ref{#1}}
\newcommand{\Figref}[1]{Figure~\ref{#1}}
\newcommand{\tabref}[1]{Tab.~\ref{#1}}
\newcommand{\Tabref}[1]{Table~\ref{#1}}
\newcommand{\eqnref}[1]{Eqn.~\ref{#1}}
\newcommand{\secref}[1]{Sec.~\ref{#1}}

\newcommand*\samethanks[1][\value{footnote}]{\footnotemark[#1]}

\definecolor{green}{rgb}{0, 0.8, 0.2}

\title{Plenodium: UnderWater 3D Scene Reconstruction with Plenoptic Medium Representation}

\author{%
Changguang Wu\quad
Jiangxin Dong\thanks{Jiangxin Dong and Jinhui Tang are corresponding authors.}\quad
Chengjian Li\quad
Jinhui Tang\samethanks\\
Nanjing University of Science and Technology\\
\tt{\{changguangwu, jxdong, lichengjian, jinhuitang\}@njust.edu.cn}
}

\begin{document}

\maketitle
\begin{abstract}

We present \emph{Plenodium} (\emph{plenoptic medium}), an effective and efficient 3D representation framework capable of jointly modeling both objects and participating media.
In contrast to existing medium representations that rely solely on view-dependent modeling, our novel plenoptic medium representation incorporates both directional and positional information through spherical harmonics encoding, enabling highly accurate underwater scene reconstruction.
To address the initialization challenge in degraded underwater environments, we propose the pseudo-depth Gaussian complementation to augment COLMAP-derived point clouds with robust depth priors.
In addition, a depth ranking regularized loss is developed to optimize the geometry of the scene and improve the ordinal consistency of the depth maps.
Extensive experiments on real-world underwater datasets demonstrate that our method achieves significant improvements in 3D reconstruction.
Furthermore, we conduct a simulated dataset with ground truth and the controllable scattering medium to demonstrate the restoration capability of our method in underwater scenarios. Our code and dataset are available at: \url{https://plenodium.github.io/}.
\end{abstract}

\begin{wrapfigure}{r}{0.41\textwidth}  
    \centering
    \footnotesize
    \vspace{-4.5mm}
    \includegraphics[width=1\linewidth]{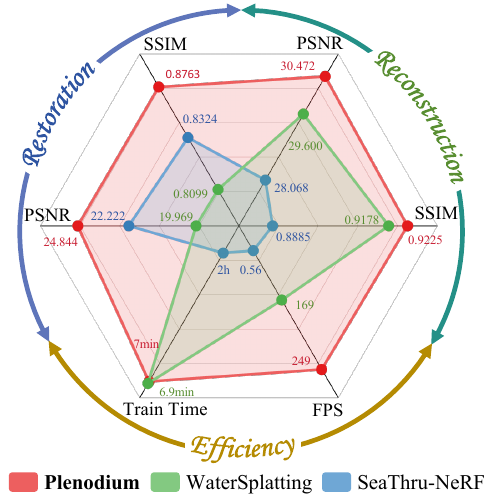}
    \vspace{-6mm}
    \caption{
    Comparison of Plenodium, WaterSplatting~\cite{watersplatting}, and SeaThru-NeRF~\cite{Seathru-nerf} on reconstruction and restoration performance (PSRN, SSIM), as well as efficiency (FPS, training time). 
    }
    \vspace{-8mm}
    \label{fig: comp}
  \end{wrapfigure}


\section{Introduction}
\label{sec: intro}


Underwater imaging plays a critical role in diverse applications, including underwater construction, marine sciences, etc.
However, its efficacy is significantly hindered by the complex optical properties of the aquatic environment.
These properties lead to wavelength- and distance-dependent attenuation and scattering of light, resulting in degraded image quality characterized by the color cast, diminished contrast, and loss of detail.
Given the expanding scientific and industrial focus on oceanic exploration, the reconstruction of scattering-affected underwater scenes becomes increasingly important.



Pioneering works~\cite{mipnerf,zipnerf,mini-splatting,taming3dgs,instantnpg} based on Neural Radiance Fields (NeRF)~\cite{NeRF} and 3D Gaussian Splatting (3DGS)~\cite{3dgs} have significant contributions to 3D reconstruction. These methods achieve effective surface modeling by constraining their representations to scene surfaces, assuming a vacuum-like medium between the observer and the objects. However, such approaches neglect the influence of light scattering in participating media, limiting their applicability in real underwater environments.

To achieve underwater 3D scene reconstruction, SeaThru-NeRF~\cite{Seathru-nerf} extends the NeRF framework by introducing an additional medium volume field, which is parameterized by an MLP and characterizes the color and density attributes of the medium, enabling accurate modeling of light-medium interactions. WaterSplatting~\cite{watersplatting} leverages 3DGS as an alternative geometric representation to replace the computational NeRF in SeaThru-NeRF~\cite{Seathru-nerf}, while preserving the core medium representation capability, achieving more efficient rendering. However, there are three significant limitations affecting its accuracy, efficiency, and robustness:
1) Existing methods estimate medium properties based exclusively on viewing directions, failing to account for the spatial relationship between camera positions and scattering effects in heterogeneous media;
2) Despite its advantages, WaterSplatting still employs an implicit MLP for medium representation, which introduces substantial computational costs limiting its efficiency;
3) The 3DGS-based approaches rely on COLMAP~\cite{schoenberger2016sfm,schoenberger2016mvs} for initializing Gaussian primitives, but underwater image degradation severely impairs COLMAP's feature extraction and matching, compromising the reliability of the initialization process.

In this paper, we present an effective and efficient method for underwater 3D reconstruction.
Different from existing methods that rely on the medium representation with limited directional information, we are the first to take into account the positional information and develop a plenoptic medium representation.
Notably, the proposed plenoptic medium representation is modeled by explicit Spherical Harmonics (SH), rather than implicit MLPs.
Specifically, we positionally encode the SH coefficients via a trilinear interpolation mechanism to capture accurate scattering effects in arbitrary positions, while achieving faster rendering than MLP-based methods.
Then to address the limitations of COLMAP in degraded underwater scenes, we propose a pseudo-depth Gaussian complementation method that enriches the sparse point clouds with pseudo-depth estimated from the Depth Anything Model~\cite{Depthanything,Depthanythingv2}, improving the robustness of the initialization for 3DGS. %
Furthermore, we introduce a depth ranking regularized loss to optimize the geometry of the scene, enhancing the ordinal stability of the depth maps.
Taken together, the proposed approach can effectively improve the reconstruction quality while speeding up rendering.
In addition, we created a simulated dataset for validating the restoration performance of our approach across various types of media and different degradation intensities, as well as analyzing degradation impacts on 3D reconstruction.

The contributions can be summarized as follows:
1) We propose \emph{Plenodium}, which introduces a novel plenoptic medium representation that characterizes both the directional and positional information and then incorporates it with 3DGS for effective underwater 3D reconstruction. 
2) To improve the robustness of 3DGS-based reconstruction in underwater scenarios, we introduce a pseudo-depth Gaussian complementation to enrich COLMAP-initialized Gaussian primitives and a depth ranking regularized loss to enhance the geometric consistency.
3) We construct a simulated dataset with ground truth (GT) and controllable scattering medium, which enables systematic evaluation of image restoration performance across degradation levels.
4) Extensive experiments demonstrate the effectiveness and efficiency of our approach. As shown in \figref{fig: comp},
Plenodium outperforms prior methods, increasing the PSNR by at least 0.872dB and speeding up the rendering efficiency by 47\% in real-world reconstruction scenarios.

\section{Related Work}
%
\noindent\textbf{3D Gaussian splatting}.
3DGS~\cite{3dgs} constructs a 3D scene representation with a set of 3D Gaussians, where the $i$-th Gaussian is defined by a center position $\mu_i \in \mathbb{R}^3$, a 3D covariance matrix $\Sigma_i \in \mathbb{R}^{3\times3}$, an opacity $\sigma_i \in \mathbb{R}$, and color features $A_i$.
Specifically, the rendered color $\hat C$ is computed by a blending process that combines the color contributions $\{\hat C_i\}_{i=1}^N$ from $N$ individual Gaussians:
\begin{equation}
    \hat C =\sum_{i=1}^{N}\hat C_i =\sum_{i=1}^{N} c_i \alpha_i T_i,\text{     where } T_i = \prod_{j=1}^{i-1}(1-\alpha_j),
    \label{eq: blend}
\end{equation}
where Gaussian color $c_i=\text{SH}( d, A_i)$ is derived from spherical harmonics~\cite{Plenoxels} with ray direction $d$ and its color features $A_i$, and $\alpha_i$ is computed by multiplying its $\sigma_i$ and its projected 2D Gaussian.

3DGS addresses limitations of the reconstruction efficiency in 3D scene modeling by leveraging explicit 3D Gaussian primitives for real-time rendering~\cite{mini-splatting,taming3dgs} and minute-level training~\cite{distwar,gsplat} on consumer GPUs.
3DGS has been proven to be effective in a wide range of applications, including digital human reconstruction~\cite{hugs,gaussianavatars}, Artificial Intelligence Generated Content~\cite{gaussiandreamer,text23dgs,repaint123}, and autonomous driving~\cite{drivinggaussian,sgsslam}.
In computational imaging, 3DGS has also shown significant improvements in super-resolution~\cite{wan2025s2gaussian}, deblurring~\cite{badgs,gsonmove}, derain~\cite{DeRainGS}, and low-light enhancement~\cite{LED3DGS}.
These advancements demonstrate the capacity of 3DGS to invert ill-posed imaging problems, transforming low-quality inputs into high-quality 3D reconstructions~\cite{uwnerf,deblurnerf,hdrnerf,Nan,nerfsr} with photometric consistency. 

\noindent\textbf{Computer vision with scattering medium}.
Underwater computer vision faces significant challenges due to complex optical phenomena, particularly light scattering and wavelength-dependent attenuation. These effects degrade image quality by introducing color distortion, reduced contrast, and haze, rendering traditional computer vision methods~(designed for clear-air environments) ineffective in underwater applications.
To address the ill-posed nature of the problem, earlier work introduces domain-specific priors to restore the scenes~\cite{berman2020underwater,peng2018generalization,peng2017underwater,zhao2024wavelet,xie2024uveb,li2019underwater}.
The method~\cite{underwatermodel} proposes a general image formation model in scattering media under ambient illumination, expressing per-pixel color $C$ as:
\begin{equation}
    C = \underbrace{c^{obj} \cdot (e^{-\sigma^{att}\cdot z})}_{\text{direct}} + \underbrace{c^{med} \cdot (1 - e^{-\sigma^{bs} \cdot z})}_{\text{backscatter}},
    \label{eq: watermodel}
\end{equation}
where $c^{obj}$ denotes the intrinsic color of the object at depth $z$, $c^{med}$ represents the ambient medium color at infinite distance, $\sigma^{att}$ and $\sigma^{bs}$ are the attenuation coefficients for the direct and backscatter components.
Building on this framework, SeaThru~\cite{Seathru-nerf} leverages depth maps to decouple attenuation and backscatter estimation, achieving robust restoration by explicitly modeling depth-dependent light propagation.
Osmosis~\cite{Osmosis} adopts an unsupervised diffusion framework that iteratively refines images using priors derived from unpaired clean and degraded datasets.
Furthermore, Seathru-NeRF~\cite{Seathru-nerf} and WaterSplatting~\cite{watersplatting} extend NeRF and 3DGS architectures by embedding the physical model into their rendering equations, enabling simultaneous 3D scene reconstruction and water removal.

\section{Preliminaries}
\label{sec: render}
In contrast to vanilla 3DGS, our Plenodium incorporates explicit modeling of medium-induced light attenuation through absorption and scattering effects, following~\cite{watersplatting}.
The transmittance $T_i(z)$ at a given depth $z$ along the ray, situated between the $(i$-$1)$-th and $i$-th Gaussian splat is formulated as:
\begin{equation}
    T_i(z) =T^{med}(z)\cdot T_i^{obj},
\end{equation}
where $T^{med}(z)=e^{-\sigma^{med}z}$ represents the exponential attenuation due to medium absorption, characterized by the medium's extinction coefficient $\sigma^{med}$ along the path from the camera to depth $z$, and $T^{obj}_i = \prod_{j=1}^{i-1}(1-\alpha_j)$ captures the cumulative transmittance through all preceding Gaussian primitives, quantifying their occlusion effects on downstream geometry.

Meanwhile, the medium's contribution to color $\hat C_i^{med}$ between these Gaussian splats is computed as:
\begin{equation}
        \hat C^{med}_i = \int_{z_{i-1}}^{z_i} c^{med}\sigma^{med}T_i^{obj}T^{med}(z) \,dz= c^{med}T^{obj}_i(e^{-\sigma^{med}z_{i-1}}-e^{-\sigma^{med}z_i}),
\end{equation} 
where $z_i$ denotes the depth of the $i$-th Gaussian in the camera coordinate system ($z_{0}$ is set to 0). 

Following~\cite{watersplatting,Seathru-nerf}, we utilize two sets of parameters: object attenuation $\sigma^{att}$ for object color $\hat C^{obj}$ and medium backscatter $\sigma^{bs}$ for medium color $\hat C^{med}$. The comprehensive rendering equation is:
\begin{equation}
    \begin{aligned}
        \hat C  =& \hat C^{obj} +\hat C^{med}
        = \sum_{i=1}^{N} \hat C_i^{obj} + \sum_{i=1}^{N} \hat C_i^{med} +\hat C_{\infty}^{med}\\
        =& \sum_{i=1}^{N} c_i \alpha_i T_i^{obj} e^{-\sigma^{att}z_{i}} 
        + \sum_{i=1}^N c^{med}T_i^{obj} (e^{-\sigma^{bs}z_{i-1}}-e^{-\sigma^{bs}z_i})
        + c^{med}T_{N+1}^{obj}e^{-\sigma^{bs}z_{N}}.
    \end{aligned}
        \label{eq: render_color}
\end{equation}
where $\hat C_{\infty}^{med}$ presents the medium's color contribution from the last Gaussian to infinitely far.

\begin{figure*}
    \includegraphics[width=\linewidth]{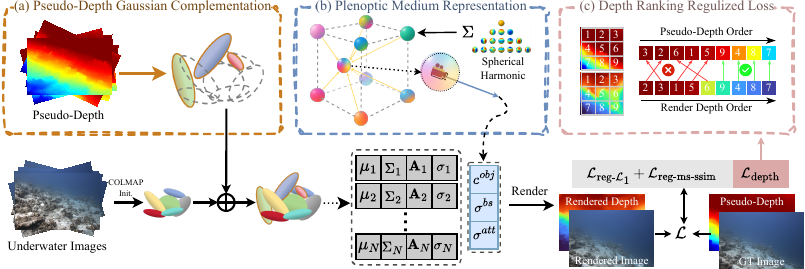}
    \vspace{-4mm}
    \caption{{Overview of our \emph{Plenodium.}}
    We first employ the pseudo-depth Gaussian complementation to enrich the primitives initialized by COLMAP.
    Then we utilize the plenoptic medium representation to estimate the medium parameter and render the underwater images following \eqnref{eq: render_color}.
    Our network is optimized with the loss function in \eqnref{eq: loss}, including a new depth ranking regularized loss.
    }
    \label{fig: overview}
    \vspace{-3mm}
\end{figure*}

\section{Plenodium}
We aim to develop an efficient and robust 3D reconstruction method for underwater scenarios.
Specifically, our approach begins with an explicit plenoptic medium representation (\ie, \figref{fig: overview}(b)) to accurately model the scattering medium in \secref{sec: sh}.
We then present a pseudo-depth Gaussian complementation (\ie, \figref{fig: overview}(a)) that leverages robust depth priors to enhance Gaussian primitive initialization in \secref{sec: PDGC}.
Finally, a depth ranking regularized loss (\ie, \figref{fig: overview}(c)) is introduced to enhance the geometric consistency in \secref{sec: loss}.
The whole framework is summarized in \figref{fig: overview}.


\subsection{Plenoptic Medium Representation}
\label{sec: sh}

Scattering effects in heterogeneous media exhibit complex dependencies on both ray direction $d$ and the observer's spatial coordinates $(x,y,z)$.
While existing methods only account for directional dependency while neglecting positional variations, we propose a novel plenoptic medium representation that explicitly incorporates both directional and positional information.
Specifically, we parameterize the properties of the medium along each ray (\ie, the medium color $c^{med}$, the object attenuation coefficient $\sigma^{att}$, and the medium backscatter coefficient $\sigma^{bs}$ in \eqnref{eq: render_color}) by Spherical Harmonics (SH):
\begin{equation}
    c^{med} = \text{SH}(  d, {A}^{c^{med}}_{x,y,z}),\quad
    \sigma^{att} = \text{SH}( d, {A}^{\sigma^{att}}_{x,y,z}),\quad
    \sigma^{bs} = \text{SH}(  d, {A}^{\sigma^{bs}}_{x,y,z}),
\end{equation}
where ${A}^{c^{med}}_{x,y,z},{A}^{\sigma^{att}}_{x,y,z},{A}^{\sigma^{bs}}_{x,y,z}$ denote the SH coefficients for medium color, object attenuation, and medium backscatter at position $(x,y,z)$, respectively.

To enable efficient learning of spatially-varying medium parameters from sparse camera observations, we store eight fundamental SH coefficients at each vertex of a 3D spatial grid. Through trilinear interpolation~\cite{zeng2020lut} of these, we can reconstruct the SH coefficient set at any arbitrary spatial position:
%
\begin{equation}
    {A}^t_{x,y,z} =\frac{1}{8} \sum_{u,v,w \in \{-1,1\}} (1+u\cdot x)(1+v\cdot y)(1+w\cdot z){A}_{u,v,w}^t, \quad \text{where } t \in \{ c^{med},\sigma^{att},\sigma^{bs} \}.
\end{equation}
%
The proposed plenoptic medium representation employs explicit SH encoding~\cite{Plenoxels}, reducing the time required for medium parameter retrieval during inference to less than 5\% of that needed by implicit MLP-based methods~\cite{watersplatting, Seathru-nerf}, while simultaneously improve the accuracy of scattering simulations in heterogeneous media (as illustrated in \secref{sec: alb}).

\subsection{Pseudo-Depth Gaussian Complementation}
\label{sec: PDGC}
The initialization of 3DGS in underwater environments presents unique challenges due to severe light attenuation and scattering.
These phenomena significantly degrade the performance of conventional structure-from-motion~(SfM) pipelines~\cite{schoenberger2016sfm}, particularly affecting tools like COLMAP~\cite{schoenberger2016mvs} that are commonly used for point cloud generation in 3DGS-based methods.
To overcome these limitations, we propose the Pseudo-Depth Gaussian Complementation (PDGC), which combines COLMAP's geometry with monocular depth priors to initialize Gaussian primitives.

First, similar to~\eqnref{eq: blend}, we render the pixel-wise depth $\hat D$ via $\alpha$-blending according to:  
\begin{equation}
    \hat D =   \sum_{i=1}^N (z_i \alpha_i \prod_{j=1}^{i-1}(1-\alpha_{i}))/(1 - T^{obj}_{N+1}).
    \label{eq: depth}
\end{equation}
Here, $T^{obj}_{N+1}$ represents the accumulated object transmission behind all $N$ Gaussians.

We then estimate the pseudo-depth map $\tilde D$ using Depth Anything Model~\cite{Depthanything,Depthanythingv2}, chosen for its superior generalization ability across diverse scenarios.
To address the scale ambiguity and offset biases in monocular depth estimation, we formulate an affine correction to calibrate the pseudo-depth:
\begin{equation}
        \tilde D'  = k \tilde D+b, 
\end{equation}
where the correction parameters $k$ and $b$ are optimized via least-squares minimization over well-initialized regions $\Omega_w = \{(x,y)| T_{N+1}^{obj}{(x,y)} <\tau_w\}$ ($\tau_w$ is the threshold for well-initialized regions):
\begin{equation}
        \mathop{\arg\min}\limits_{k,b}\sum_{(x,y)\in \Omega_w}\left(\hat D(x,y) - k \tilde{D}(x,y) -b\right)^2.
    \label{eq: fit}
\end{equation}
The refined pseudo-depth $\tilde{D}'$ exhibits significantly mitigated scale ambiguity and reduced offset biases compared to the original pseudo-depth $\tilde{D}$, as quantified through alignment with the depth estimated from initialized scenes. 
Subsequently, we insert new Gaussian primitives in domain $\Omega_n \cap \Omega_p$, specifically targeting regions exhibiting both proximal camera distance ($\Omega_n =  \{ (x,y)| \tilde{D}(x,y) < \tau_{near}\cdot \text{max}(\tilde{D})\}$) and elevated transmittance values ($\Omega_p = \{(x,y)| T^{obj}_{N+1}{(x,y)} \geq \tau_w\}$). 
This targeted placement strategy effectively minimizes background interference while preventing oversampling, thereby improving both the computational efficiency and reconstruction accuracy of our approach.

%

Our pseudo-depth Gaussian complementation method enriches the initialized Gaussian primitives across diverse scenes (\tabref{tab: pdgc}), improving the robustness of 3DGS initialization against degradations.

\subsection{Loss Function}
\label{sec: loss}
Building upon differentiable 3DGS frameworks, we develop a multi-objective optimization pipeline for primitive refinement, jointly enforcing photometric accuracy, structural coherence, and depth consistency.
Following \cite{watersplatting,rawnerf}, we incorporate a weighting matrix $W = \frac{1}{\text{sg}(\hat C)+\epsilon}$ (with $\epsilon= 10^{-6}$, $\text{sg}(\cdot)$ means stop gradient) to emphasize dark regions during optimization, aligning with human perceptual sensitivity to dynamic range.
Then, based on the L1 loss and the multi-scale differentiable SSIM loss~\cite{msssim}, we employ two regularized losses:
\begin{equation}
        \mathcal{L}_{\text{reg-}\mathcal{L}_1} = \mathcal{L}_{1}(W\odot \hat C,W\odot C),\quad
        \mathcal{L}_\text{reg-ms-ssim} = \mathcal{L}_{\text{ms-ssim}}(W\odot \hat C,W\odot C),
    \label{eq: reg}
\end{equation}
for photometric accuracy and structural coherence, respectively.
Notably, we utilize the regularized multi-scale differentiable SSIM loss $\mathcal{L}_\text{reg-ms-ssim}$ rather than the single-scale version $\mathcal{L}_\text{reg-ssim}$~\cite{ssim} used in previous works~\cite{watersplatting} for a larger perceptual field.

In addition, to enforce the ordinal stability on depth maps~\cite{xian2020structure,wang2023sparsenerf} for 3DGS, we first downsample the pseudo-depth map $\tilde D$ and the rendered depth map $\hat D$ to low-resolution variants $\tilde{D}^*$ and $\hat D^*$ of size $N\times N$ pixels using bilinear pooling, which suppresses high-frequency noise while preserving relative depth ordering. We then introduce a new depth ranking regularized loss that penalizes violations of ordinal relationships between depth estimates:
%
\begin{equation}
    \mathcal{L}_{depth} = \frac{1}{N^4}\sum_{i,j} \text{min}(-(\tilde D^*_i - \tilde D^*_j)(\hat D^*_i - \hat D^*_j),0).
\end{equation}
This loss function maintains scale invariance and reduces dependence on fine-grained structural details by operating on the downsampled representations.
Benefiting from this loss function, our approach achieves robust depth optimization even when initialized with imprecise pseudo-depth estimates, as demonstrated in \tabref{tab: loss}.

The final loss function is given as:
\begin{equation}
        \mathcal{L} = \lambda_{\mathcal{L}_1}\mathcal{L}_{\text{reg-}\mathcal{L}_1} + \lambda_{\text{ssim}}\mathcal{L}_\text{reg-ms-ssim}  + \lambda_{\text{depth}}\mathcal{L}_\text{depth},
    \label{eq: loss}
\end{equation}
where the parameters $\lambda_{\mathcal{L}_1}$, $\lambda_{\text{ssim}}$, and $\lambda_{\text{depth}}$ are utilized to balance the different loss components.
  
\section{Experimental Results}
\label{sec:experiments}
In this section, we first describe the implementation details of our approach and the datasets we used.
Then we evaluate the proposed method on both simulated and real underwater scenarios.

\subsection{Implementation and Datasets}
\noindent\textbf{Implementation details.}
Our implementation is based on the Nerfstudio\cite{gsplat,nerfstudio}.
After COLMAP initialization, we use the pseudo-depth Gaussian complementation with $\tau_w = 0.99$ and $\tau_\text{near}=0.5$ to enrich the initial set of 3D Gaussians.
During training, we empirically set $\lambda_{\mathcal{L}_1}=0.8$, $\lambda_{\text{ssim}}=0.2$, and $\lambda_{\text{depth}}=5$ in \eqnref{eq: loss}. The patch number $N$ is set to 16.
The max degree of the SH coefficients for the medium and Gaussian primitives is set as 3.
We accumulated the absolute gradient norms of $\mu$ for finer densification following~\cite{absgs}. All the experiments are conducted on an RTX4090 GPU.

\noindent\textbf{SeathruNeRF dataset.}
The SeaThru-Nerf dataset~\cite{Seathru-nerf} includes four real underwater scenes: IUI3 Red Sea, Cura\c{c}ao, Japanese Gardens Red Sea, and Panama. There are 29, 20, 20, and 18 images in each scene, respectively, where 25, 17, 17, and 15 images are used for training and the rest of the images are for testing. These images are captured in RAW format by a Nikon D850 SLR camera in underwater conditions. Following~\cite{watersplatting,Seathru-nerf}, these images are downsampled to an averaged resolution of $900\times 1400$ pixels and COLMAP~\cite{schoenberger2016sfm,schoenberger2016mvs} is employed to determine the camera poses and produce sparse point clouds for the initialization of 3DGS-based methods.

\noindent\textbf{Our simulated dataset.}
We utilize Blender to simulate a dataset with precise GT for restoration evaluation. The dataset includes two scenes (beach and street), each degraded by two types of media (fog~\cite{schechner2001instant} and water) at three incremental intensity levels (easy, medium, and hard), yielding 12 systematically structured subsets. Each subset contains 100 images at a resolution of 512$\times$512 pixels. We split them evenly into 50 training and 50 testing samples to ensure a balanced evaluation.
Compared to existing simulated datasets~\cite{watersplatting,Seathru-nerf}, our dataset provides enhanced accuracy, scale, and diversity, enabling robust evaluation across different media types and degradation intensities.
To isolate restoration quality from exposure, we preprocess restored images by linearly scaling their intensity to align with the GT's mean luminance before computing evaluation metrics.
Since Blender provides accurate camera poses, we use COLMAP solely to generate sparse point clouds for initializing the set of 3D Gaussians.
More details are provided in the supplemental material.

\begin{table}[t]
\vspace{-2mm}
    \centering
    \scriptsize
    \setlength{\tabcolsep}{1.1mm}
    \newcommand{\f}[1]{{\textcolor{red}{\bf{{#1}}}}}
    \newcommand{\s}[1]{{\textcolor{blue}{\underline{{#1}}}}} 
    \caption{{Recontruction performance on the SeaThru-NeRF~\cite{Seathru-nerf} dataset.} We report the PSNR$\uparrow$,SSIM$\uparrow$, and LPIPS$\downarrow$ scores for the four underwater scenes. The average FPS$\uparrow$ and Training Time$\downarrow$ on an RTX4090 are also provided. The best and second-best results are \f{bolded} and \s{underlined}, respectively.}
    \vspace{-1mm}
    \begin{tabularx}{\linewidth}{l|ccc|ccc|ccc|ccc|c|c}
        \toprule[1pt]
                                             & \multicolumn{3}{c|}{ IUI3 Red Sea} & \multicolumn{3}{c|}{Cura\c{c}ao} & \multicolumn{3}{c|}{J.G. Red Sea} & \multicolumn{3}{c|}{ Panama} &           &                                                                                                                                \\

        \multirow{-2}{*}{Methods}        & PSNR                                  & SSIM                                 & LPIPS                                 & PSNR                            & SSIM      & LPIPS     & PSNR       & SSIM      & LPIPS     & PSNR       & SSIM      & LPIPS     & \multirow{-2}{*}{\makecell{Avg.          \\FPS}} & \multirow{-2}{*}{\makecell{Avg.\\Time}} \\\midrule
        ZipNeRF~\cite{zipnerf}               & 16.937                                & 0.474                                & 0.412                                 & 19.956                          & 0.442     & 0.421     & 19.022     & 0.349     & 0.483     & 19.012     & 0.349     & 0.482     & 0.17                            & 5h     \\
        SeaThru-NeRF~\cite{Seathru-nerf}     & 26.755                                & 0.826                                & \f{0.168}                             & 30.959                          & 0.915     & 0.133     & 23.282     & 0.876     & \f{0.111} & 31.276     & 0.937     & \f{0.071} & 0.68                            & 2h     \\
        3D-Gauss.~\cite{3dgs}                & 22.980                                & 0.843                                & 0.246                                 & 28.313                          & 0.873     & 0.221     & 21.493     & 0.854     & 0.216     & 29.200     & 0.893     & 0.152     & 318                             & 13min  \\
        WaterSplatting~\cite{watersplatting} & \s{29.840}                            & \s{0.889}                            & \s{0.203}                             & \s{32.203}                      & \s{0.948} & \s{0.116} & \s{24.741} & \s{0.892} & \s{0.116} & \s{31.616} & \s{0.942} & 0.080     & 169                             & 6.9min \\
        \rowcolor{gray!20}
        \bf Plenodium                    & \f{30.275}                            & \f{0.895}                            & {0.205}                               & \f{34.120}                      & \f{0.953} & \f{0.110} & \f{25.058} & \f{0.896} & {0.121}   & \f{32.435} & \f{0.946} & \s{0.074} & 249                             & 7.0min \\\bottomrule[1pt]
    \end{tabularx}
    \vspace{-3mm}
    \label{tab: rec}
\end{table}
\begin{table}[t]
    \centering
    \footnotesize
    \scriptsize
    \setlength{\tabcolsep}{1.32mm}
    \newcommand{\f}[1]{{\textcolor{red}{\bf{{#1}}}}}
    \newcommand{\s}[1]{{\textcolor{blue}{\underline{{#1}}}}} 
    \newcommand{\tc}[1]{\multicolumn{3}{c|}{#1}}
    \newcommand{\sixc}[1]{\multicolumn{6}{c|}{#1}}
    \caption{{Restoration performance on our simulated dataset.} We report the PSNR$\uparrow$, and SSIM$\uparrow$ for each subset scenes. The best and second-best results are \f{bolded} and \s{underlined}, respectively.}
    \vspace{-1mm}
    \begin{tabularx}{\linewidth}{l|ccc|ccc|ccc|ccc}
        \toprule[1pt]
        \multicolumn{1}{c}{}                     & \sixc{Beach} & \multicolumn{6}{c}{Street}                                                                                                                                                 \\\midrule
        \multirow{2}{*}{Methods}             & \tc{PSNR}     & \tc{SSIM}                     & \tc{PSNR}  & \multicolumn{3}{c}{SSIM}                                                                                                         \\
                                                 & easy          & medium                        & hard       & easy                     & medium     & hard       & easy       & medium     & hard       & easy       & medium     & hard       \\\midrule[1pt]
        \multicolumn{13}{c}{\it {FOG}}                                                                                                                                                                                                    \\\midrule
        SeaThru-NeRF~\cite{Seathru-nerf}     & \f{28.416} & \s{24.799} & \s{16.439} & \s{0.9228} & \s{0.8888} & \s{0.7856} & \s{27.677} & 23.562 & 17.730 & 0.8697 & 0.8286 & 0.7244   \\
         WaterSplatting~\cite{watersplatting} & 17.725 & 16.724 & 15.492 & 0.8549 & 0.7728 & 0.7475 & 26.595 & \s{24.816} & \s{22.510} & \s{0.8924} & \s{0.8551} & \f{0.8003}\\
        \rowcolor{gray!20}
        \bf Plenodium                        & \s{26.372} & \f{26.248} & \f{25.275} & \f{0.9435} & \f{0.9259} & \f{0.9051} &  \f{27.978} & \f{26.071} & \f{23.060} & \f{0.9107} & \f{0.8791} & \s{0.7982} \\\midrule[1pt]
        \multicolumn{13}{c}{ \it {WATER}}                                                                                                                                                                                                 \\\midrule
        SeaThru-NeRF~\cite{Seathru-nerf}     & \s{25.938} & \s{19.500} & \s{16.495} & \s{0.9094} & \s{0.8590} & \s{0.8085} & \s{25.268} & 21.414 & 19.420 & 0.8533 & 0.7919 & 0.7467\\
        WaterSplatting~\cite{watersplatting} & 17.592 & 15.892 & 15.021 & 0.8198 & 0.7725 & 0.7450 &  24.936 & \s{22.120} & \s{20.205} & \f{0.8793} & \f{0.8240} & \s{0.7546}  \\
        \rowcolor{gray!20}
        \bf Plenodium                        & \f{26.360} & \f{24.723} & \f{23.636} & \f{0.9406} & \f{0.9119} & \f{0.8867} & \f{25.731} & \f{22.392} & \f{20.285} & \s{0.8594} & \s{0.7983} & \f{0.7556} \\\bottomrule[1pt]
    \end{tabularx}
    \label{tab: res}
    \vspace{-5mm}
\end{table}

\subsection{Quantitative Results}
We compare our approach against several state-of-the-art methods, including two NeRF-based ones (Zip-NeRF~\cite{zipnerf} and SeaThru-NeRF~\cite{Seathru-nerf}) and two Gaussian splatting-based ones (3DGS~\cite{3dgs} and WaterSplatting~\cite{watersplatting}). 
Note that SeaThru-NeRF and WaterSplatting are specifically tailored for underwater scene reconstruction.
We use PSNR, SSIM, and LPIPS as metrics to evaluate the rendering quality. 
In addition, we benchmark the total training time and FPS (frame per second) on a machine with an RTX 4090 GPU to provide a comprehensive comparison of computational efficiency.


We first examine the 3D reconstruction performance of Plenodium on the SeaThru-NeRF dataset. As shown in \tabref{tab: rec}, Plenodium performs favorably against state-of-the-art methods, increasing the PSNR and SSIM values by 0.872dB and 0.047 on average and improving the rendering speed by 47\% (from 169FPS to 249FPS) over the best-competing method. 
Our approach substitutes the computationally expensive MLPs used in conventional methods with a lightweight SH-based spectral decomposition, significantly accelerating inference while preserving representational accuracy.

We then evaluate the restoration performance of Plenodium on our simulated dataset with controllable scattering media in \tabref{tab: res}. 
Compared to state-of-the-art methods~\cite{Seathru-nerf,watersplatting}, our Plenodium achieves better results, especially in challenging cases.
This improvement stems from two key factors.
First, benefitting from the proposed plenoptic medium representation, our approach is able to reconstruct 3D scenes more accurately.
Second, the enhanced robustness of our approach originates from the integration of learned depth priors extracted from the Depth Anything Model, 
which supports:
(i) pseudo-depth complementation to densify sparse points for initialization, and
(ii) a ranking-based depth loss that enforces ordinal consistency during optimization.
Taken together, these components allow Plenodium to achieve high-fidelity reconstructions under heterogeneous medium conditions and maintain reliable performance across varying levels of visual degradation.




\begin{figure*}[t]
    \centering
    \newcommand{\tc}[1]{\multicolumn{2}{@{}c}{#1}} 
    \setlength{\tabcolsep}{0.5pt}
    \small
    \def\fs{0.165\linewidth} 
    \def\sfs{0.0815\linewidth} 
    \begin{tabular}{@{}cccccccccccc@{}}
        \tc{\includegraphics[width=\fs]{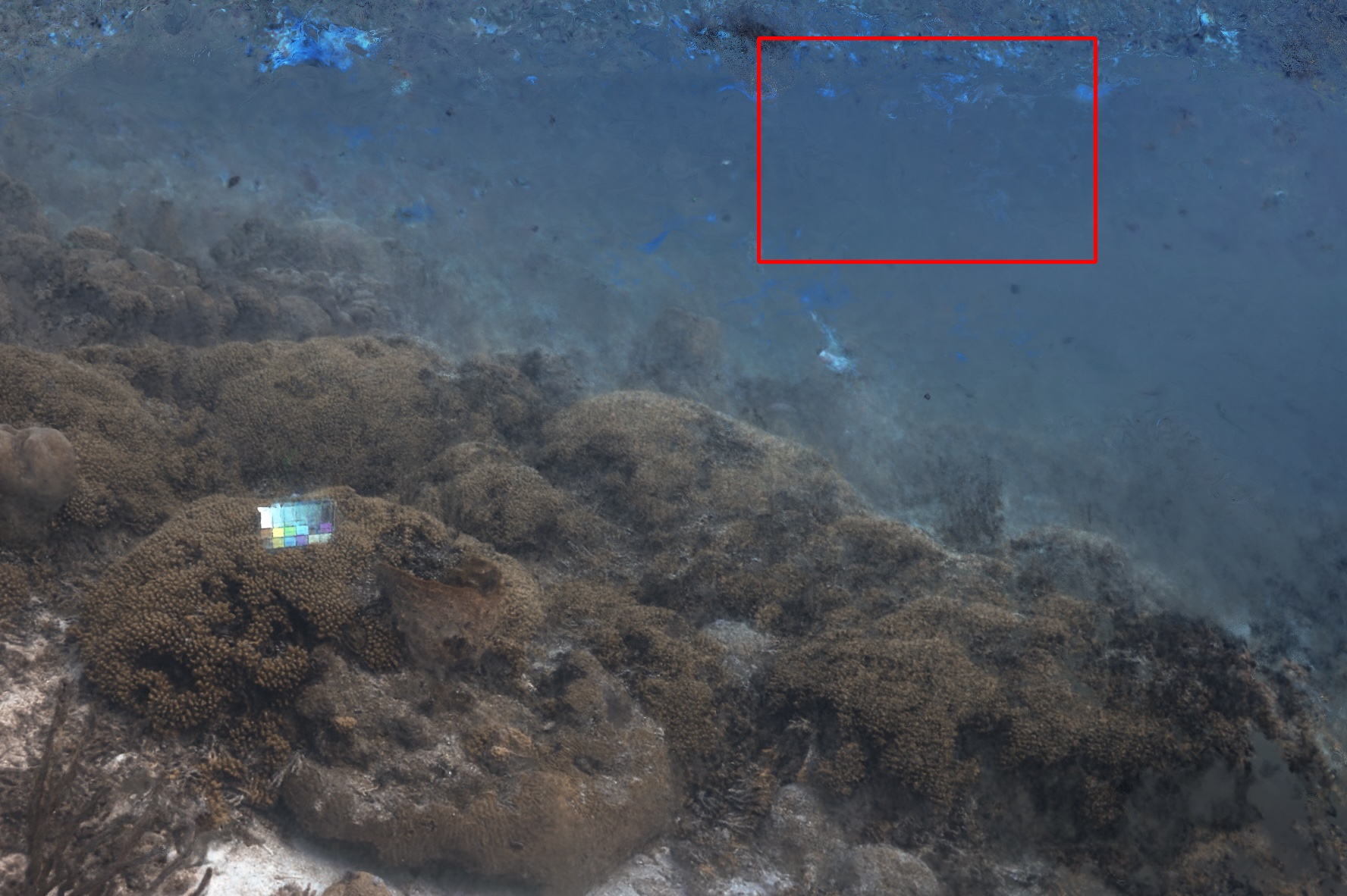}}              &
        \tc{\includegraphics[width=\fs]{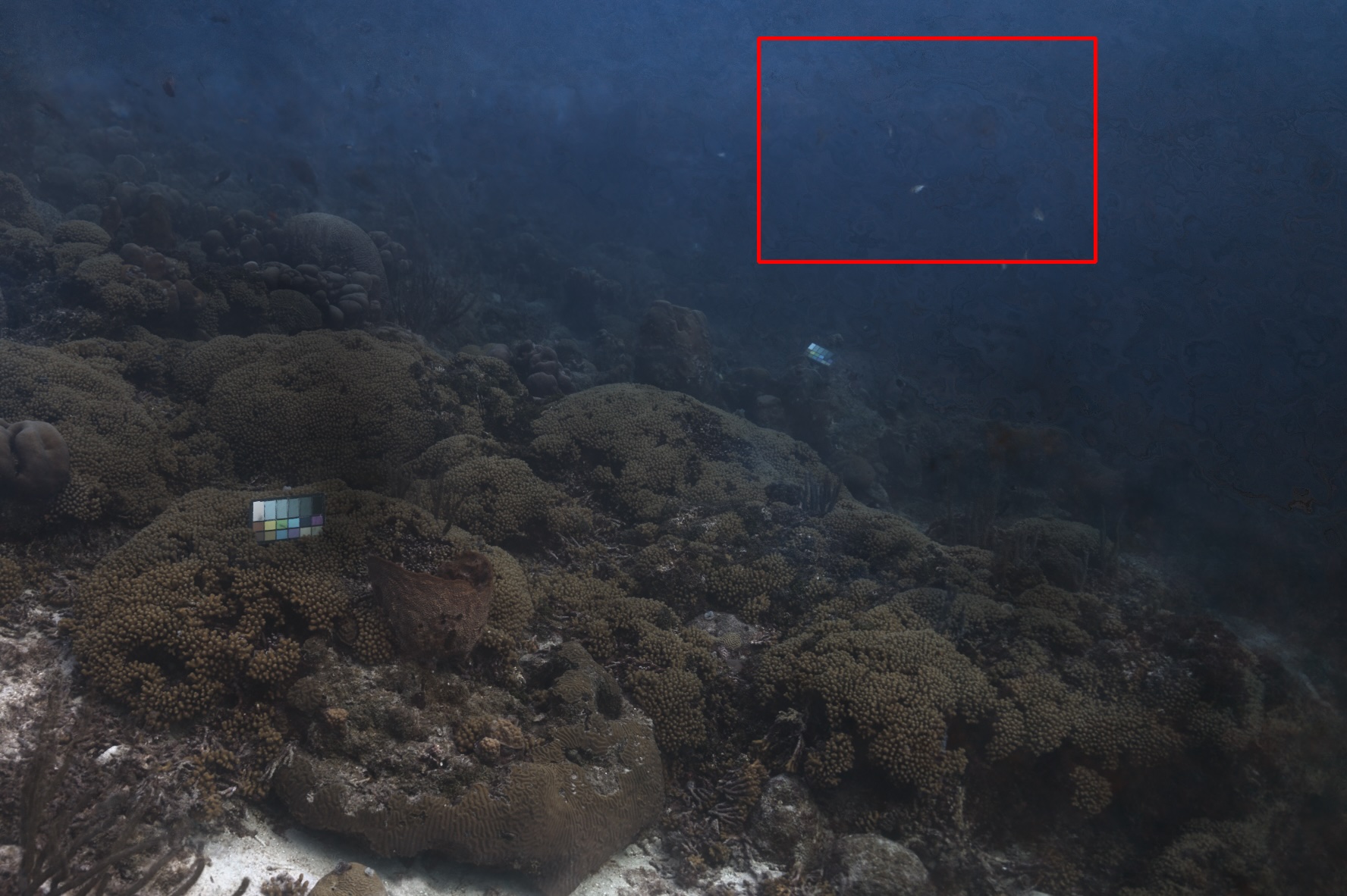}}          &
        \tc{\includegraphics[width=\fs]{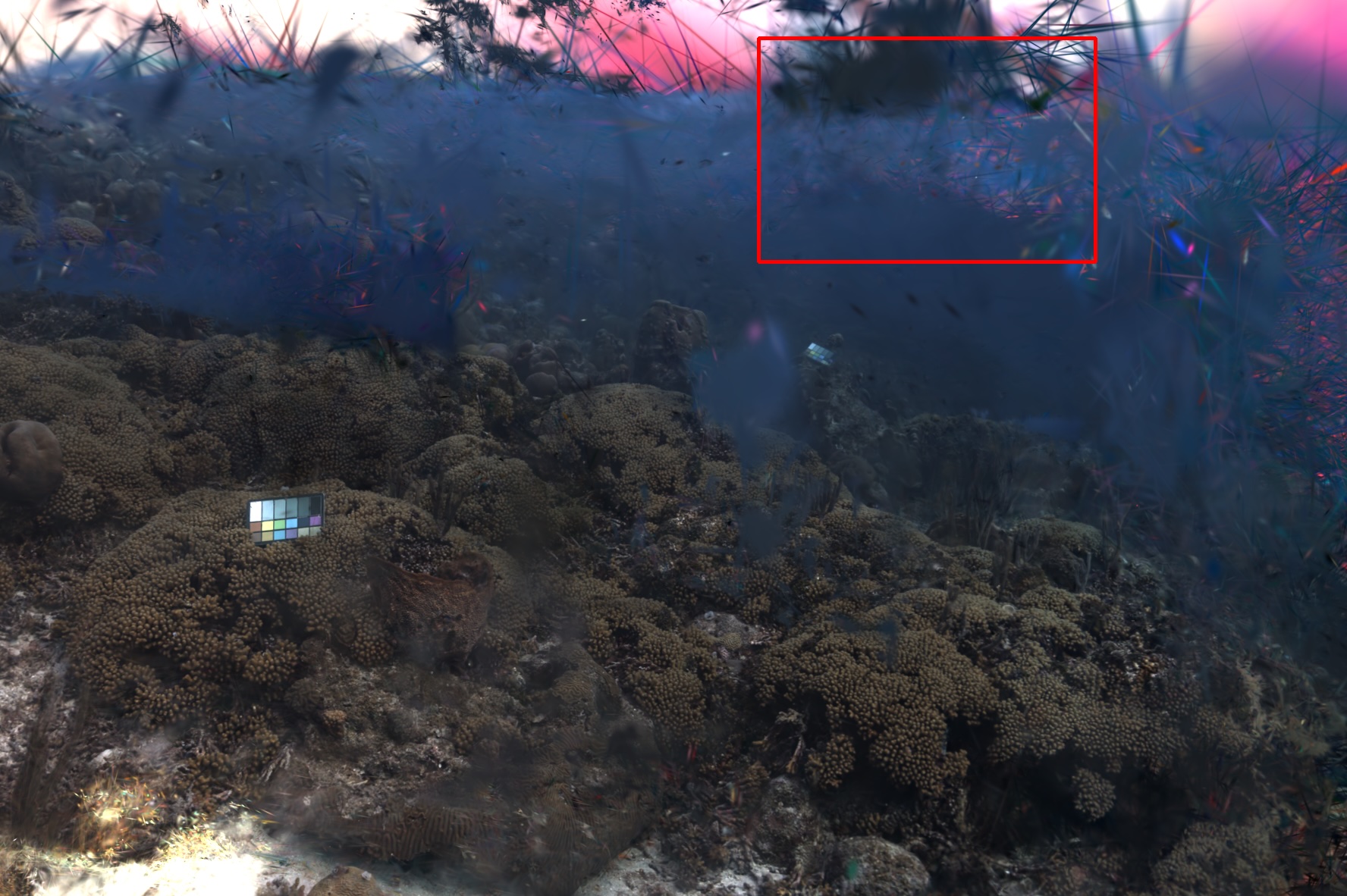}}                  &
        \tc{\includegraphics[width=\fs]{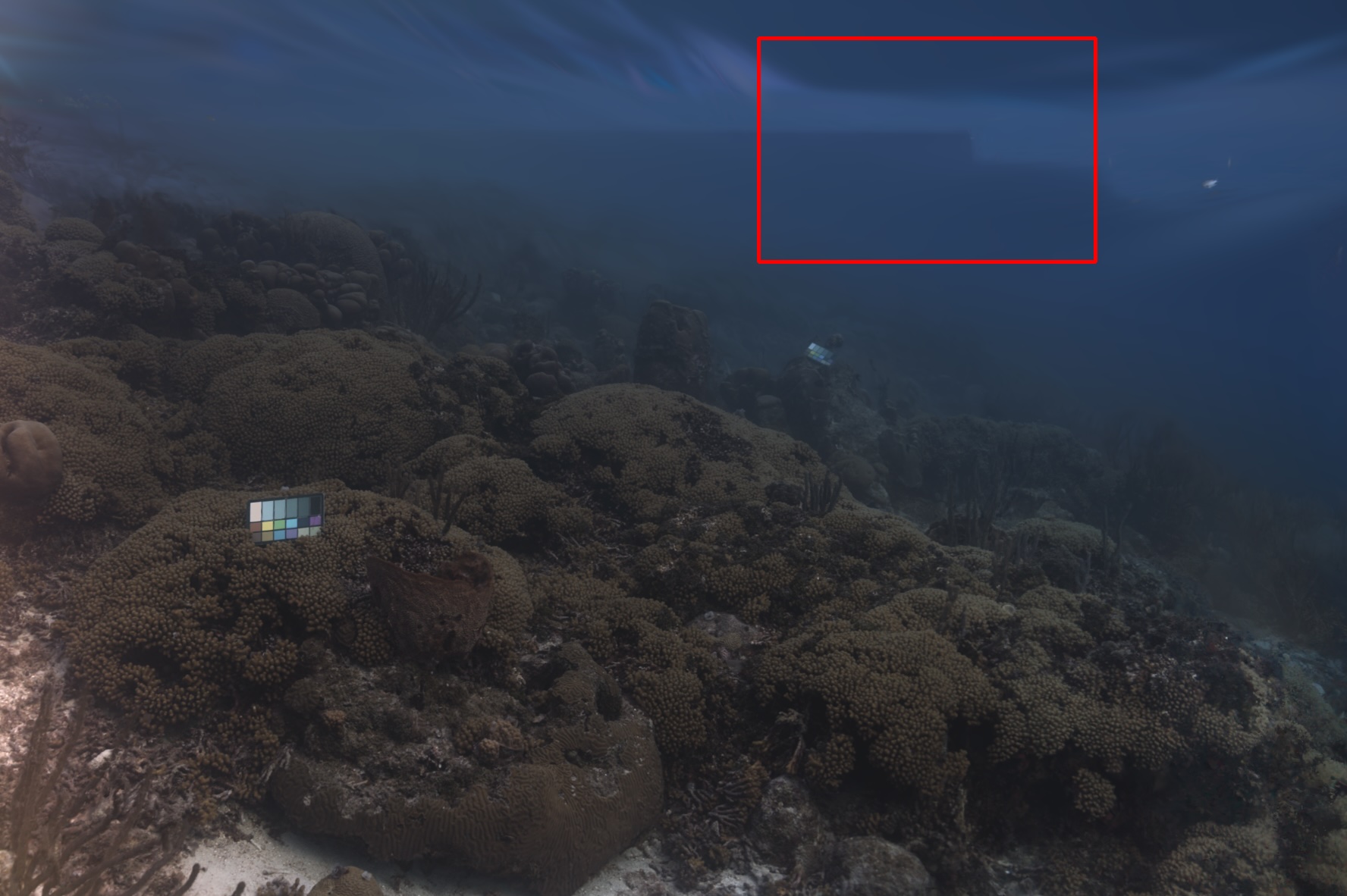}}       &
        \tc{\includegraphics[width=\fs]{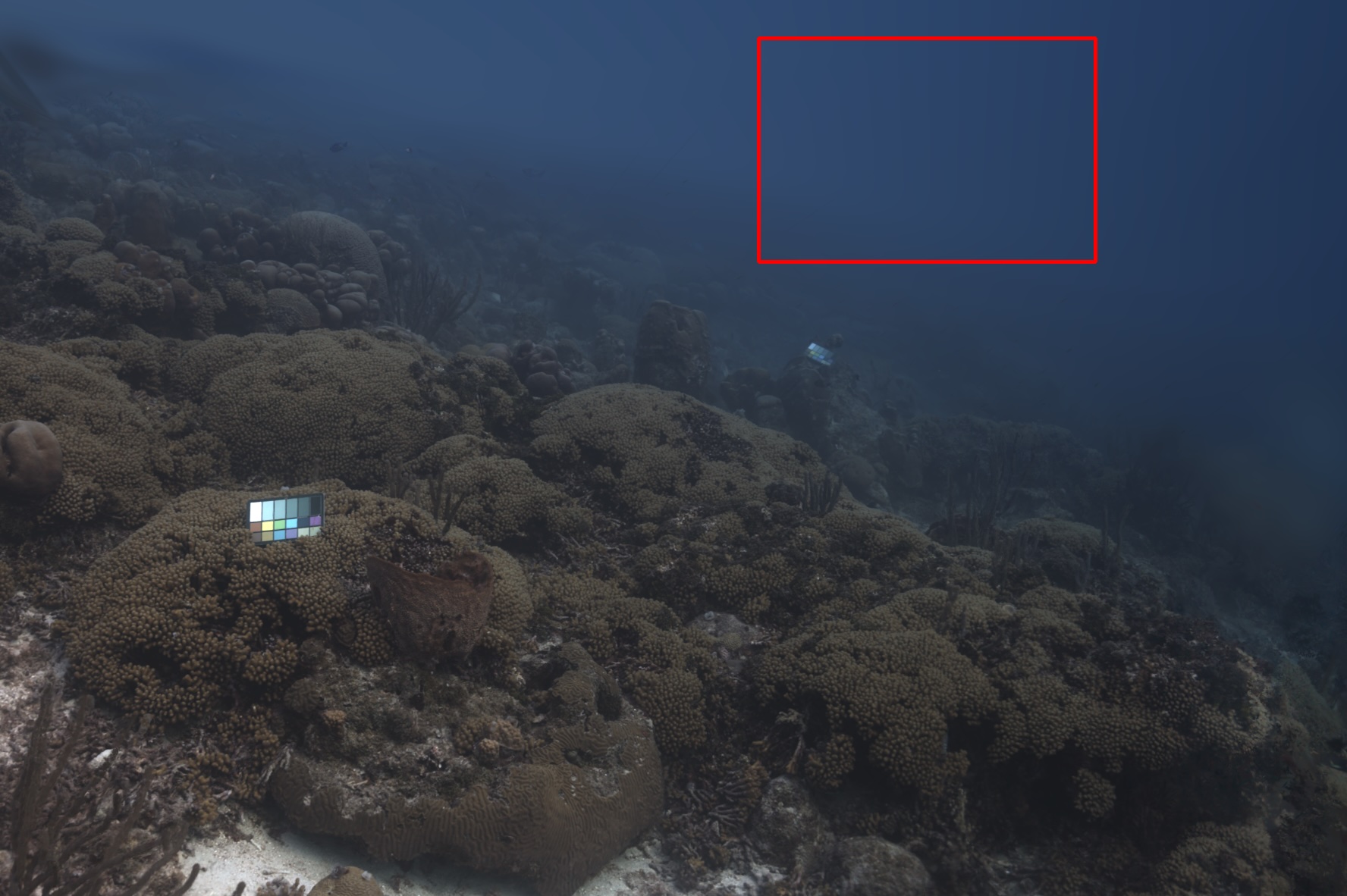}}     &
        \tc{\includegraphics[width=\fs]{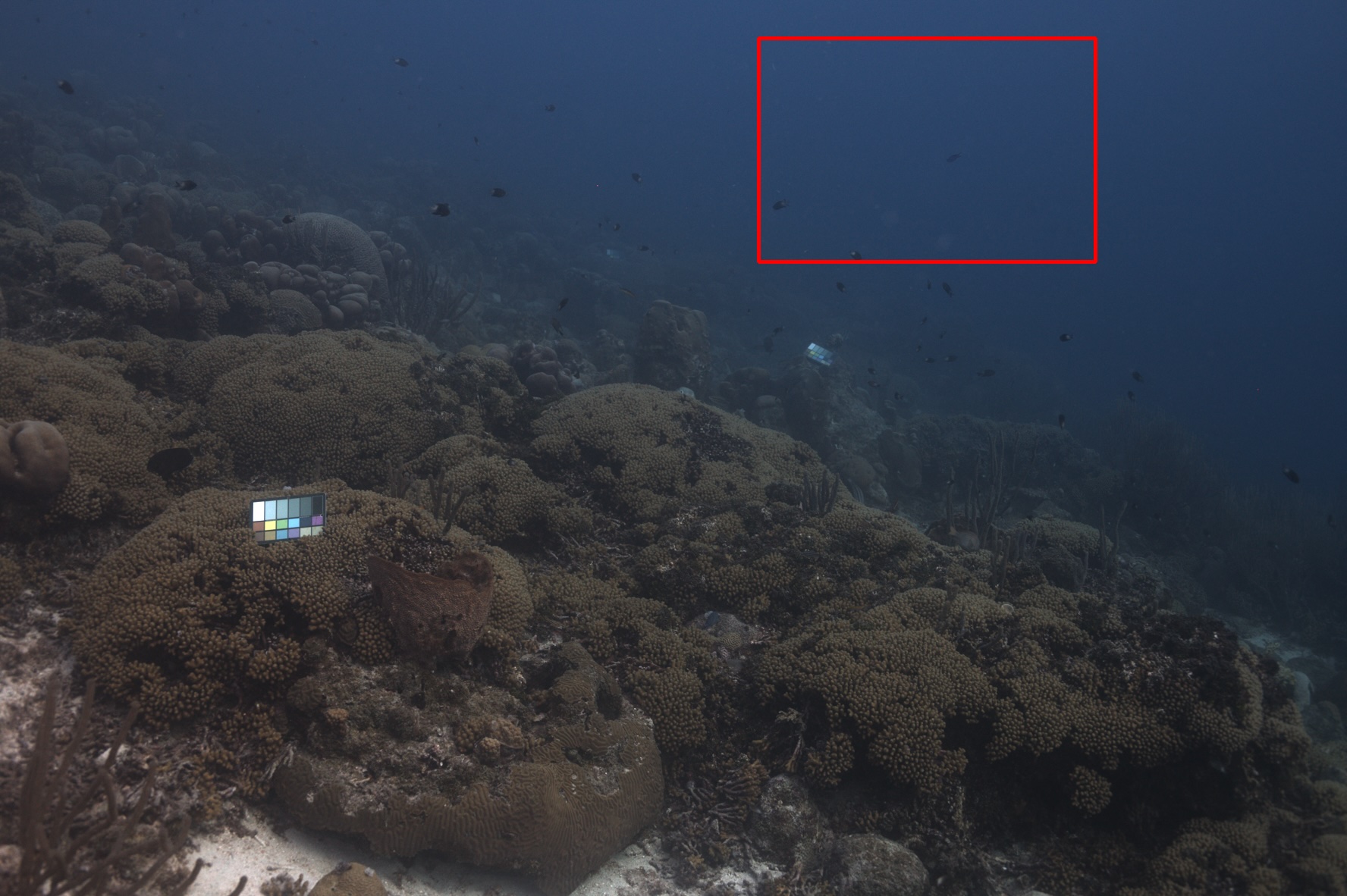}}                      \\
        \includegraphics[width=\sfs]{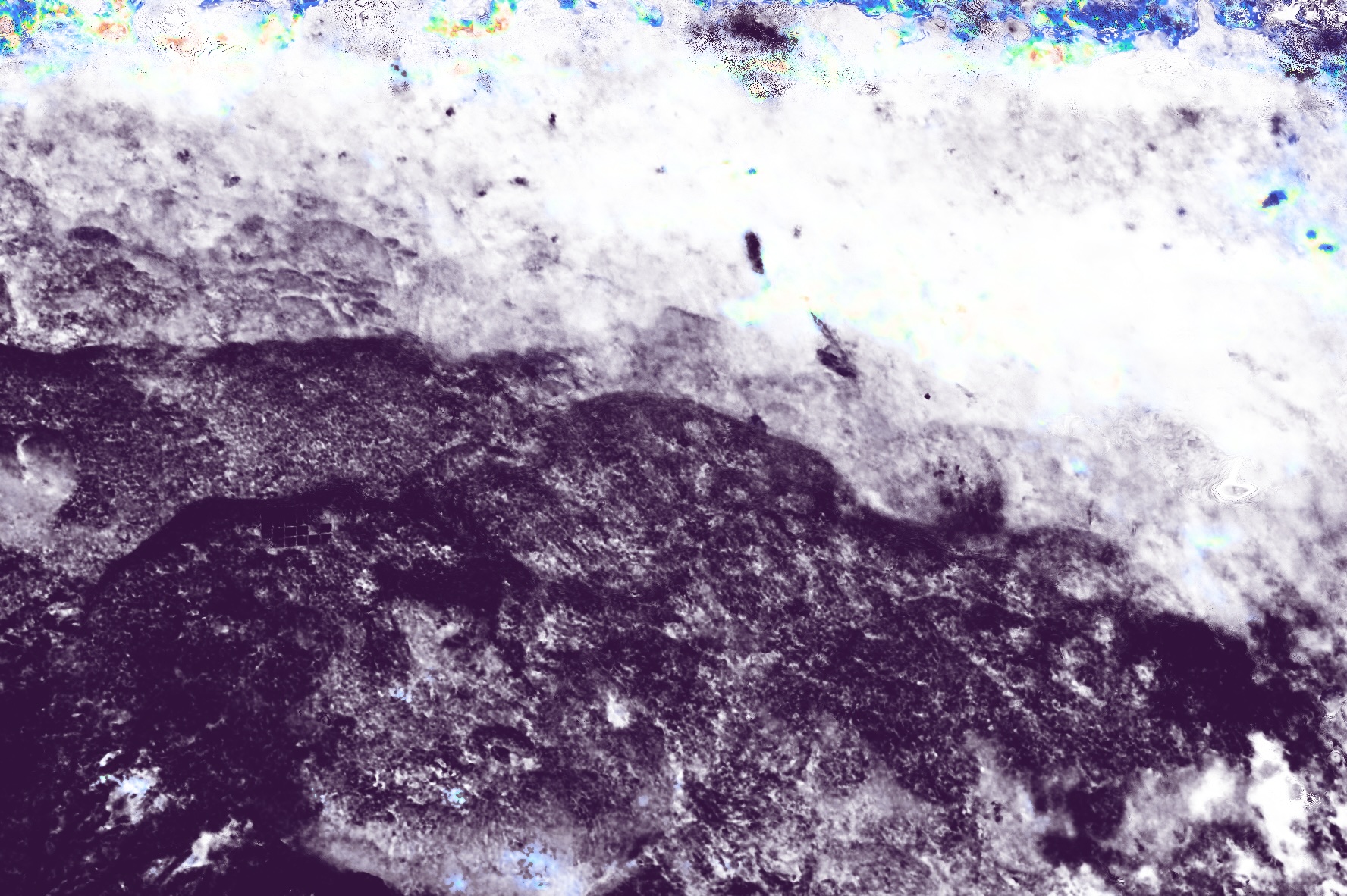}                 &
        \includegraphics[width=\sfs]{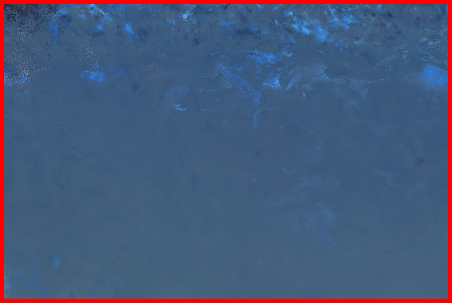}                  &
        \includegraphics[width=\sfs]{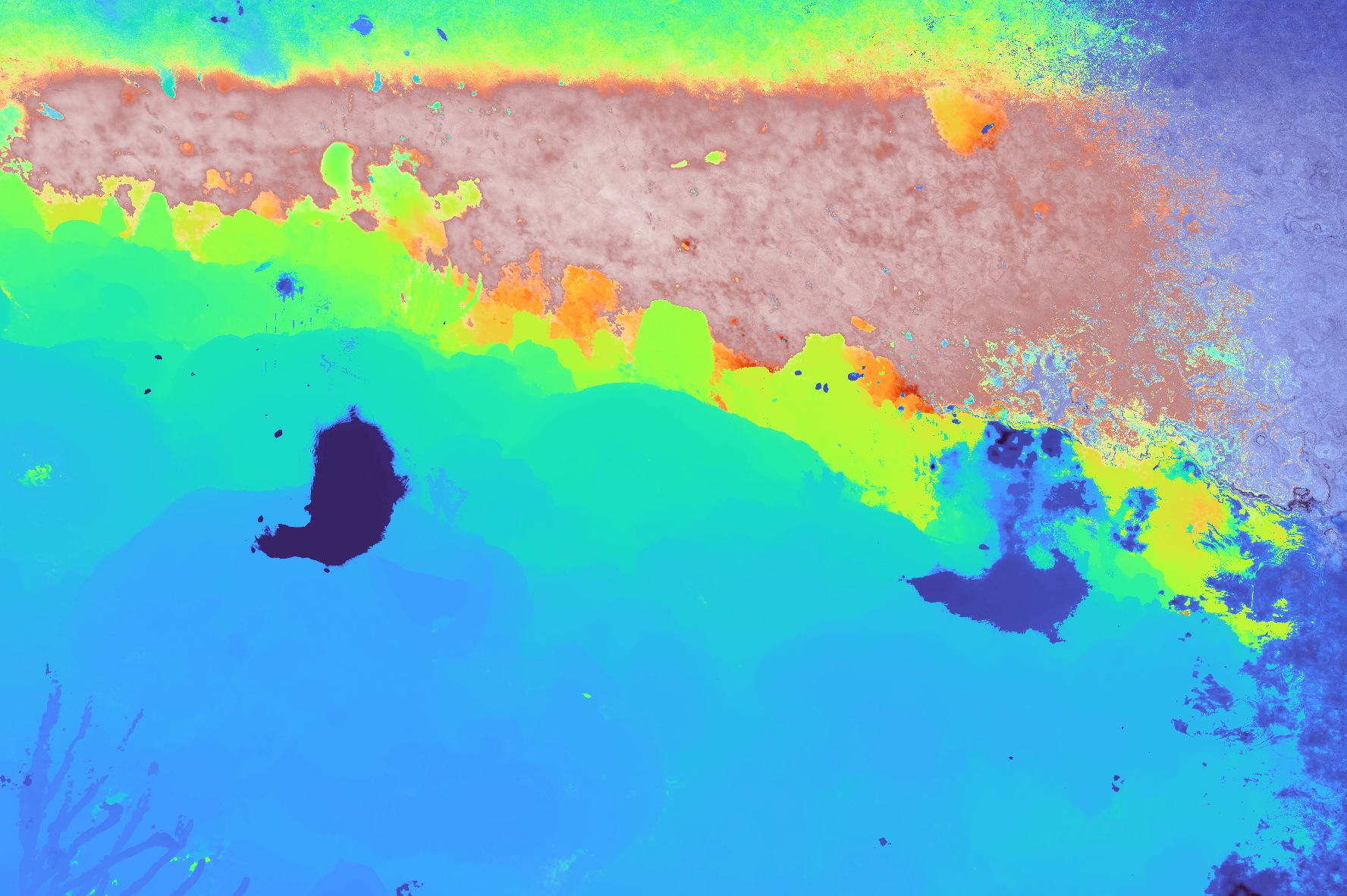}             &
        \includegraphics[width=\sfs]{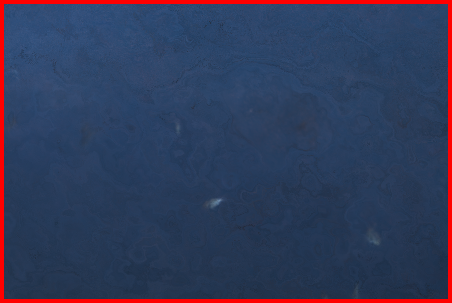}              &
        \includegraphics[width=\sfs]{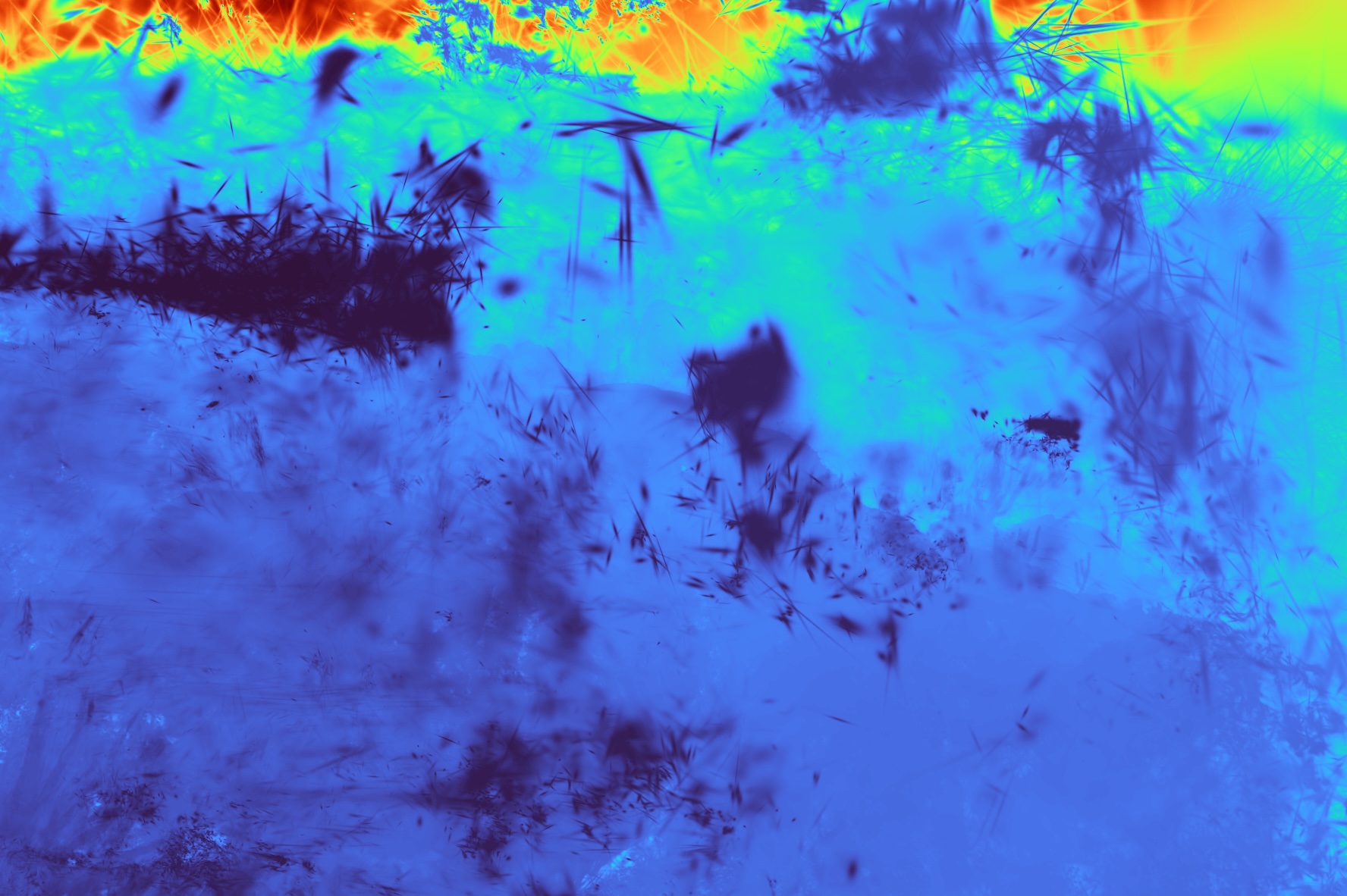}                     &
        \includegraphics[width=\sfs]{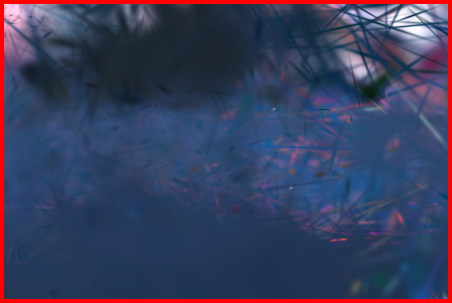}                      &
        \includegraphics[width=\sfs]{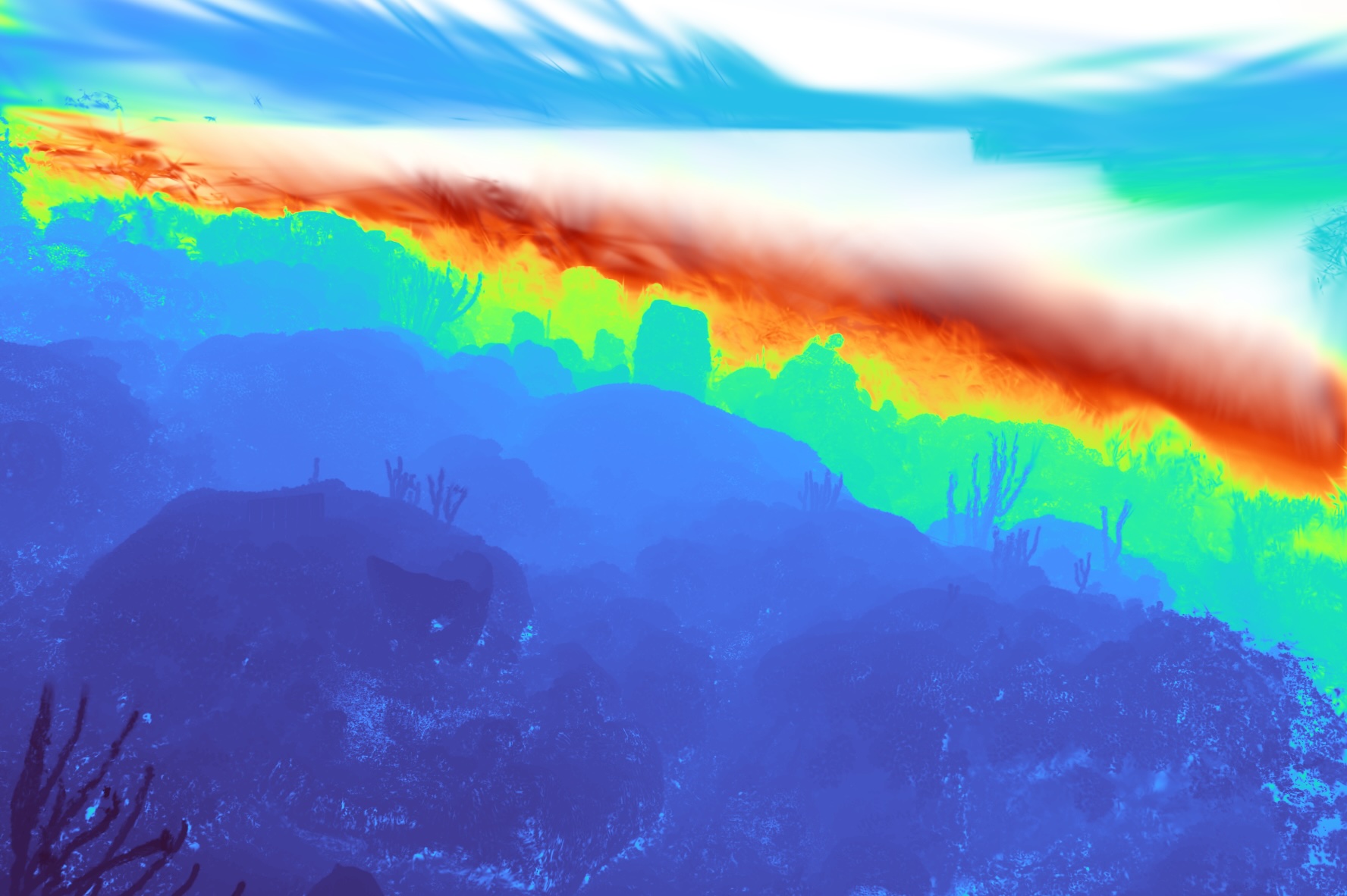}          &
        \includegraphics[width=\sfs]{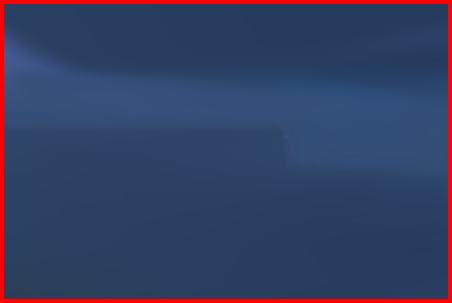}           &
        \includegraphics[width=\sfs]{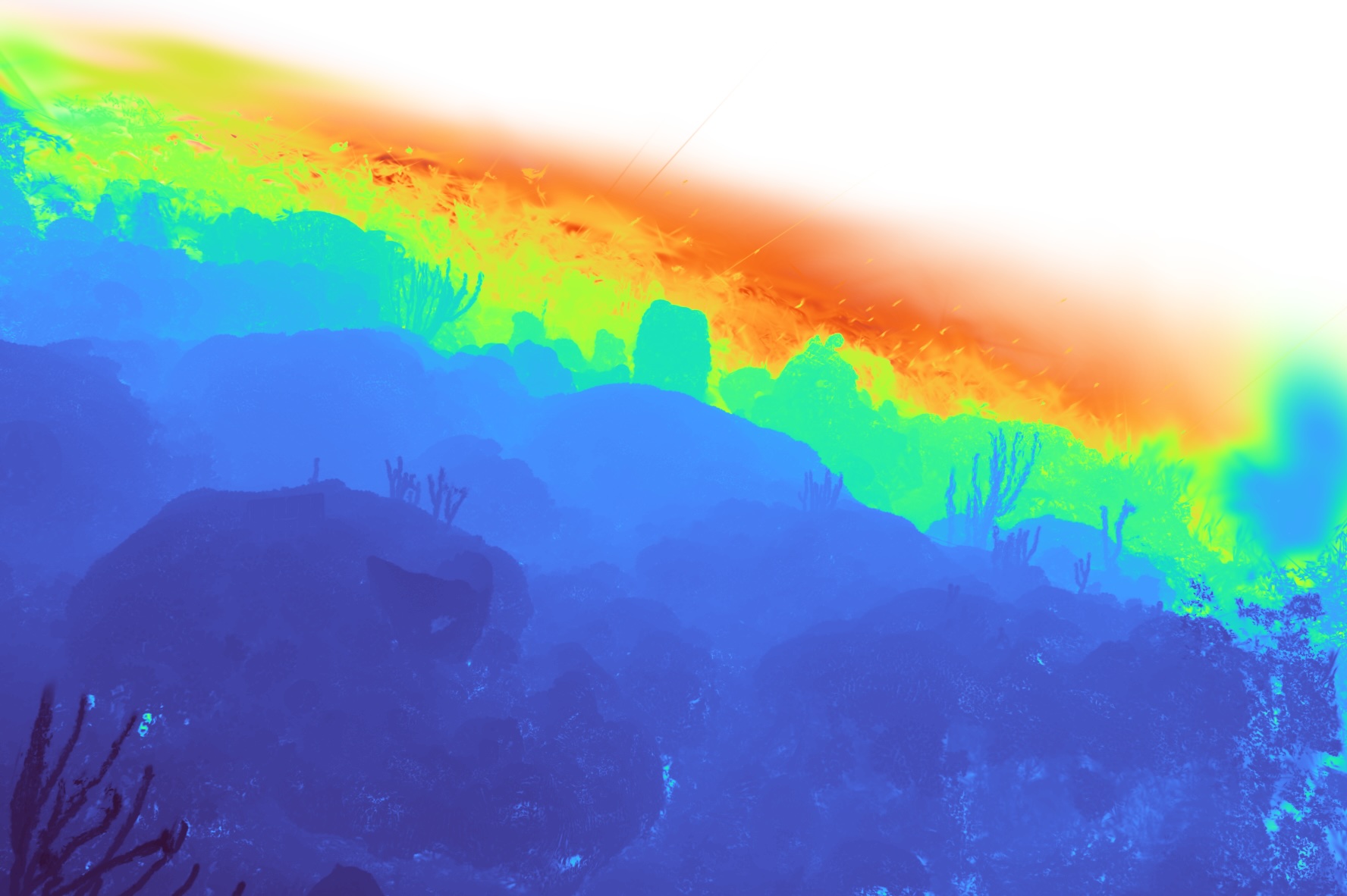}        &
        \includegraphics[width=\sfs]{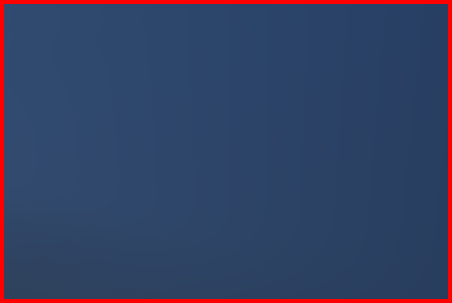}         &
        \includegraphics[width=\sfs]{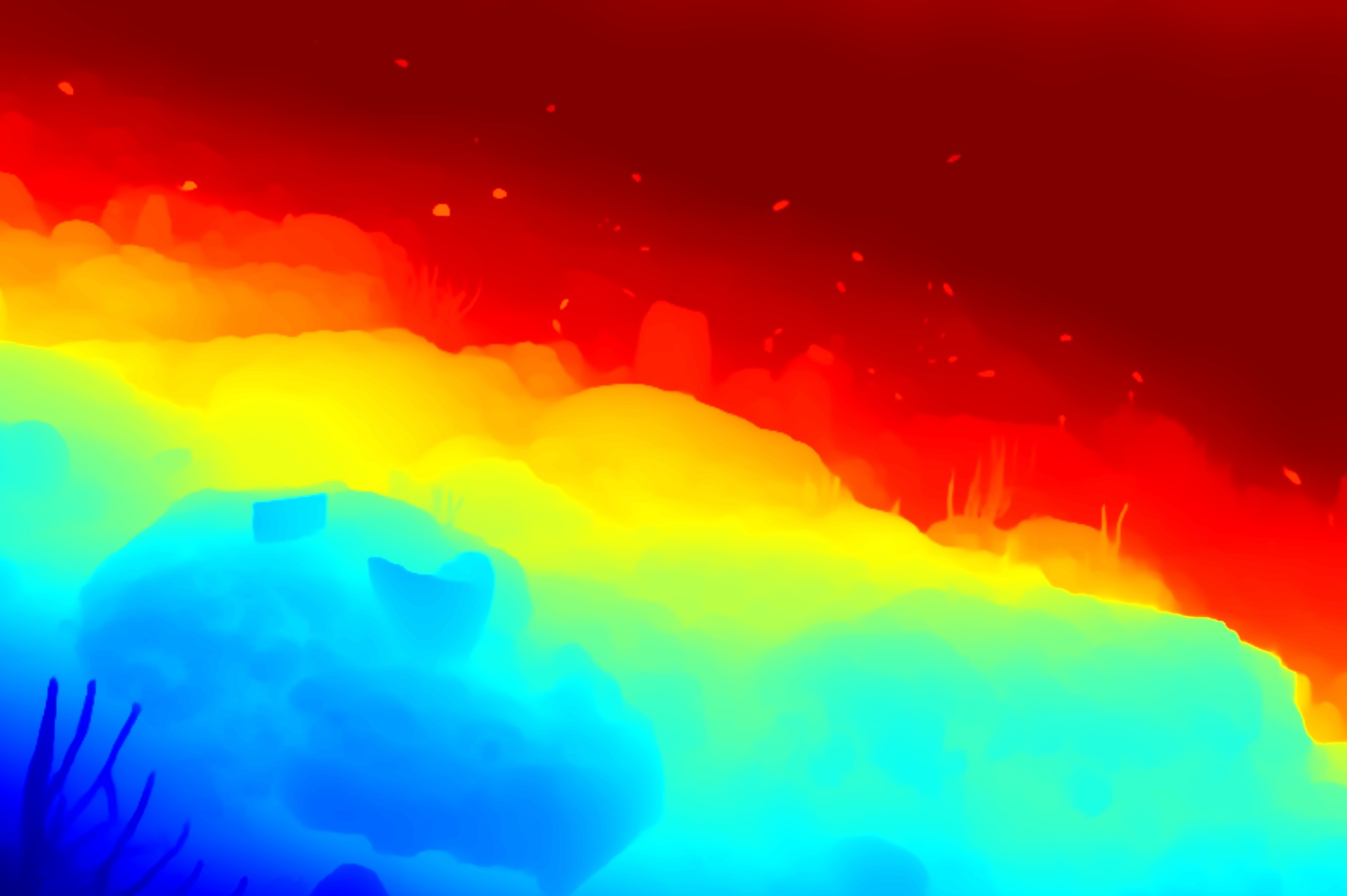}                      &
        \includegraphics[width=\sfs]{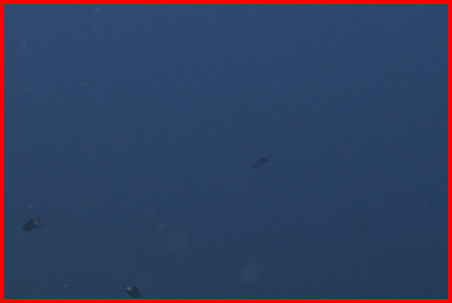}                          \\

        \tc{\includegraphics[width=\fs]{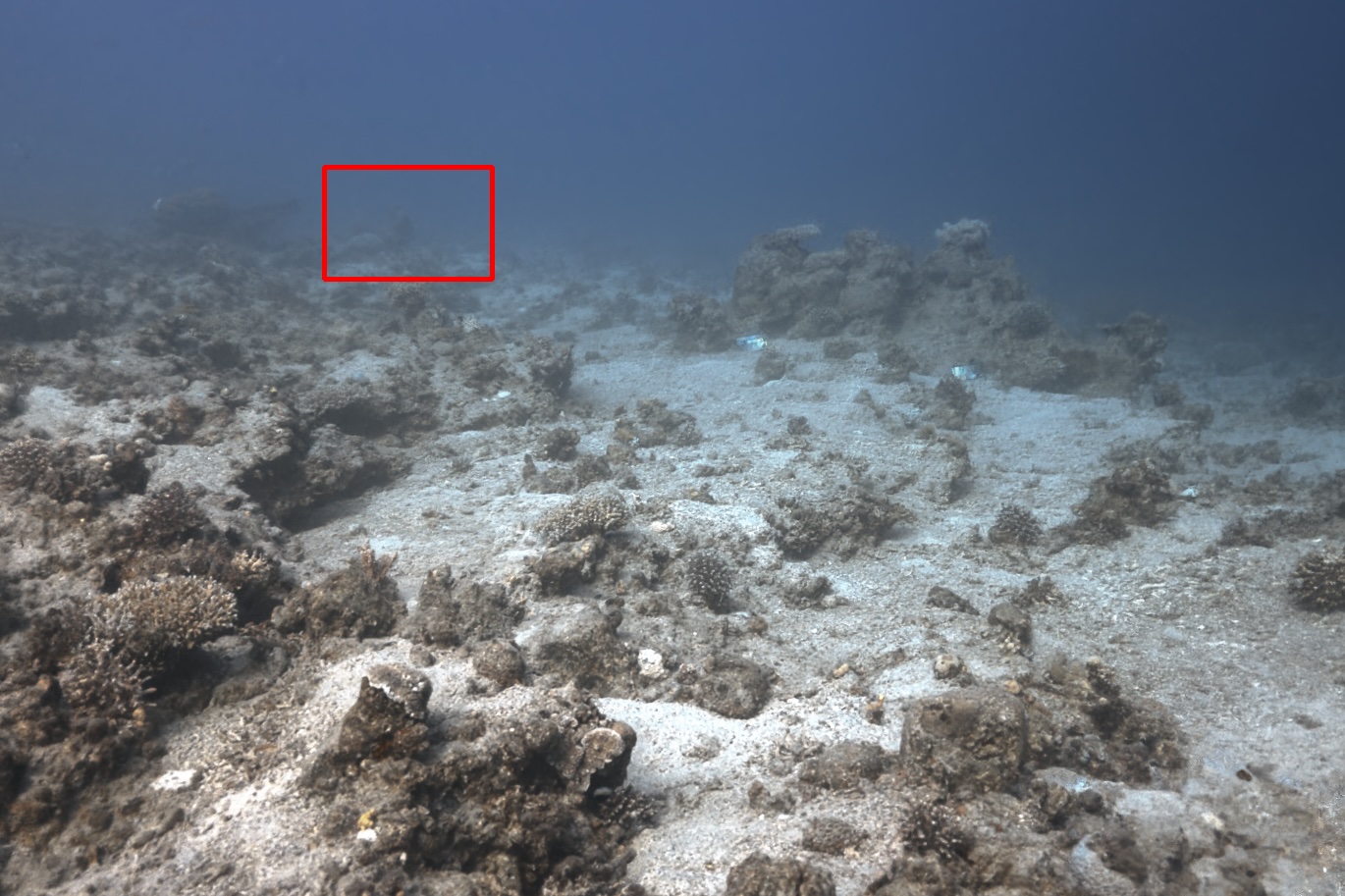}}          &
        \tc{\includegraphics[width=\fs]{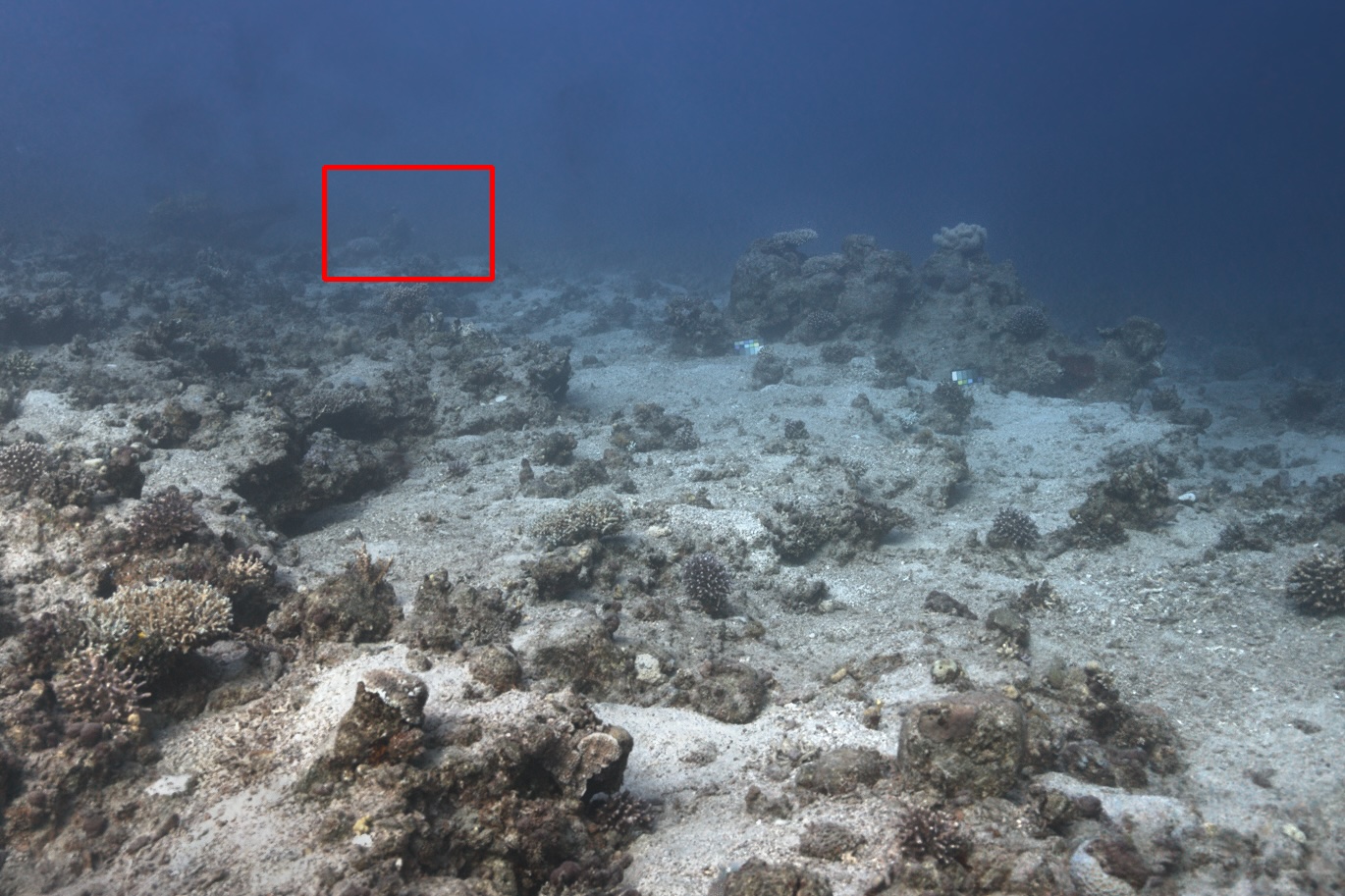}}      &
        \tc{\includegraphics[width=\fs]{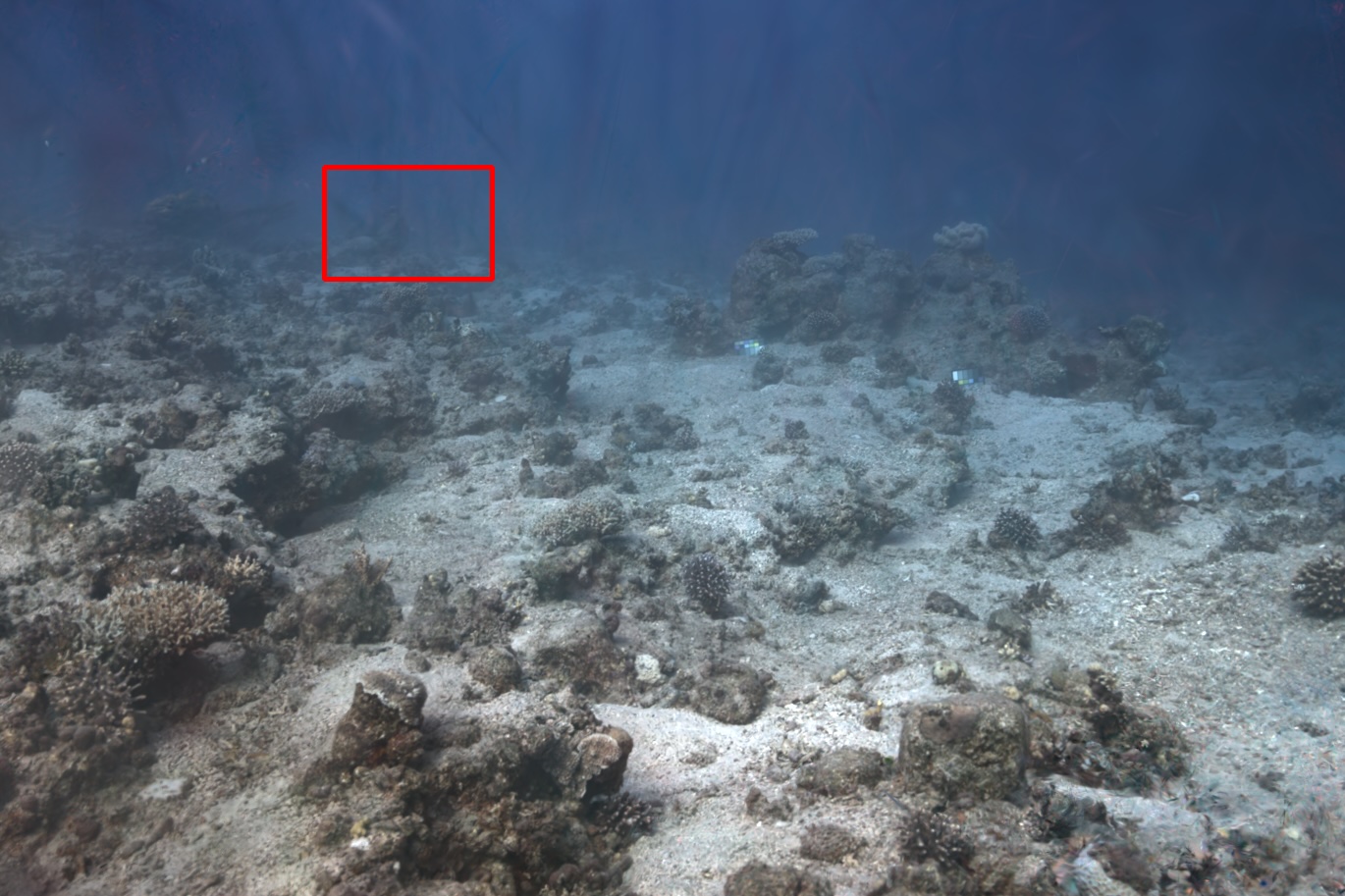}}              &
        \tc{\includegraphics[width=\fs]{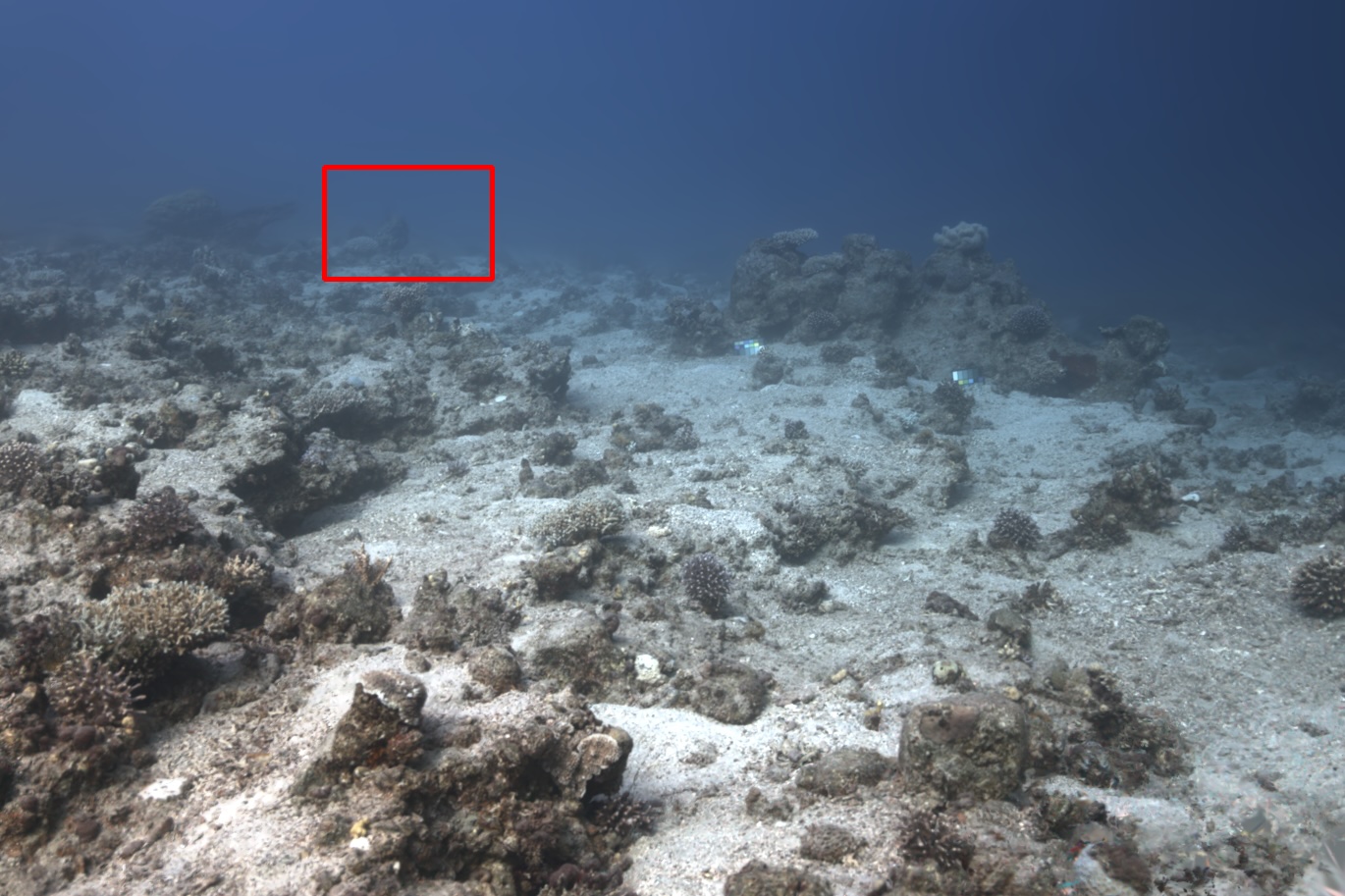}}   &
        \tc{\includegraphics[width=\fs]{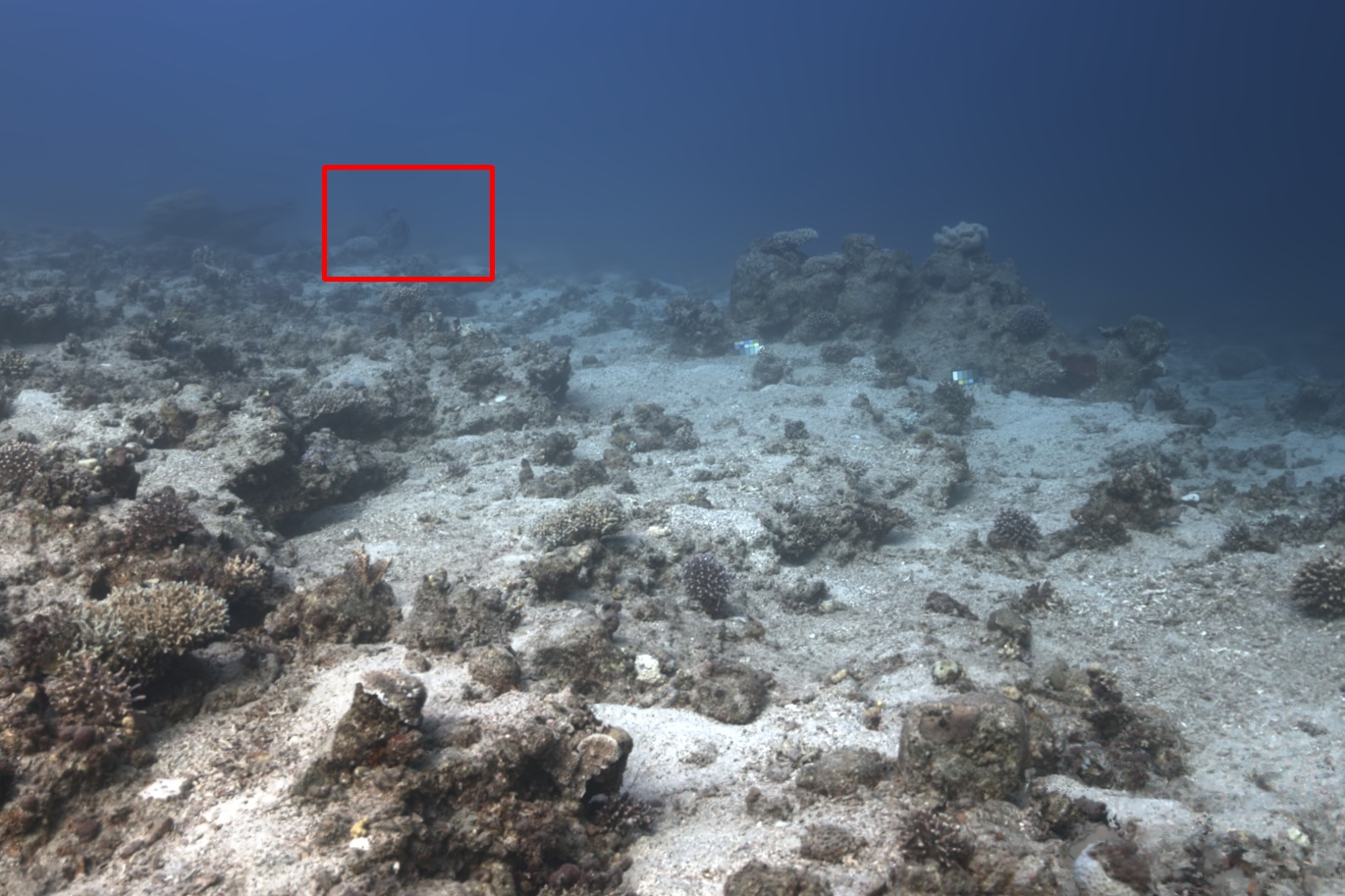}} &
        \tc{\includegraphics[width=\fs]{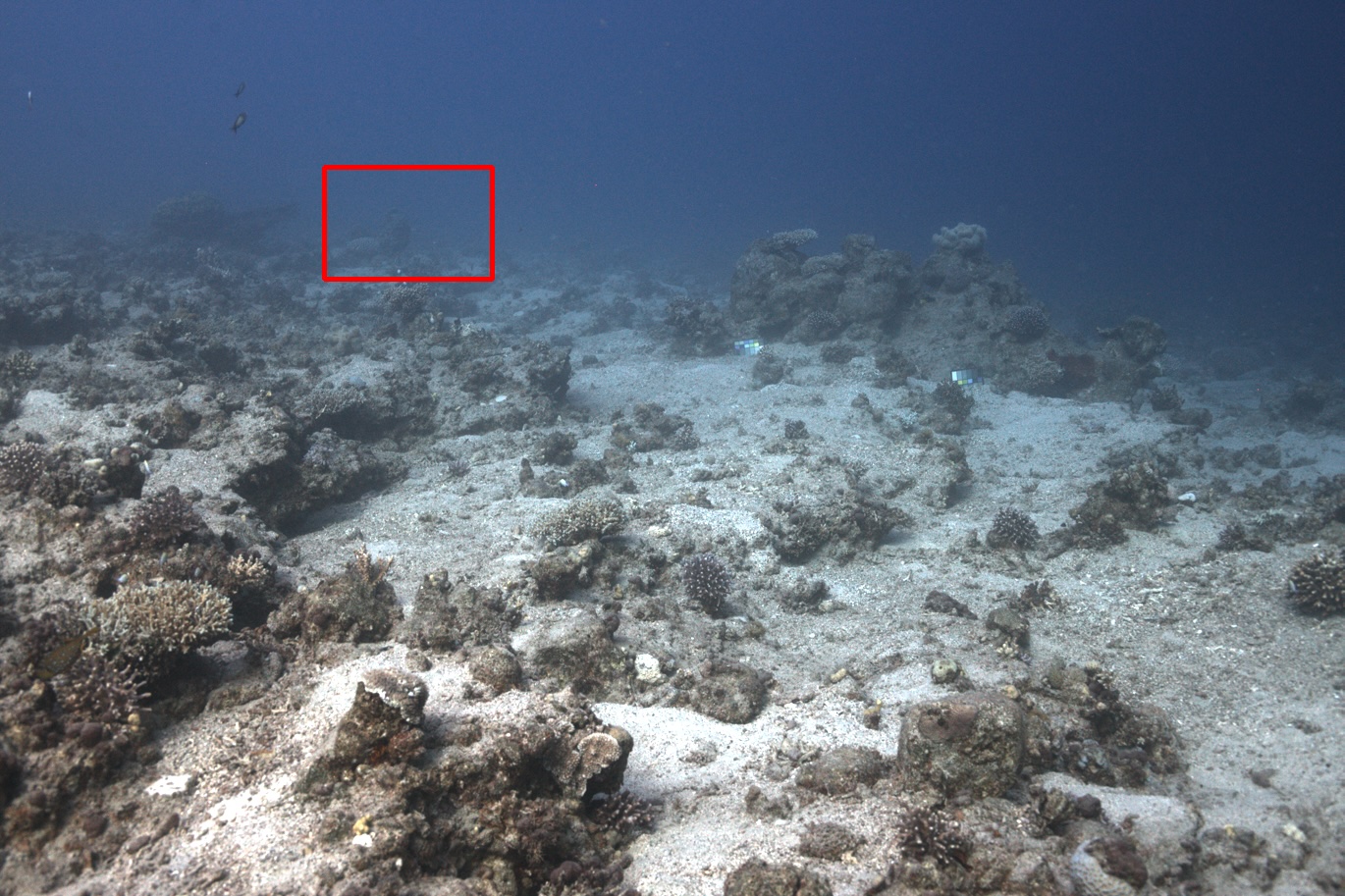}}                  \\
        \includegraphics[width=\sfs]{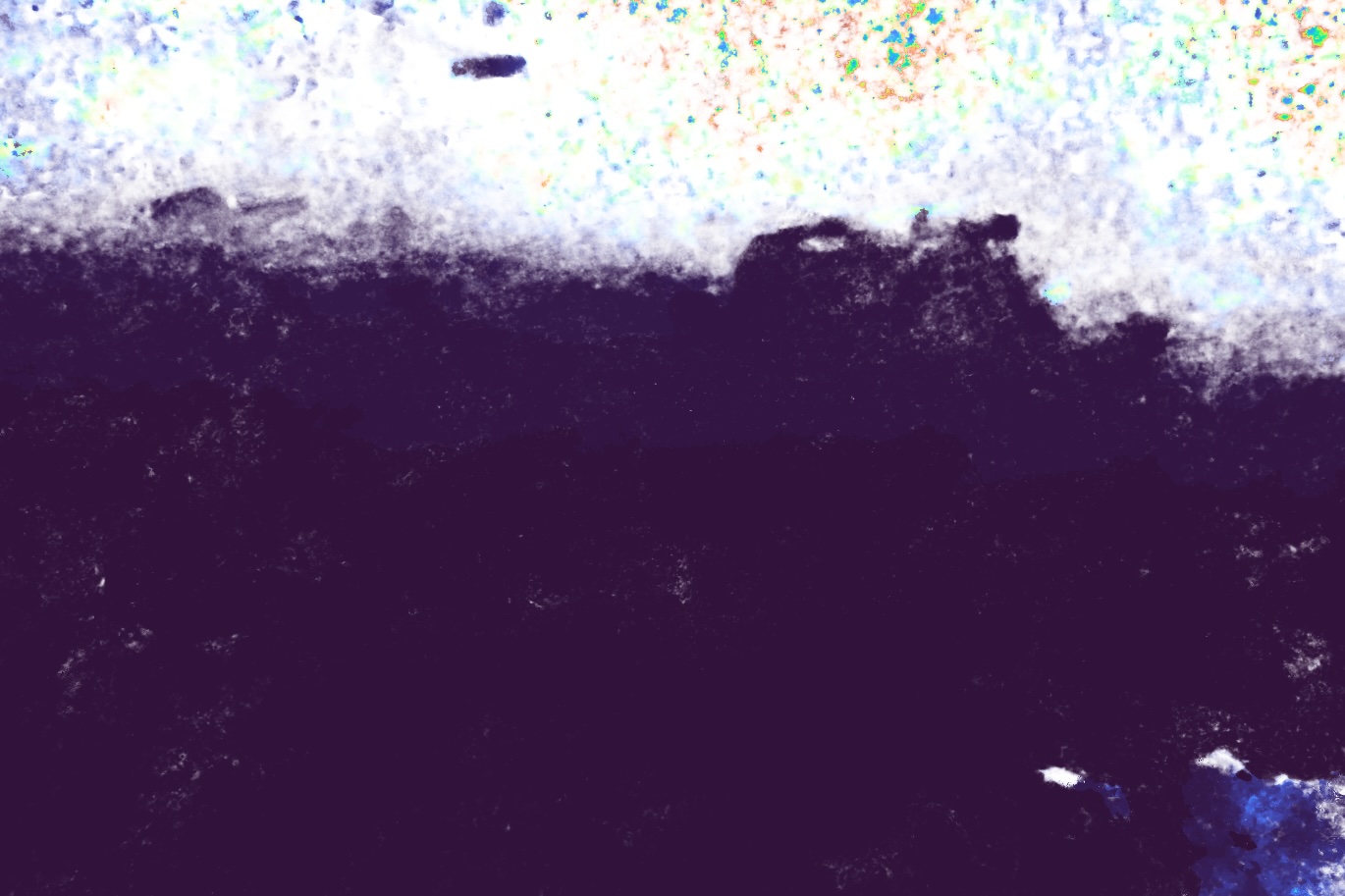}             &
        \includegraphics[width=\sfs]{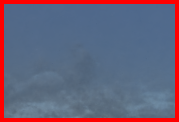}              &
        \includegraphics[width=\sfs]{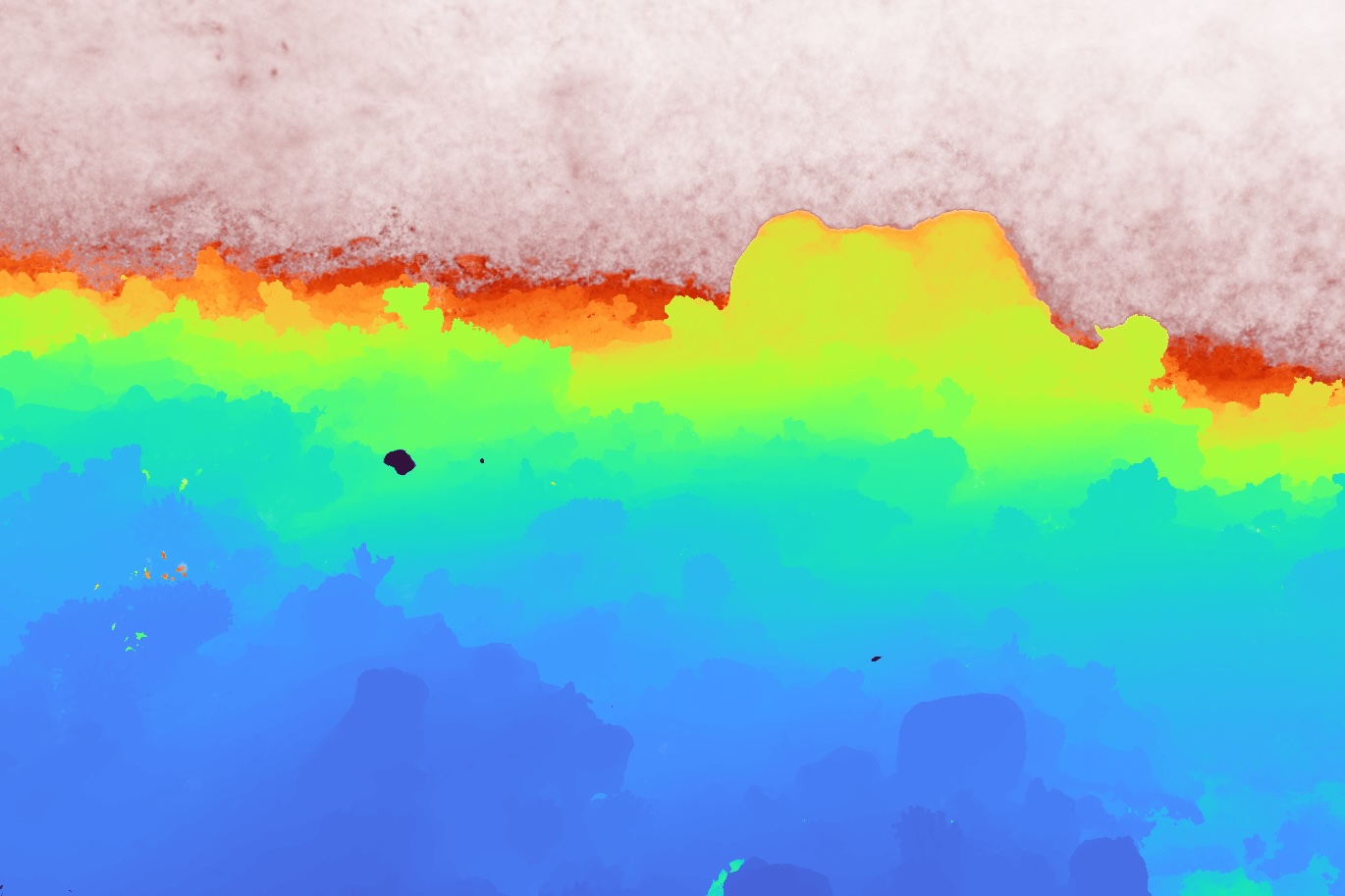}         &
        \includegraphics[width=\sfs]{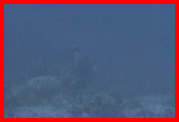}          &
        \includegraphics[width=\sfs]{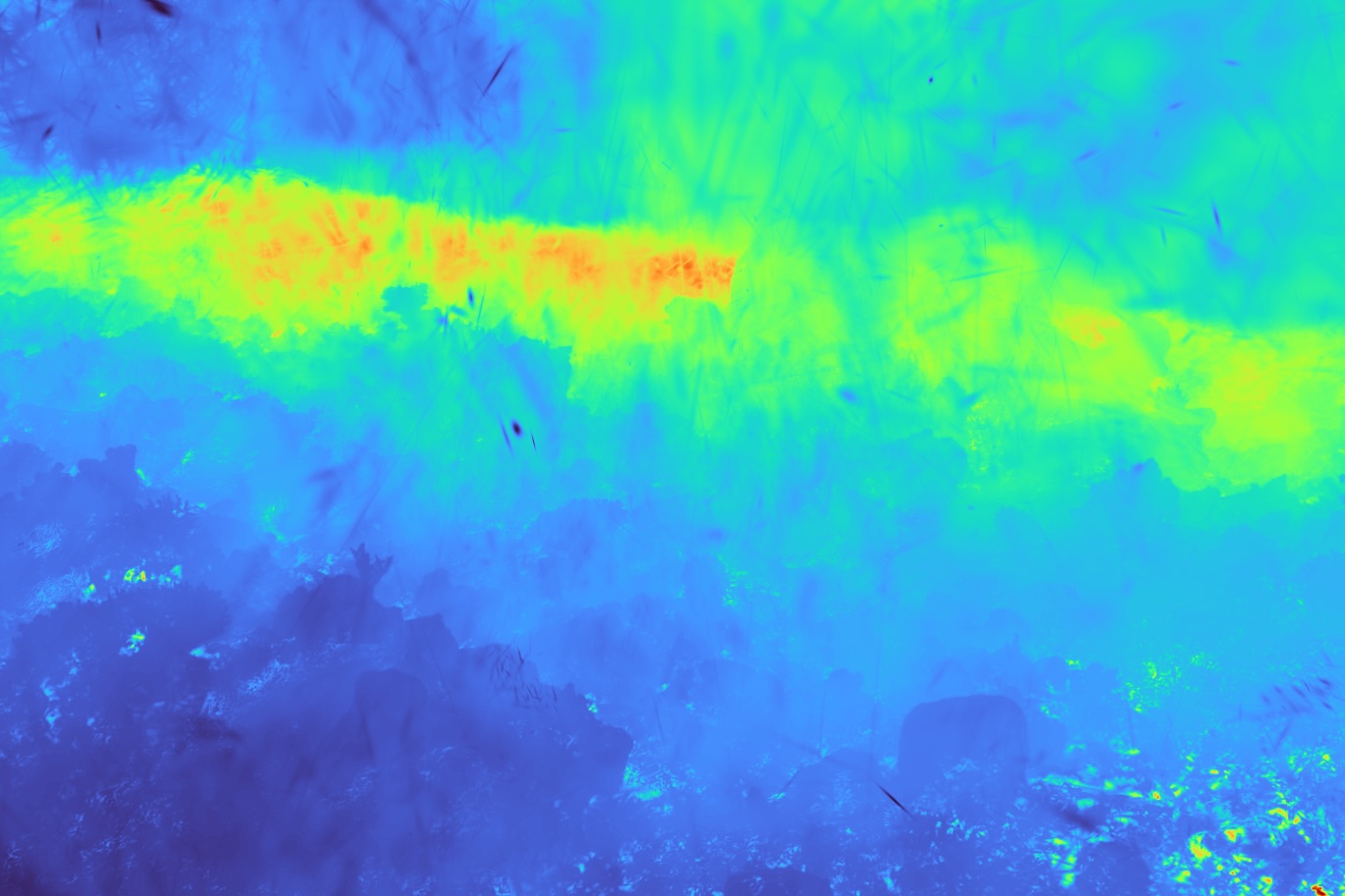}                 &
        \includegraphics[width=\sfs]{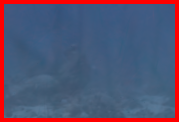}                  &
        \includegraphics[width=\sfs]{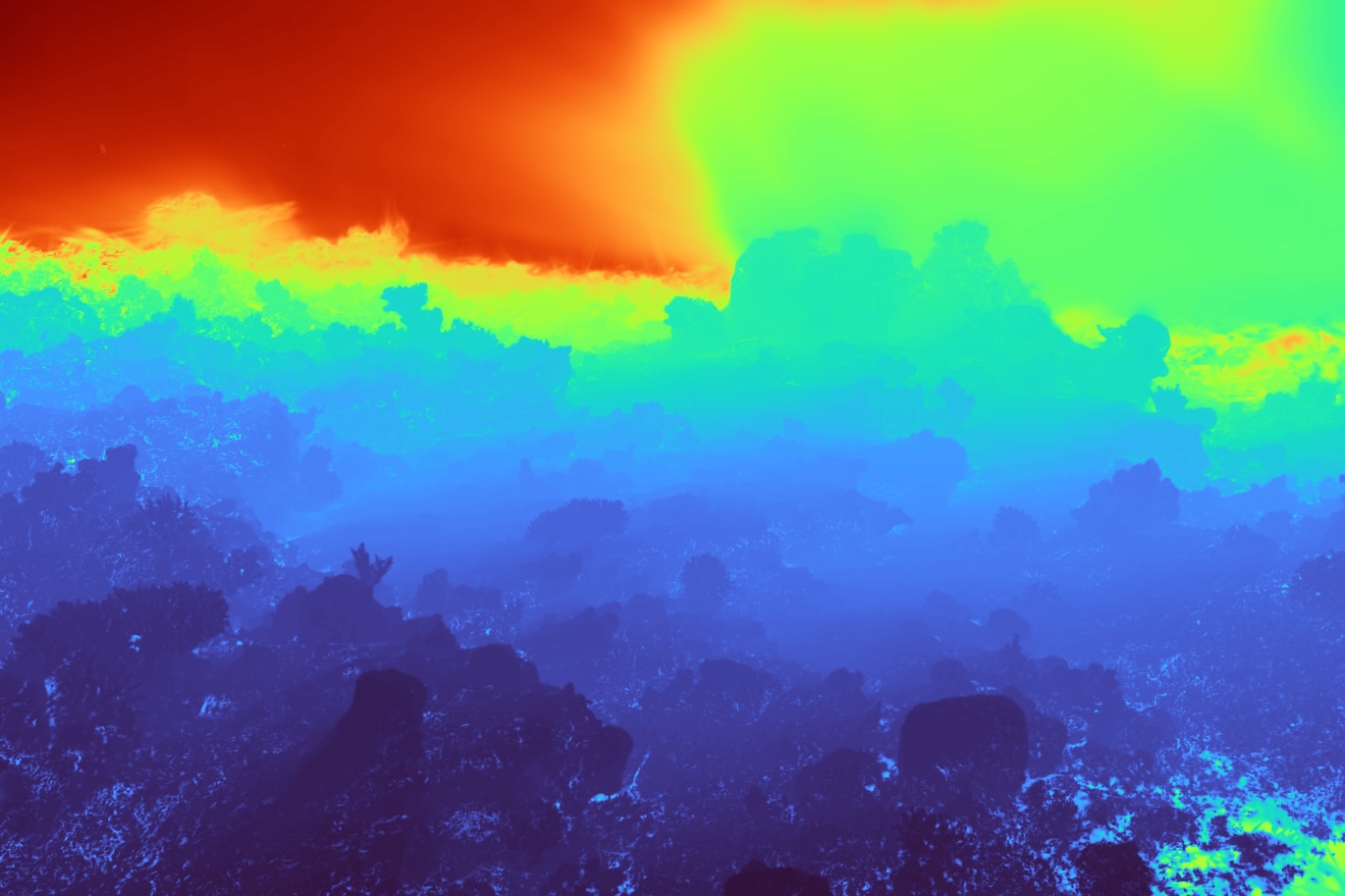}      &
        \includegraphics[width=\sfs]{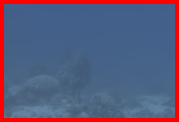}       &
        \includegraphics[width=\sfs]{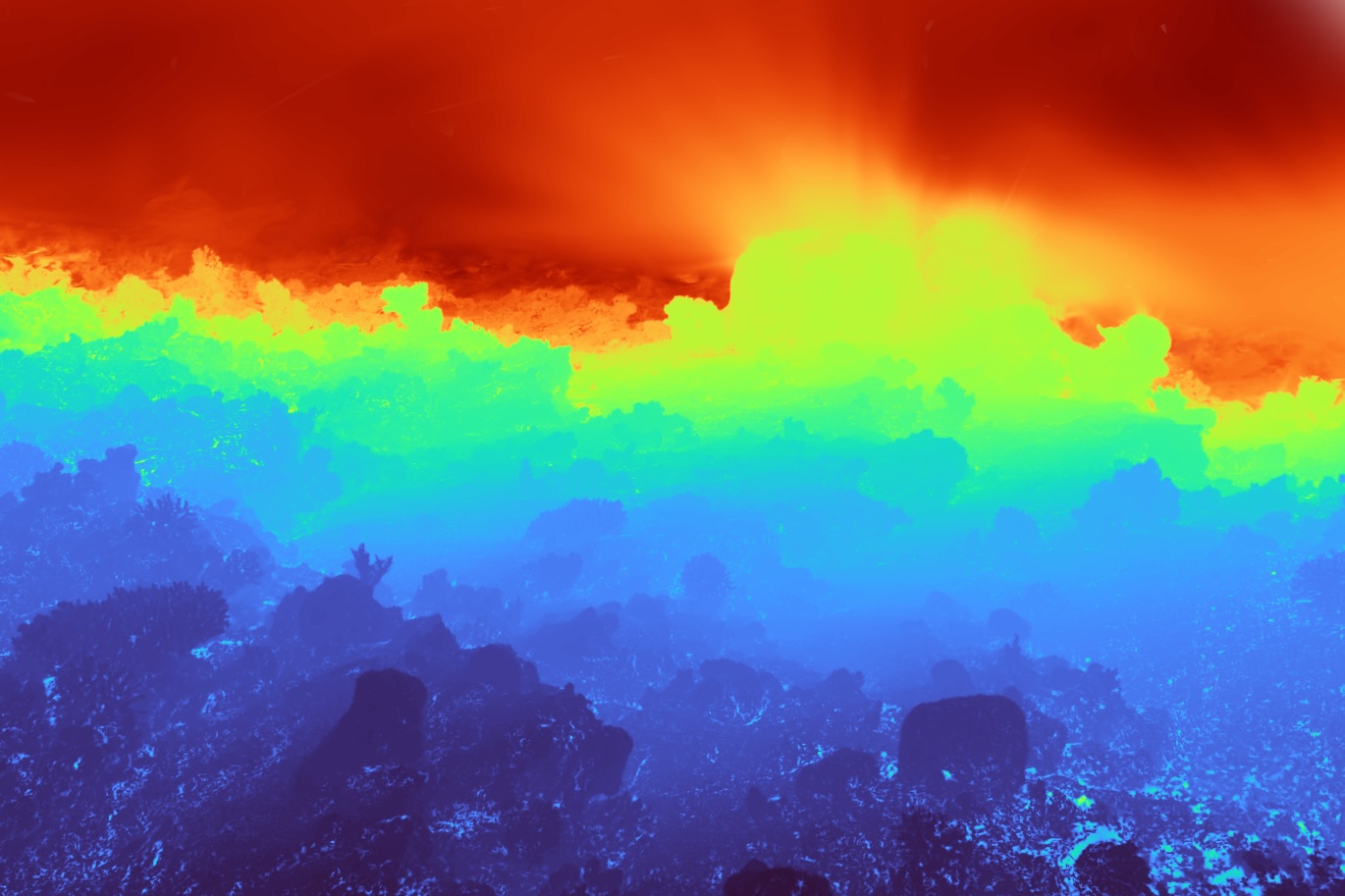}    &
        \includegraphics[width=\sfs]{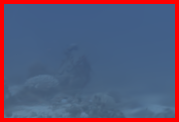}     &
        \includegraphics[width=\sfs]{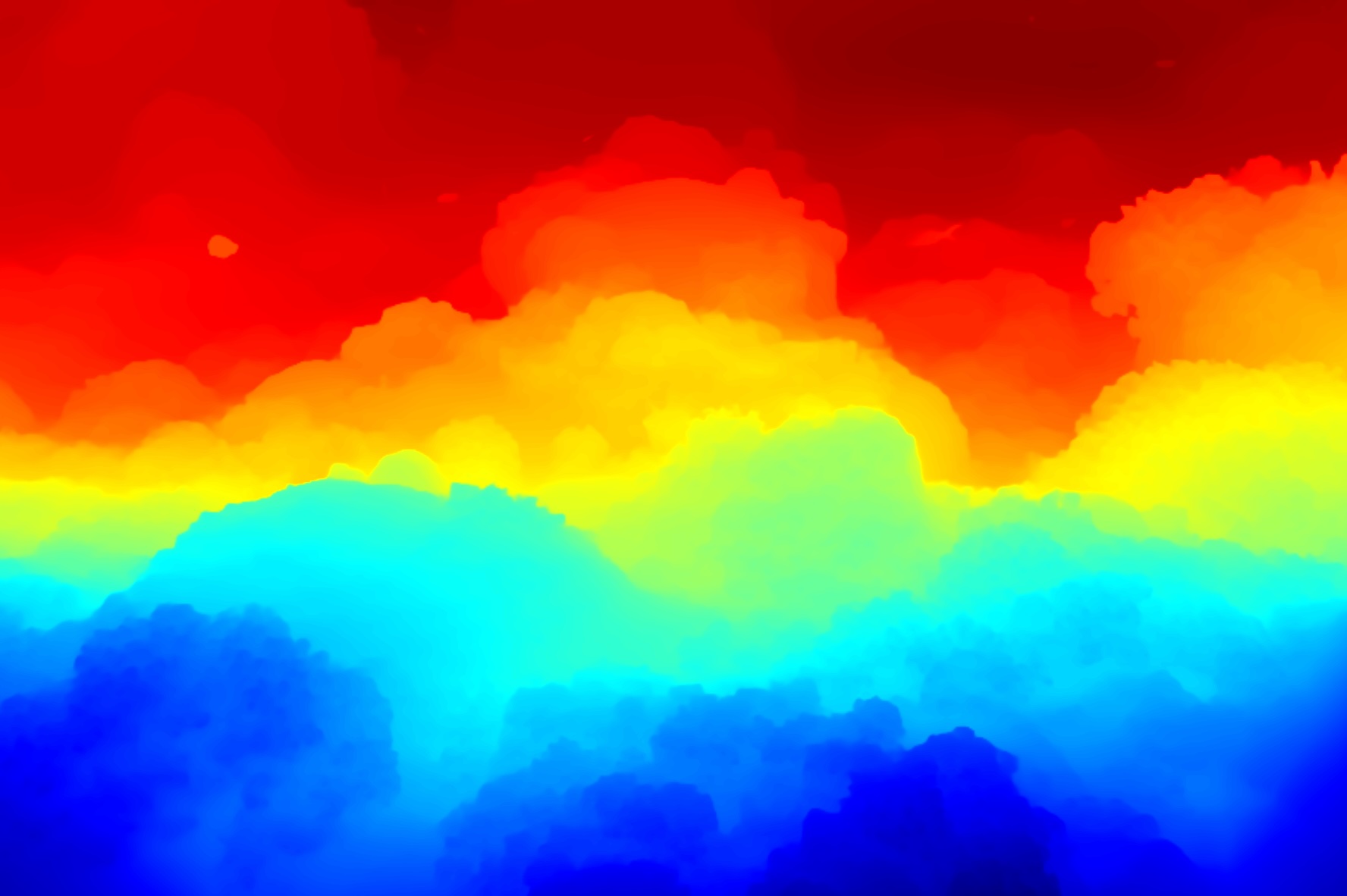}                  &
        \includegraphics[width=\sfs]{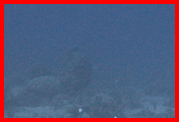}                      \\
        \tc{Zip-NeRF}                                                                         &
        \tc{SeaThru-NeRF}                                                                     &
        \tc{3DGS}                                                                             &
        \tc{WaterSplatting}                                                                   &
        \tc{Plenodium}                                                                       &
        \tc{GT}                                                                             \\
    \end{tabular}
    \vspace{-2mm}
    \caption{{Rendering performance comparison of our Plenodium against existing methods on the ``Cura\c{c}ao'' and ``IUI3 Red Sea'' scenes.} Rendered images and depths are presented for comparison. The pseudo-depth for ground truth is estimated using the Depth Any Model~\cite{Depthanything,Depthanythingv2} for reference purposes.
    Compared to competing methods, Plenodium enhances clarity for medium and distant objects, as highlighted in the red boxes, while simultaneously yielding more precise depth maps.
    }
    \label{fig: rec}
    \vspace{-2mm}
\end{figure*}
\begin{figure*}[t]
    \centering
    \newcommand{\tc}[1]{\multicolumn{2}{@{}c}{#1}} 
    \setlength{\tabcolsep}{0.5pt}
    \small
    \def\fs{0.141\linewidth} 
    \begin{tabular}{@{}ccccccc@{}}
        Reconstruction&
        Restoration&
        Reconstruction&
        Restoration&
        Reconstruction&
        Restoration&
        \\
        \includegraphics[width=\fs]{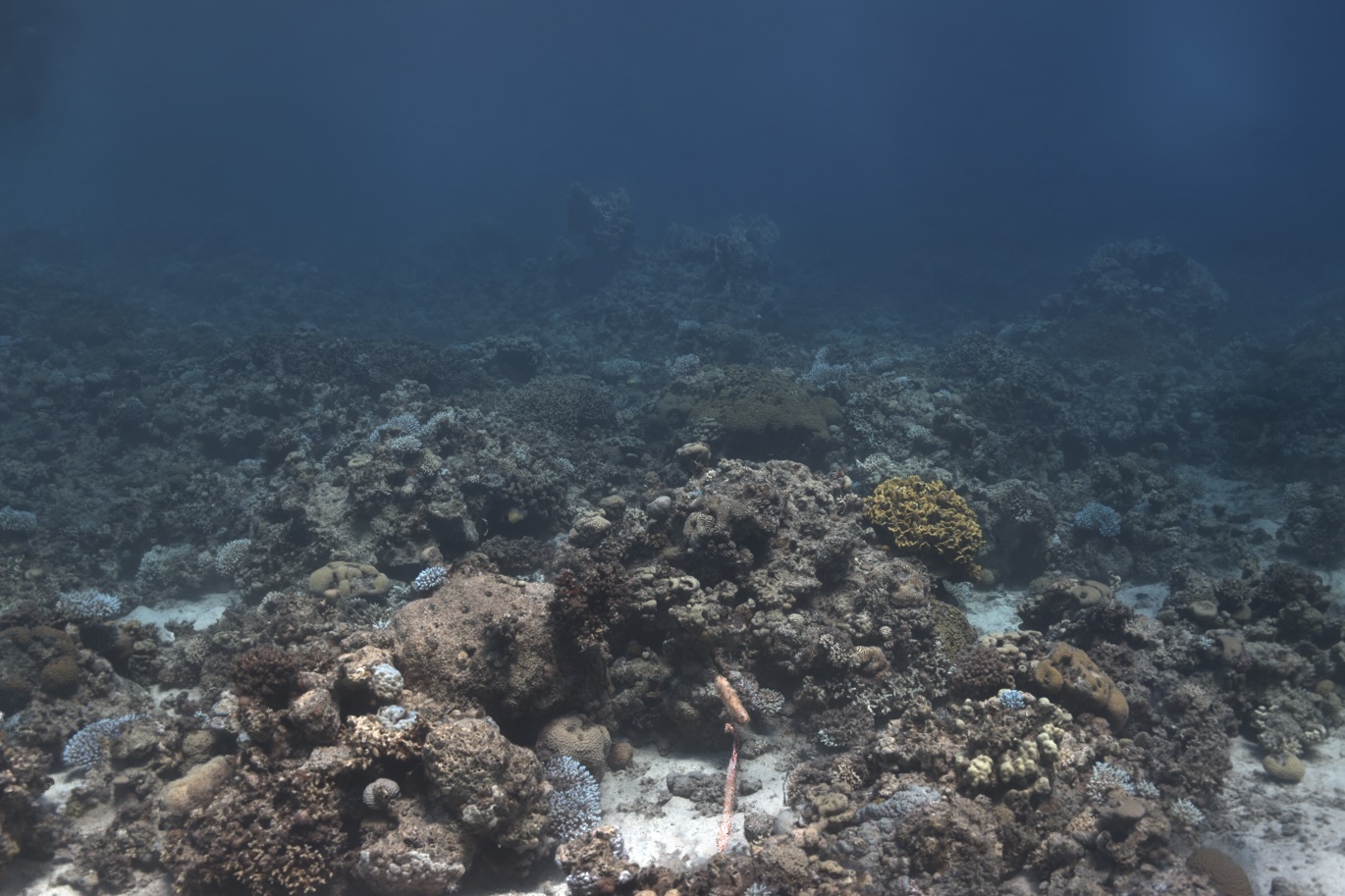}        &
        \includegraphics[width=\fs]{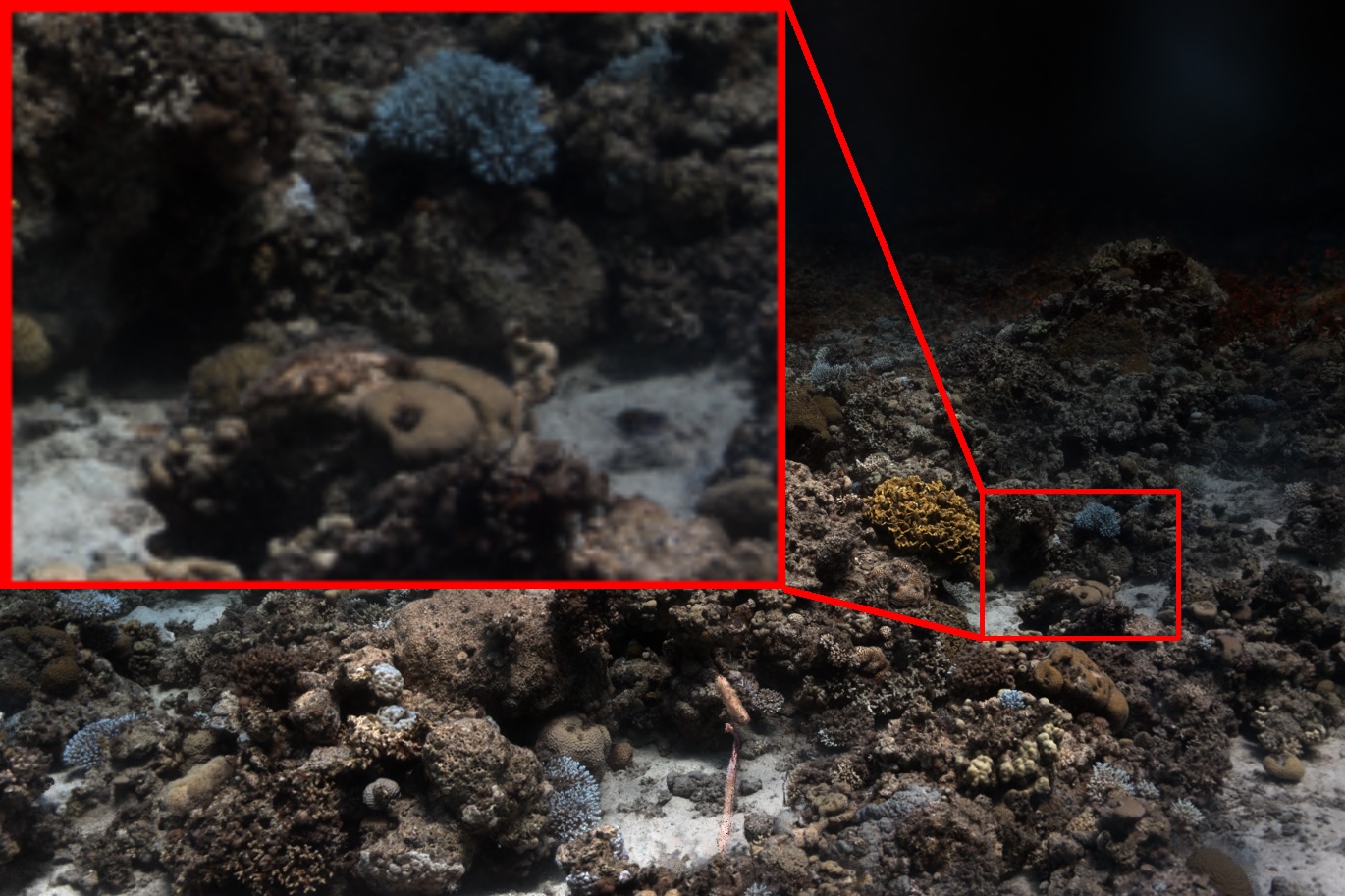}      &
        \includegraphics[width=\fs]{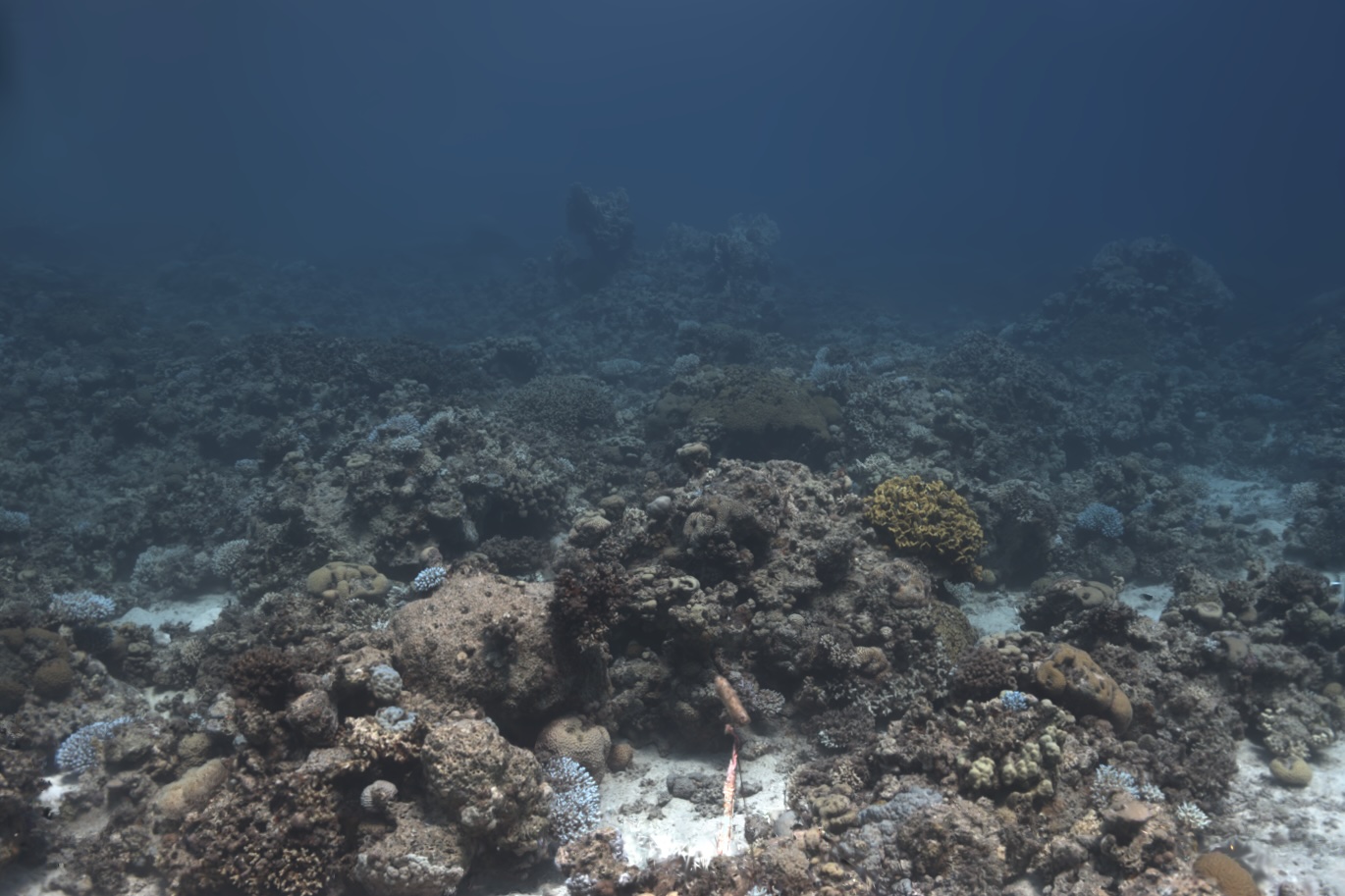}     &
        \includegraphics[width=\fs]{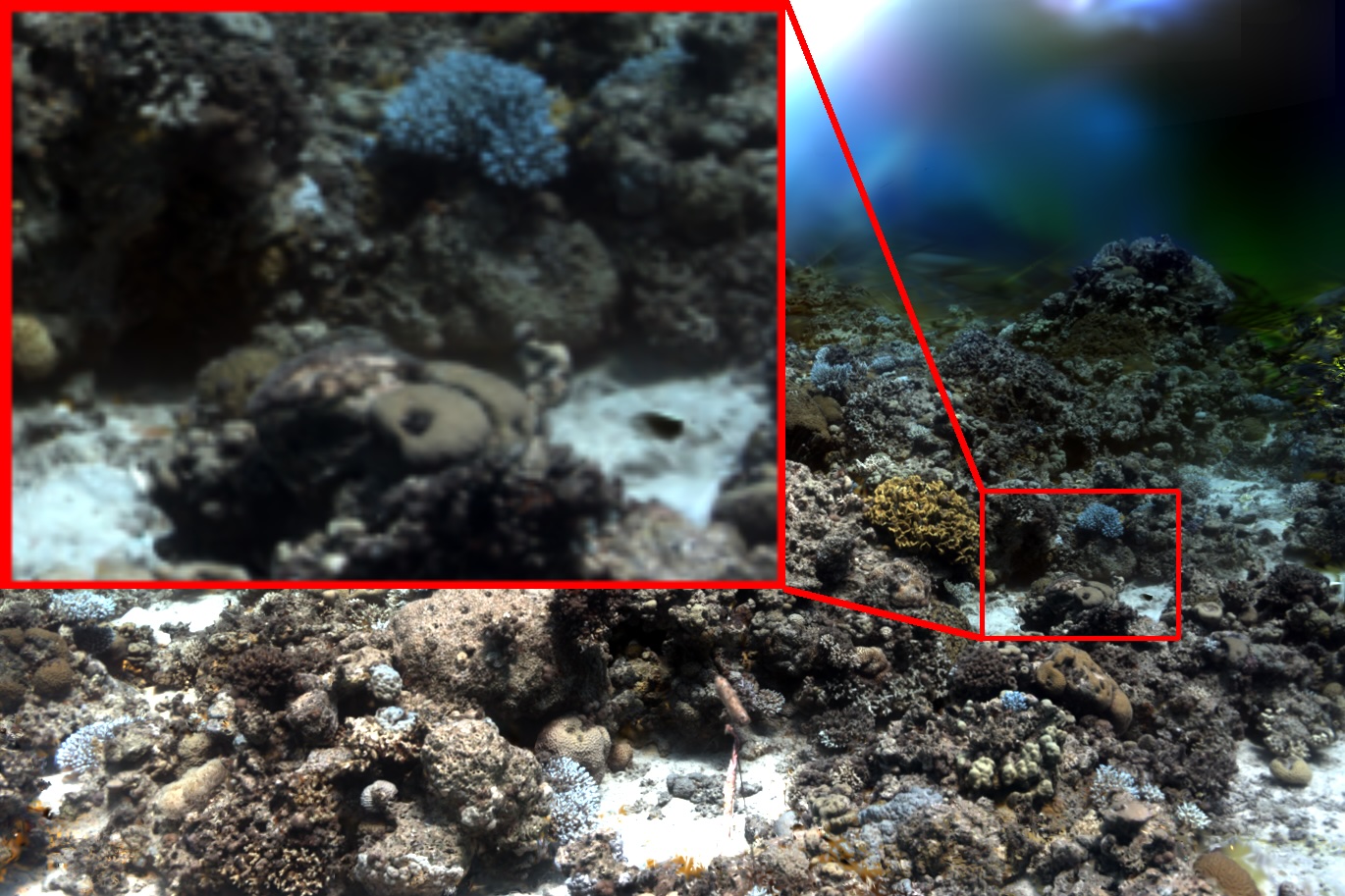}   &
        \includegraphics[width=\fs]{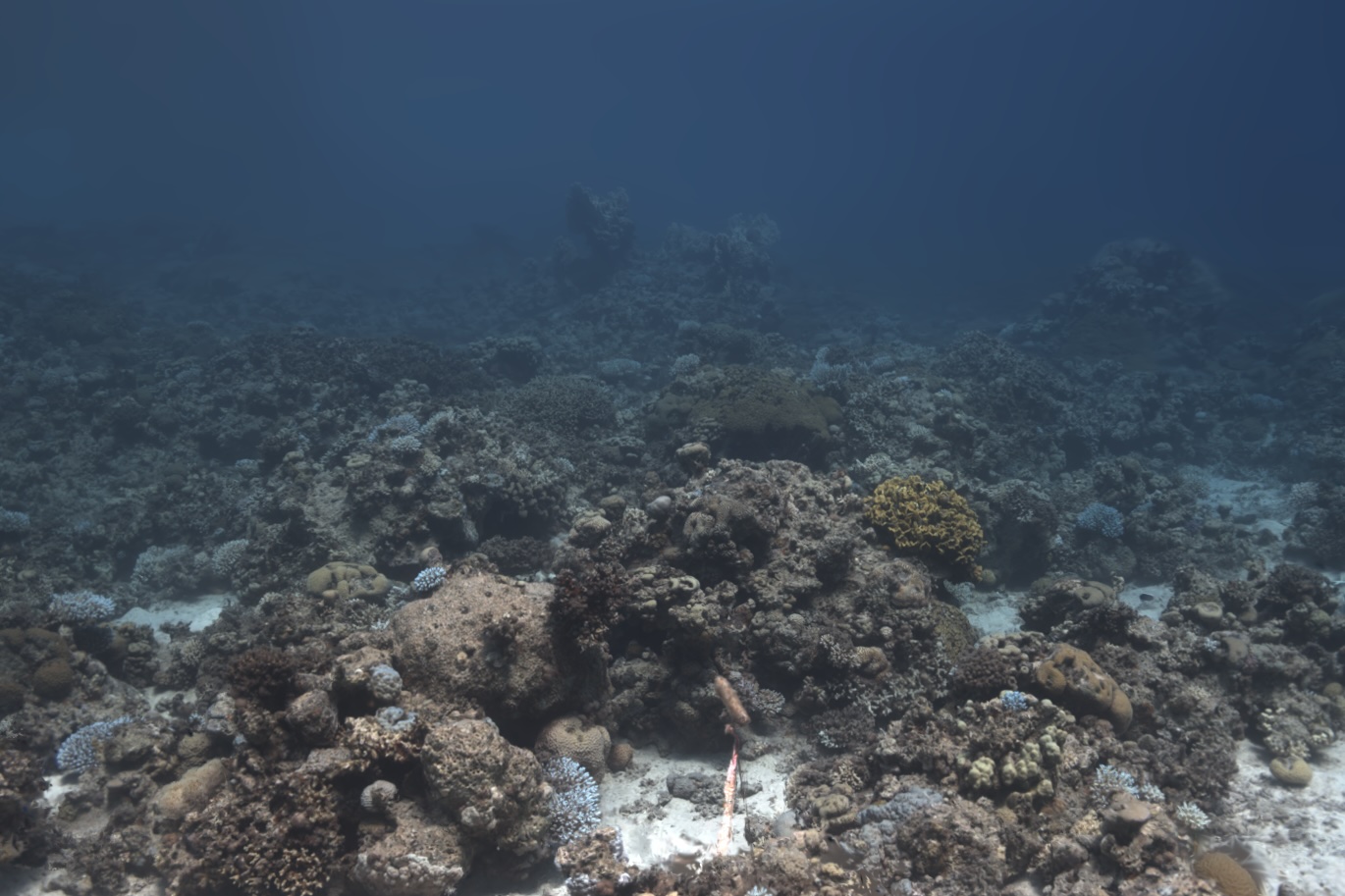}   &
        \includegraphics[width=\fs]{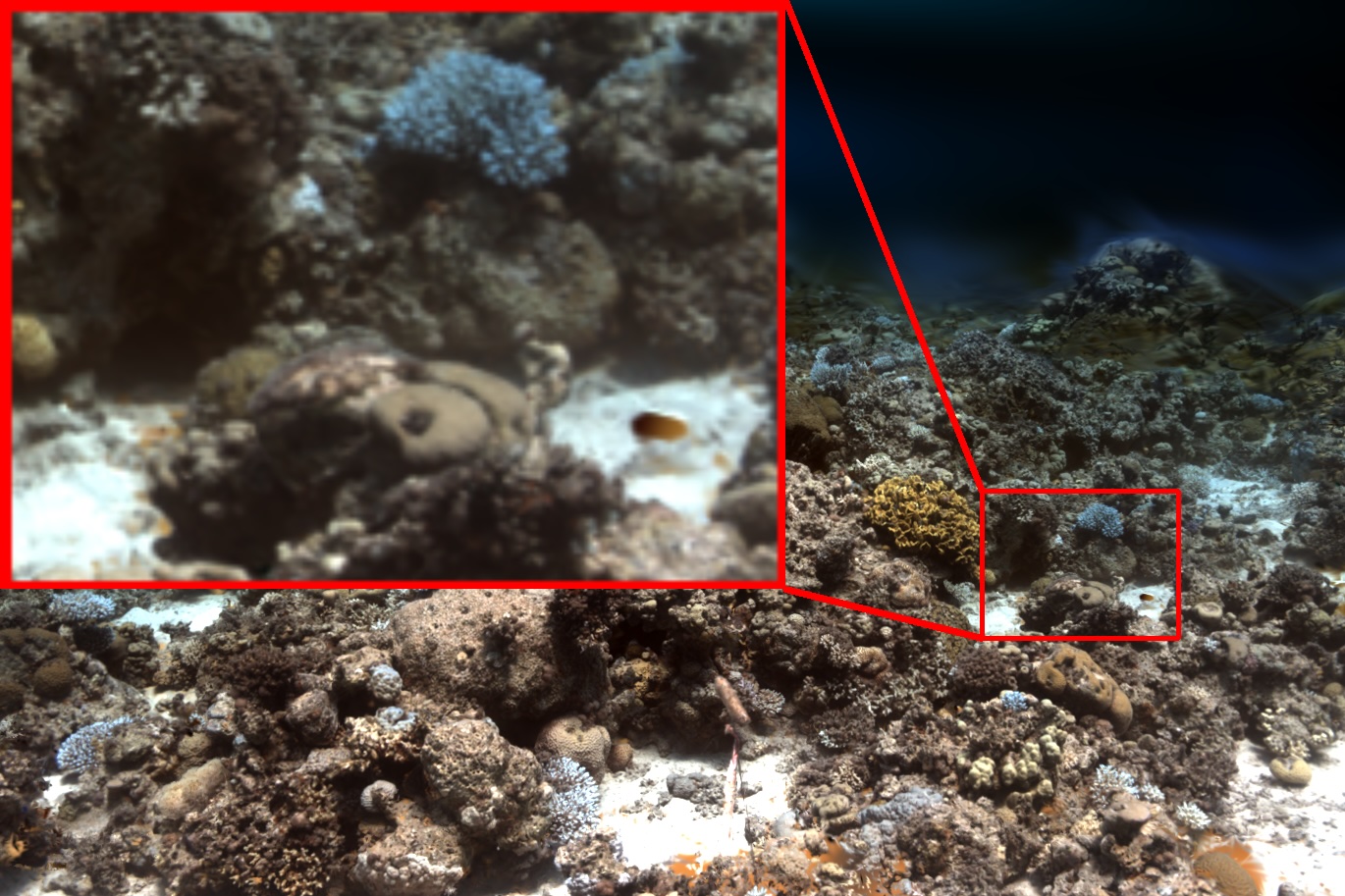} &
        \includegraphics[width=\fs]{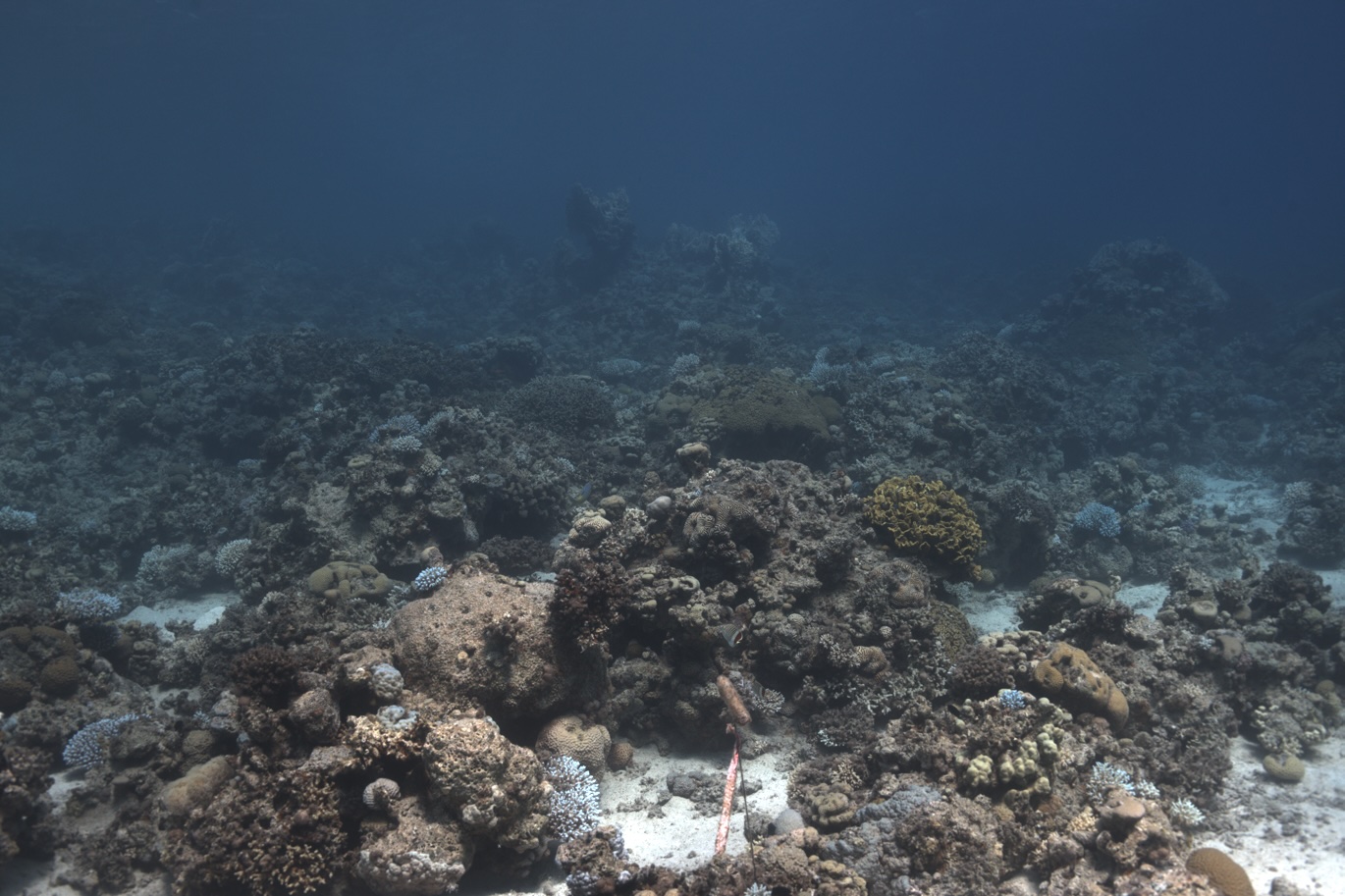}                        \\
        \includegraphics[width=\fs]{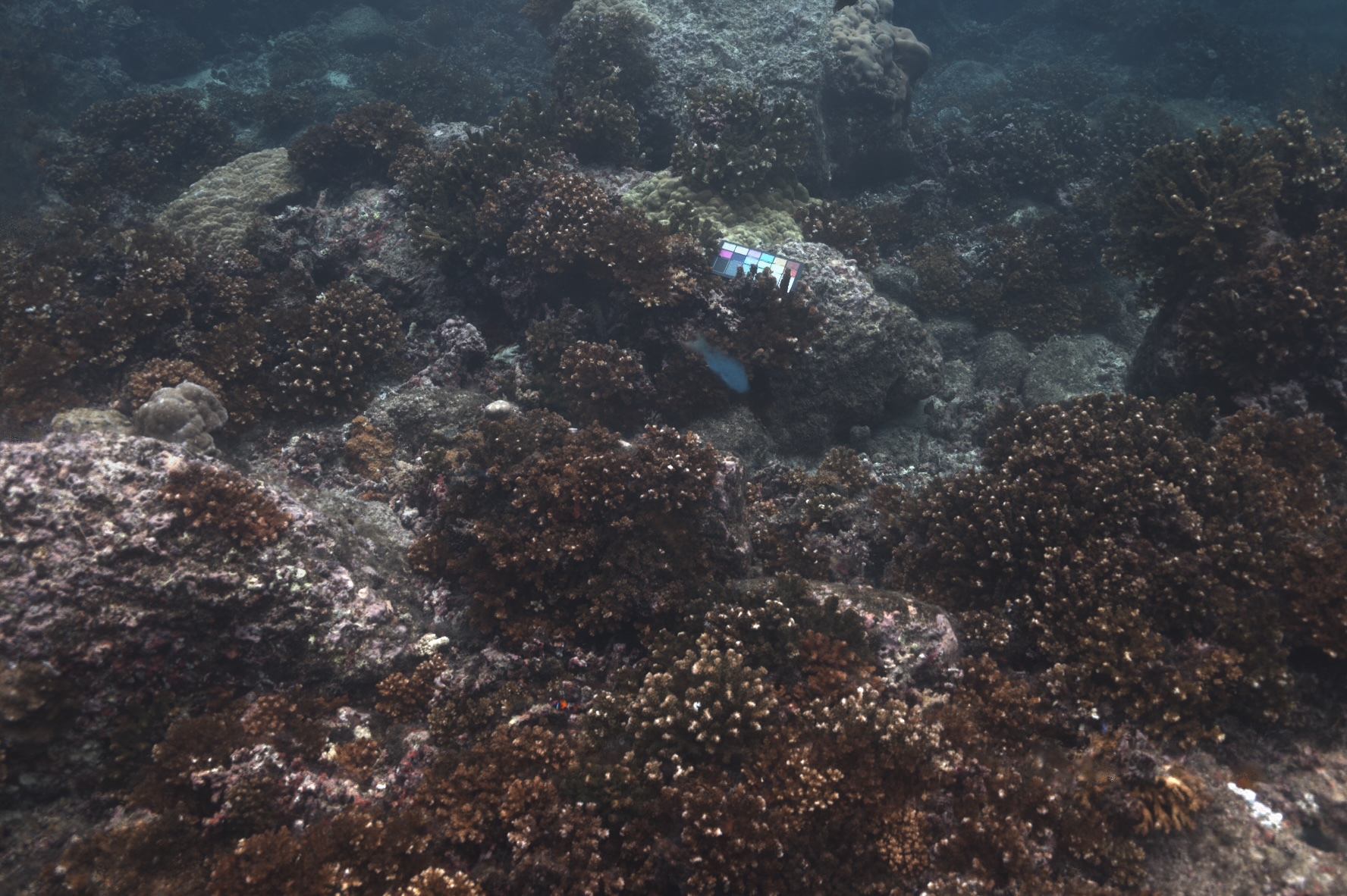}                 &
        \includegraphics[width=\fs]{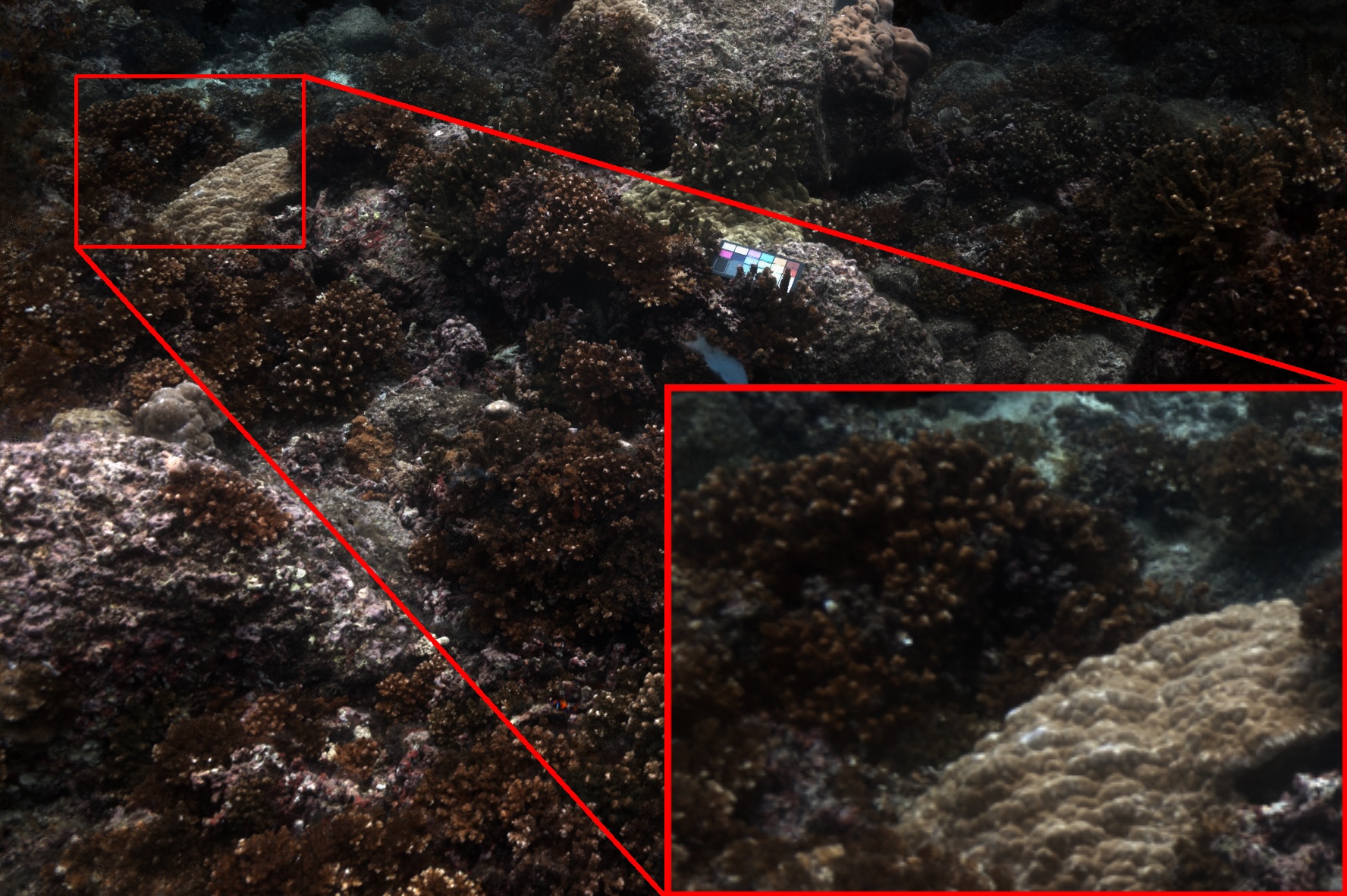}               &
        \includegraphics[width=\fs]{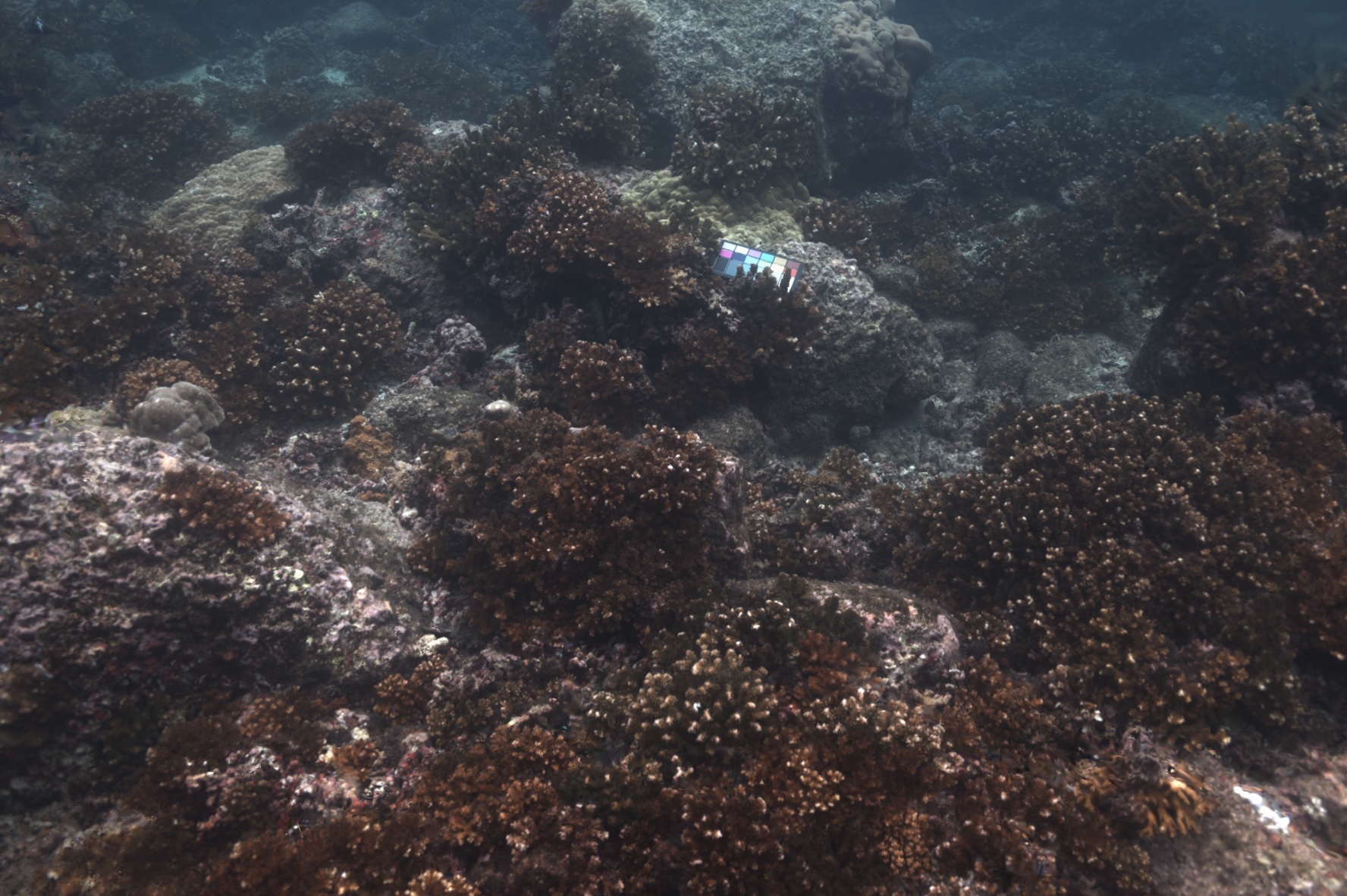}              &
        \includegraphics[width=\fs]{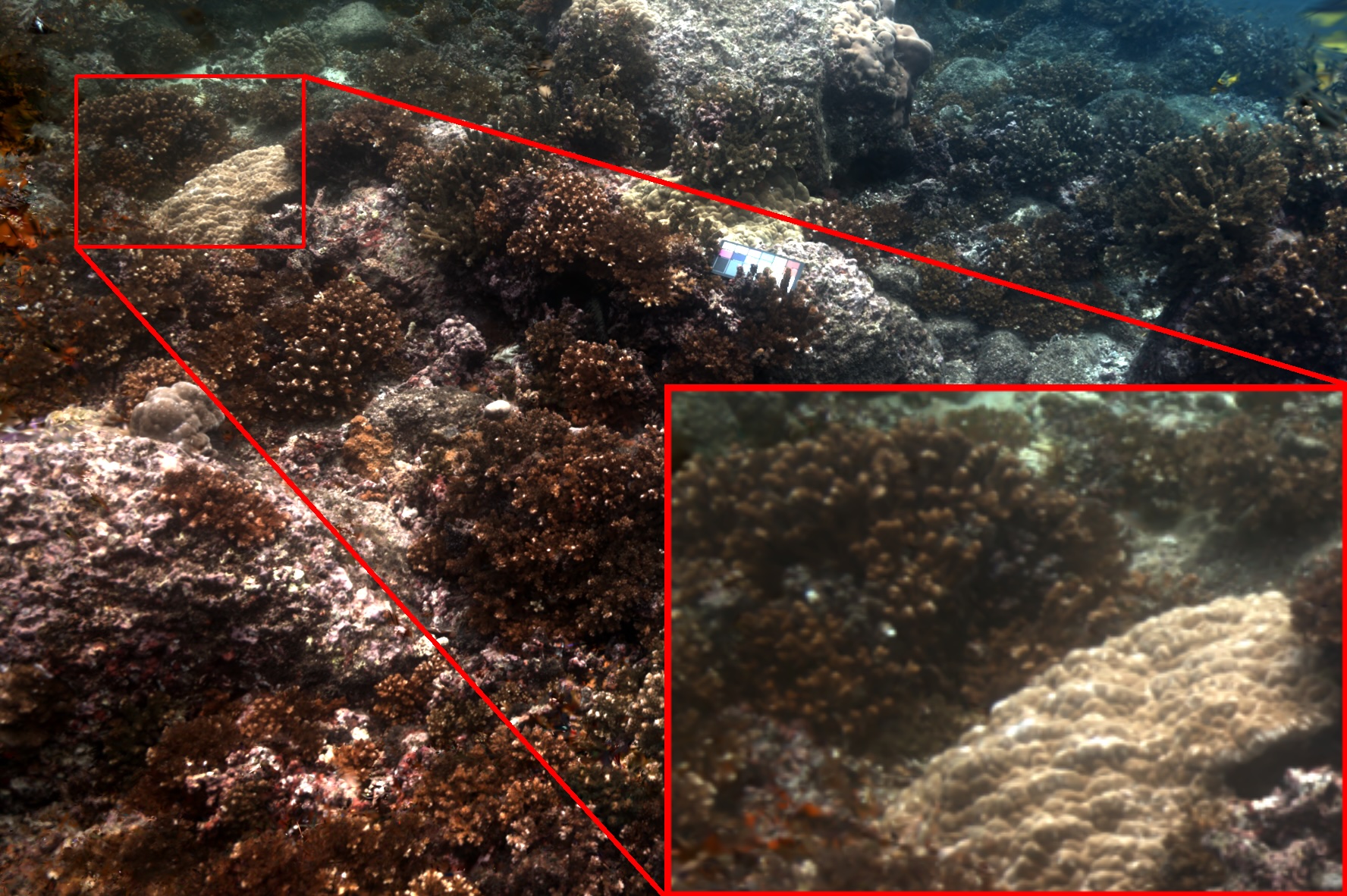}            &
        \includegraphics[width=\fs]{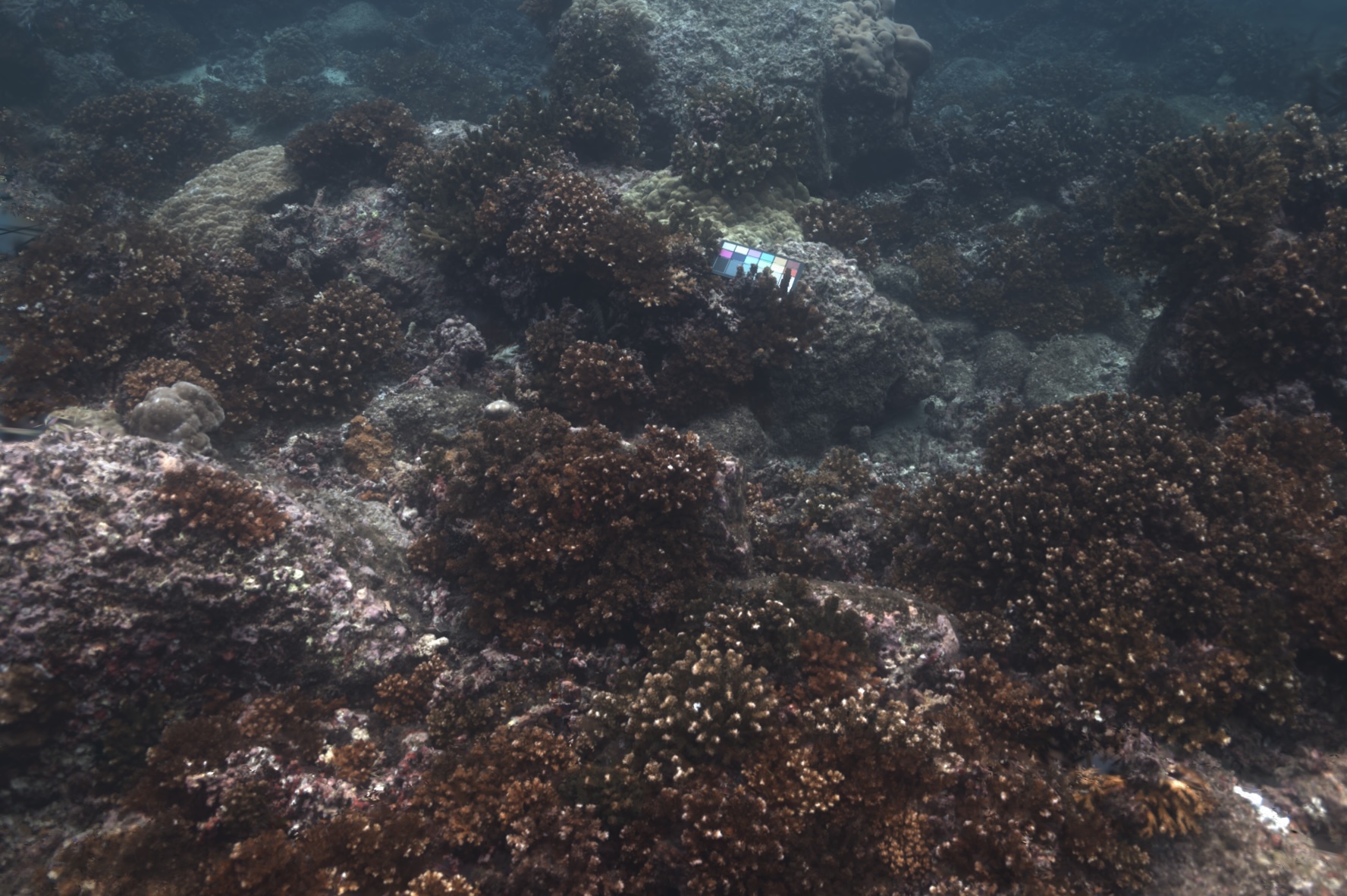}            &
        \includegraphics[width=\fs]{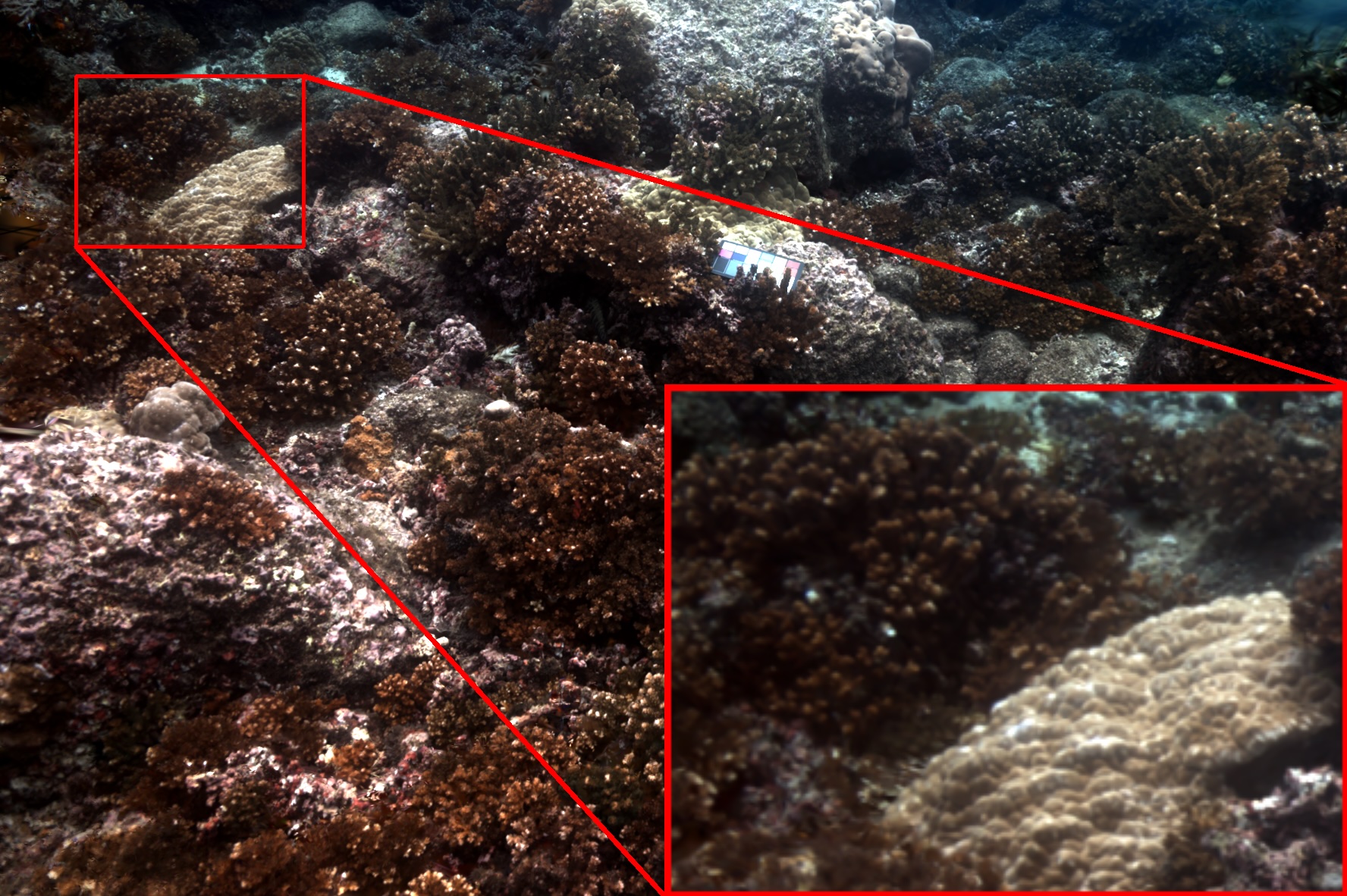}          &
        \includegraphics[width=\fs]{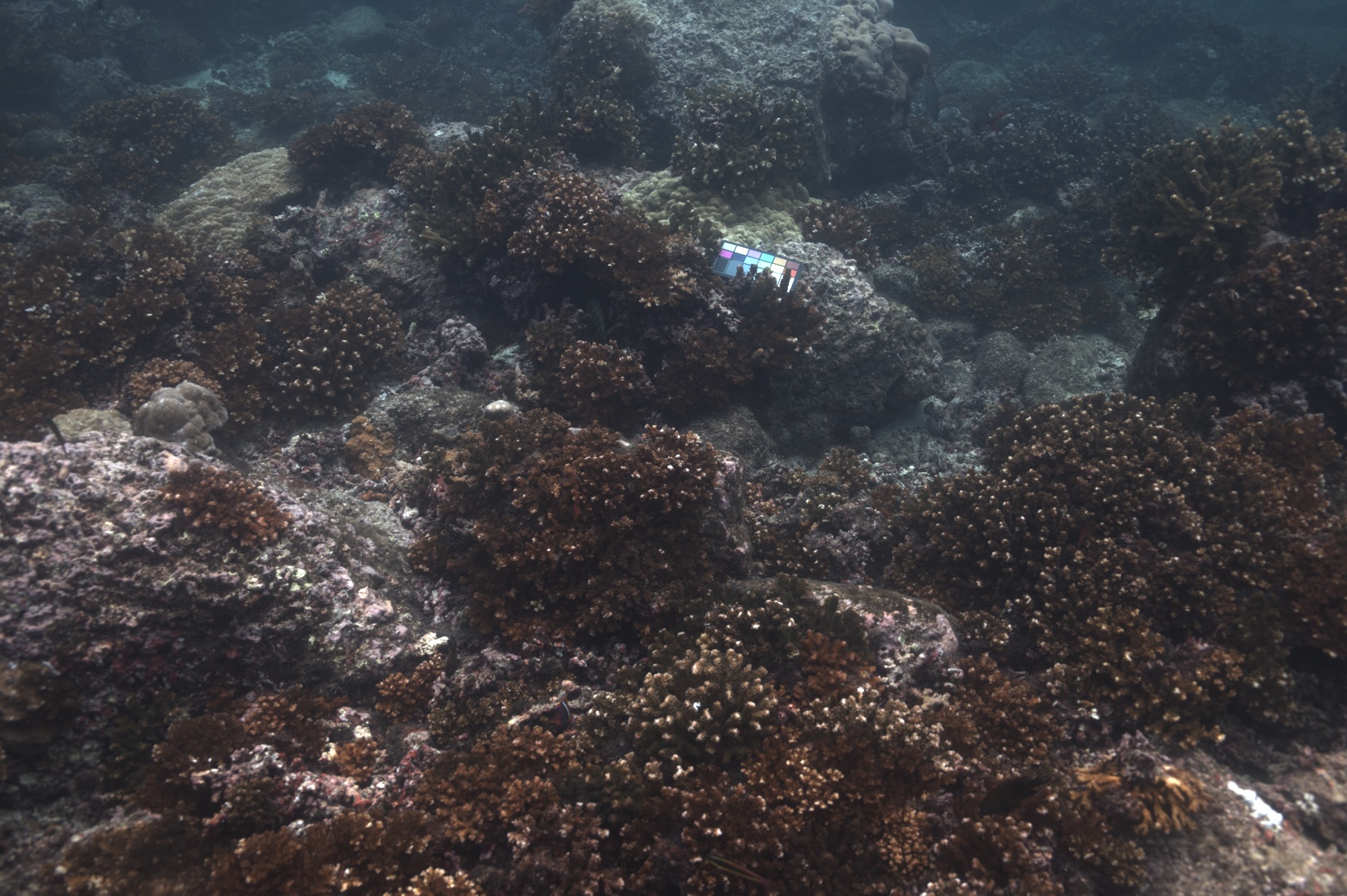}                                 \\
        \multicolumn{2}{@{}c}{SeaThru-NeRF}                               &
        \multicolumn{2}{@{}c}{WaterSplatting}                             &
        \multicolumn{2}{@{}c}{Plenodium}                                       &
        GT                                                                            \\
    \end{tabular}
    \vspace{-2mm}
    \caption{{Restoration performance comparison of our Plenodium against existing methods on the ``JapaneseGarden Red Sea'' and ``Panama'' scenes. As shown in the red boxes, the proposed Plenodium generates results with more reasonable exposure and accurate colors.
    }
    }
    \label{fig: res}
    \vspace{-3mm}
\end{figure*}
\begin{figure*}[t]
    \centering
    \def\fs{0.098\linewidth} 
    \setlength{\tabcolsep}{0.5pt}
    \renewcommand\arraystretch{0.8}
    \small
    \begin{tabular}{@{}cccccccccc@{}}
        \multicolumn{3}{c}{SeathruNeRF}                                             &
        \multicolumn{3}{c}{WaterSplatting}                                          &
        \multicolumn{3}{c}{Plenodium}                                           &
        \multirow{2}{*}{GT}                                                           \\
        \multicolumn{3}{c}{easy $\xrightarrow[]{\hspace{60pt}}$ hard}               &
        \multicolumn{3}{c}{easy $\xrightarrow[]{\hspace{60pt}}$ hard}               &
        \multicolumn{3}{c}{easy $\xrightarrow[]{\hspace{60pt}}$ hard}               &
        \\
        \includegraphics[width=\fs]{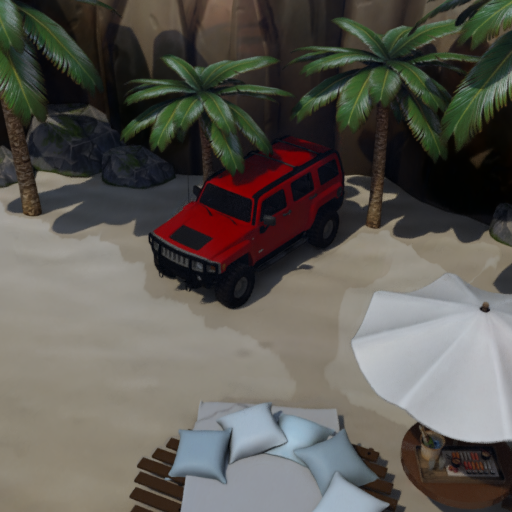}      &
        \includegraphics[width=\fs]{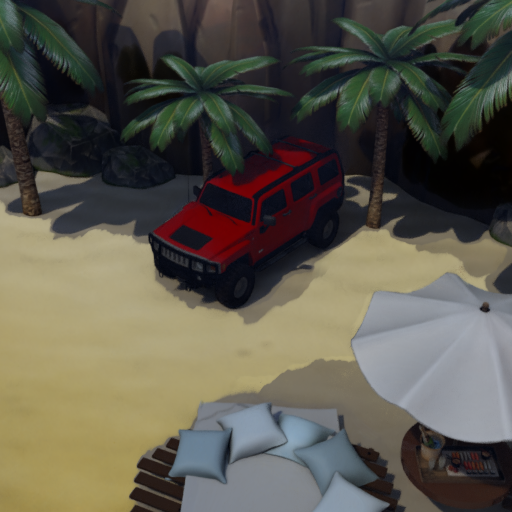}    &
        \includegraphics[width=\fs]{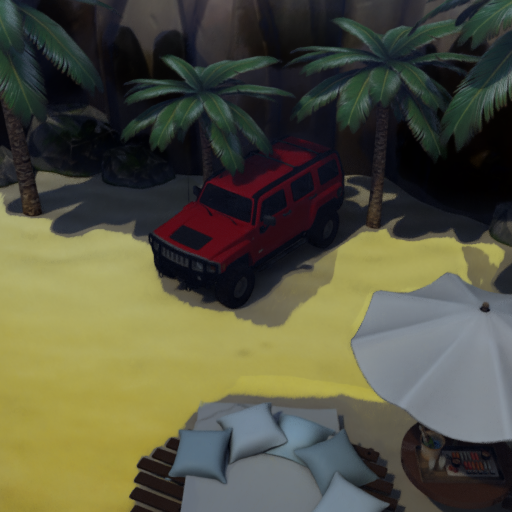}      &
        \includegraphics[width=\fs]{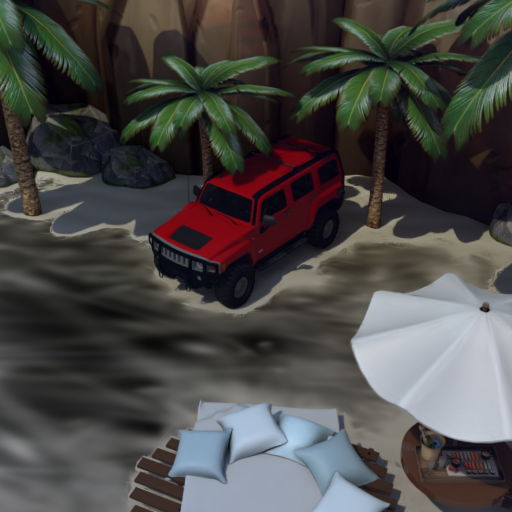}   &
        \includegraphics[width=\fs]{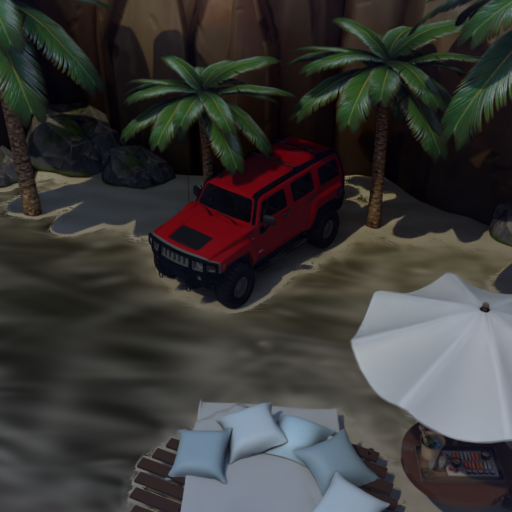} &
        \includegraphics[width=\fs]{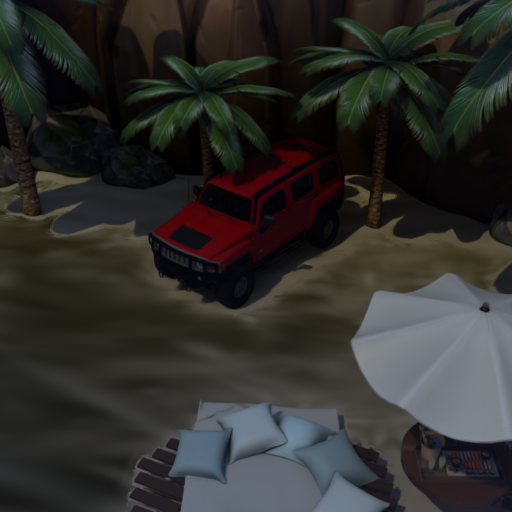}   &
        \includegraphics[width=\fs]{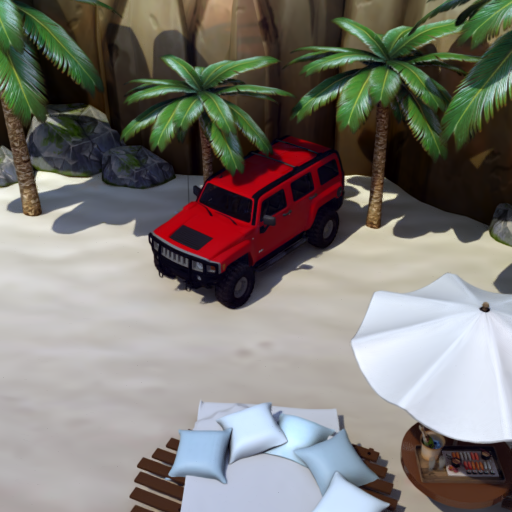}          &
        \includegraphics[width=\fs]{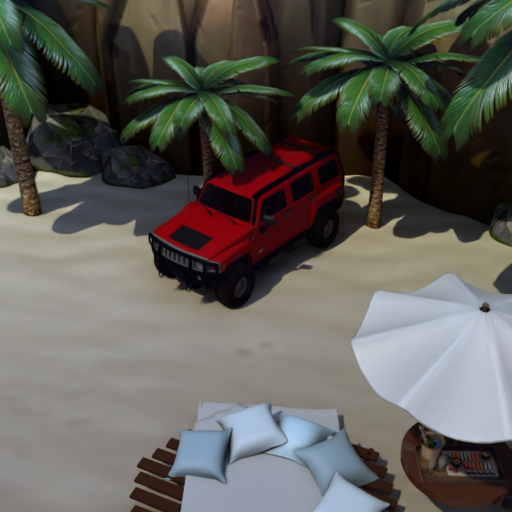}        &
        \includegraphics[width=\fs]{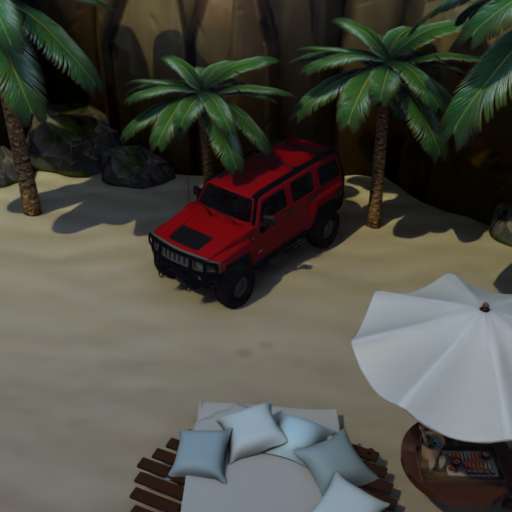}          &
        \includegraphics[width=\fs]{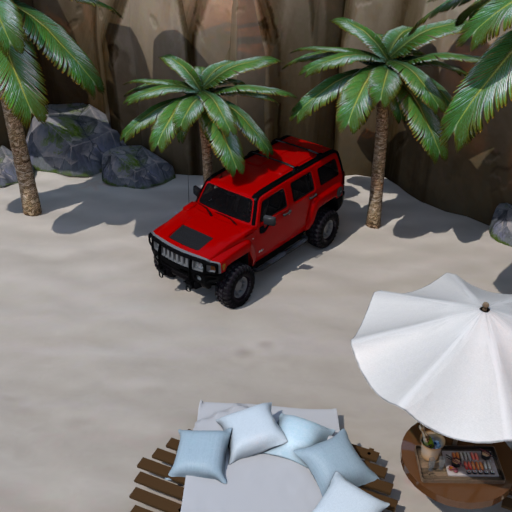}                       \\
        \scriptsize 26.41/0.908&
        \scriptsize 19.58/0.860&
        \scriptsize 16.10/0.797&
        \scriptsize 14.14/0.788&
        \scriptsize 14.45/0.788&
        \scriptsize 14.13/0.773&
        \scriptsize 27.46/0.937&
        \scriptsize 25.38/0.905&
        \scriptsize 23.67/0.878&
        \scriptsize PSNR/SSIM\\
        \includegraphics[width=\fs]{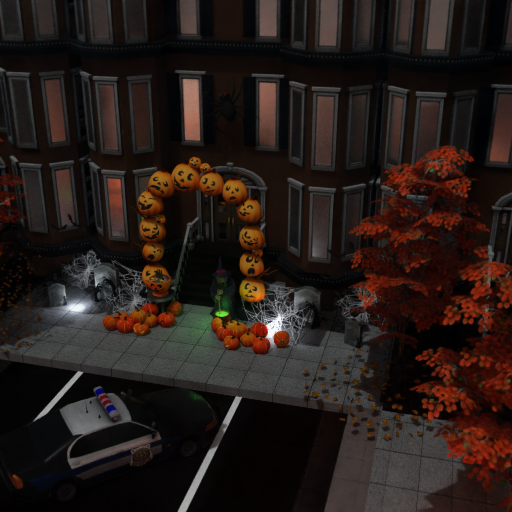}      &
        \includegraphics[width=\fs]{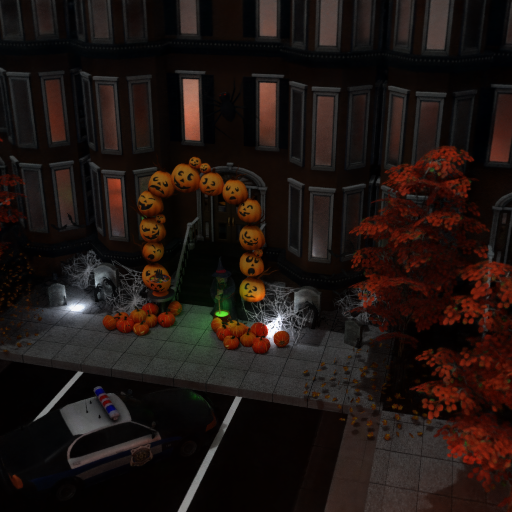}    &
        \includegraphics[width=\fs]{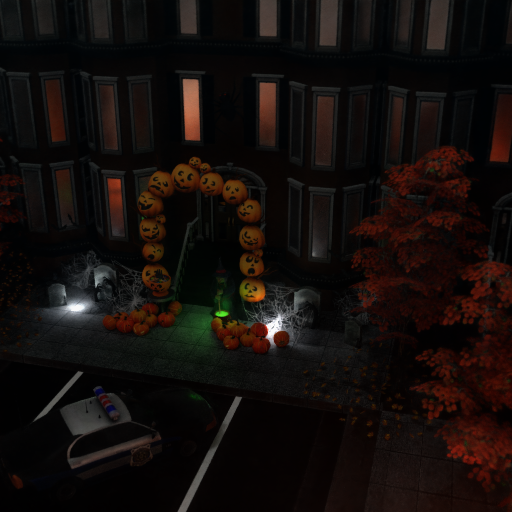}      &
        \includegraphics[width=\fs]{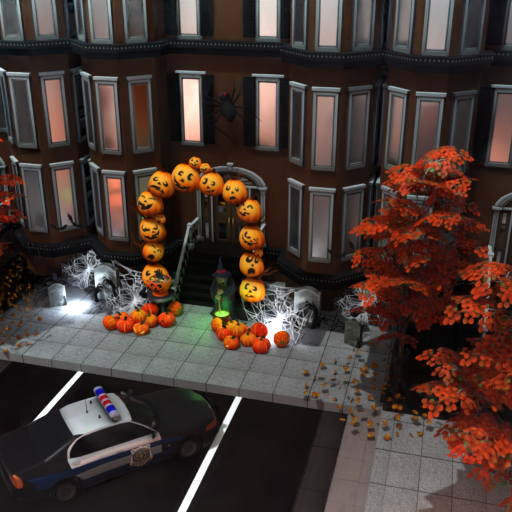}   &
        \includegraphics[width=\fs]{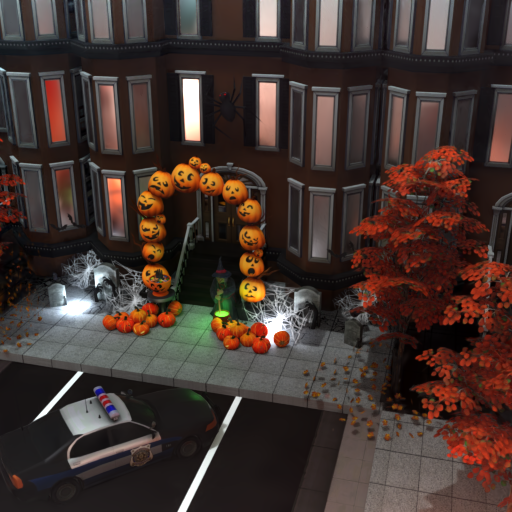} &
        \includegraphics[width=\fs]{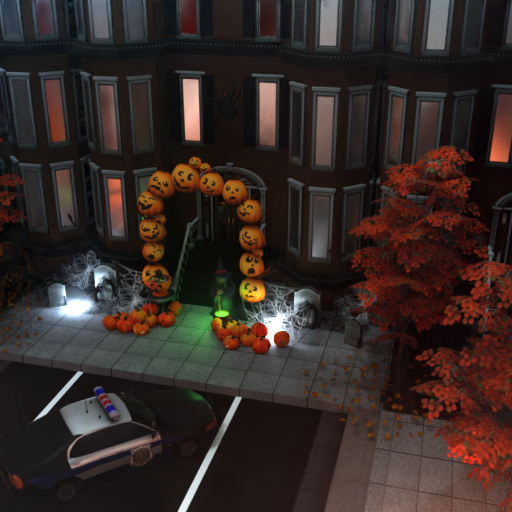}   &
        \includegraphics[width=\fs]{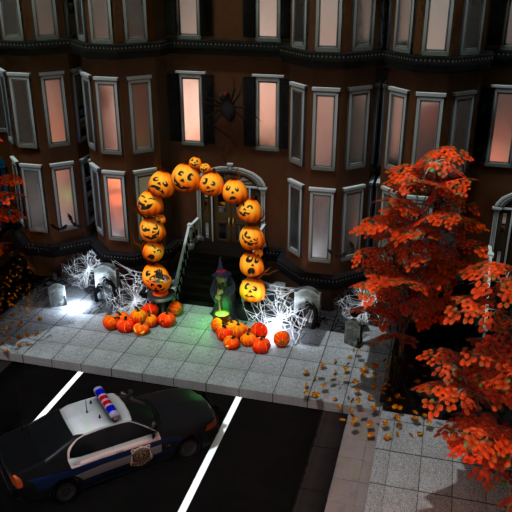}          &
        \includegraphics[width=\fs]{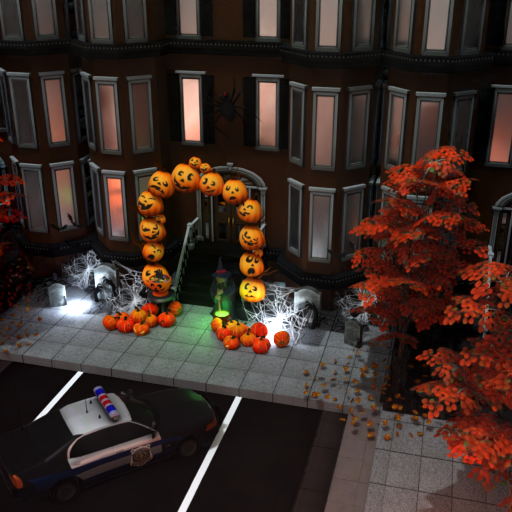}        &
        \includegraphics[width=\fs]{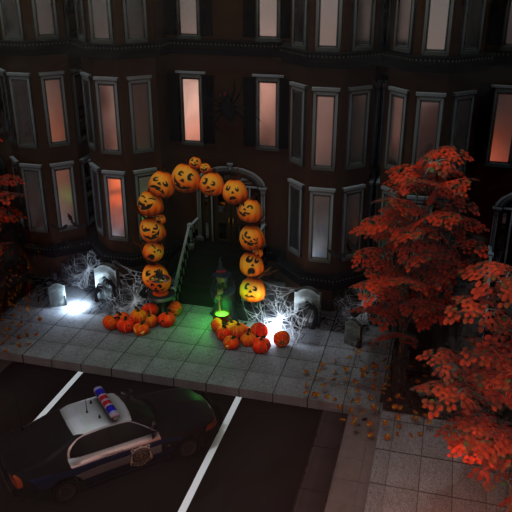}          &
        \includegraphics[width=\fs]{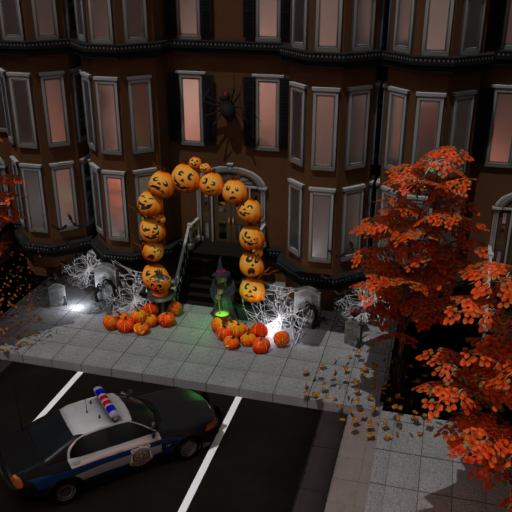}                       \\
        \scriptsize 27.18/0.881&
        \scriptsize 22.63/0.827&
        \scriptsize 16.88/0.716&
        \scriptsize 26.39/0.887&
        \scriptsize 24.63/0.841&
        \scriptsize 23.05/0.782&
        \scriptsize 28.29/0.917&
        \scriptsize 26.34/0.879&
        \scriptsize 23.38/0.784&
        \scriptsize PSNR/SSIM\\
    \end{tabular}
    \vspace{-2mm}
    \caption{{Restoration performance comparison of our Plenodium against existing methods on ``Beach'' and ``Street'' scenes from our simulated dataset. 
    Quantitative evaluations, including PSNR and SSIM metrics for each image, are beneath the corresponding figures.
    Among these methods, Plenodium produces results with more coherent textural details and superior color accuracy.
    }}
    \vspace{-5mm}
    \label{fig: res2}
\end{figure*}

\subsection{Qualitative Results}
\Figref{fig: rec} shows the visual comparisons of the rendering performance. Conventional methods, 3DGS and ZipNeRF, exhibit significant limitations when operating in scattering media, resulting in visually incoherent outputs marked by dense floaters and depth inconsistencies in underwater environments.
SeaThru-NeRF and WaterSplatting show improvements by MLP-based medium modeling, thereby improving the rendering quality with enhanced photometric consistency.
However, they still fail to accurately reconstruct fine details in water volumes and distant objects, as shown in the red boxes of \figref{fig: rec}.
In contrast, our Plenodium yields visually superior results with clear water volumes and well-defined distant objects.
As discussed in \secref{sec: alb}, our improved results stem from employing the plenoptic medium representation, the pseudo-depth Gaussian complementation, and the depth ranking regularized loss.
In addition, Plenodium produces the most accurate depth maps among compared methods, validating its effectiveness in addressing the complexity of scattering media.

%
%

\Figref{fig: res} presents qualitative comparisons the ``JapaneseGarden Red Sea'' and ``Panama'' scenes, where Plenodium achieves superior underwater restoration quality compared to state-of-the-art approaches.
Specifically, SeaThru-NeRF often produces underexposed images, leading to visually unappealing outputs in detail preservation.
WaterSplatting introduces color distortions, such as greenish or yellowish tints on object surfaces, which compromise the realism of restoration scenes.
In contrast, Plenodium generates well-exposed outputs with effective medium separation, ensuring that submerged objects are rendered with natural color fidelity.

Qualitative comparisons of the restoration performance on our simulated dataset are visualized in \figref{fig: res2}.
SeaThru-NeRF suffers from severe color distortions, most prominently manifested as unnatural yellow discoloration of sand in ``Beach''. It also frequently generates underexposed images, especially in the ``Street'' scene.
WaterSplatting exhibits hollow reconstruction artifacts in the ``Beach'' scene, due to COLMAP's failure under degraded visibility. Additionally, it and retains residual fog on windows in the “Street.” scene.
Plenodium, by comparison, achieves more accurate color restoration and preserves fine textures, demonstrating greater robustness across diverse underwater scenes.

\section{Analysis and Discussion}
\label{sec: alb}
In this section, we present ablation studies to evaluate the contributions of key components in our framework, including the plenoptic medium representation, pseudo-depth Gaussian complementation, and the loss function design. Additional details are provided in the supplementary material.

\noindent\textbf{Effect of the plenoptic medium representation.}
%
To assess the effect of position information, as well as spherical harmonics (SH) encoding, in our plenoptic medium representation, we conduct an ablation study on the WaterSplatting framework, as shown in \tabref{tab: med}. We first compare the state-of-the-art WaterSplatting~\cite{watersplatting}, which parameterizes the medium by an MLP with only the direction input (\ie, MLP w/ dir), with a modified version that replaces the MLP with SH (\ie, SH w/ dir). 
This substitution yields a 0.524 dB improvement in PSNR and an 86 FPS speedup, demonstrating SH’s superior representational power and computational efficiency. 
To further investigate the contribution of position information, We begin with a shared SH parameterization across the entire volume using only degree-zero coefficients (\ie, SH w/o dir\&pos), which lacks both directional and spatial awareness. Building upon this, we introduce trilinear interpolation over the eight voxel corners to encode position (\ie, SH w/ pos), leading to noticeable performance improvements. Finally, we combine the strengths of both directional and positional information, as well as SH encoding, in our plenoptic medium representation (\ie, SH w/ dir\&pos), which yields the highest rendering accuracy.



Furthermore, to assess the computational efficiency of our plenoptic medium representation compared to MLP-based representations, we quantified the forward and backward propagation latencies across varying pixel counts. The results, illustrated in \figref{fig: speed}, reveal that our plenoptic medium representation achieves significant speed improvements. During forward propagation, it requires only 5\% of the time needed by the MLP baseline, while during backward propagation, it is less than half as time-consuming, which highlight the computational benefits of our plenoptic medium representation.

\begin{table}
    \centering
    \scriptsize
    \def\lsize{0.48\textwidth}
    \def\rsize{0.48\textwidth}
    \begin{minipage}[t]{\lsize}
        \setlength{\tabcolsep}{1.72mm}
        \caption{Effect of the plenoptic medium presentation. All methods are evaluated on the SeaThru-NeRF~\cite{Seathru-nerf} dataset.}
        \vspace{-1.5pt}
        \vspace{-1mm}
        \begin{tabularx}{\linewidth}{l|ccc|cc}
            \toprule[1pt]
            Method                           & PSNR   & SSIM  & LPIPS & FPS & Time \\\midrule
            MLP w/ dir~\cite{watersplatting} & 29.600 & 0.918 & 0.129 & 179 & 6.9min     \\
            SH w/ dir                        & 30.124 & 0.920 & 0.129 & 265 & 6.0min     \\
            SH w/o dir\&pos                  & 29.503 & 0.917 & 0.133 & 287 & 5.6min \\
            SH w/ pos                        & 29.796 & 0.919 & 0.129 & 284 & 5.6min   \\
            \rowcolor{gray!20}
            SH w/ dir\&pos       & 30.254 & 0.921 & 0.127 & 257 & 6.0min     \\\bottomrule[1pt]
        \end{tabularx}
        \label{tab: med}
        \vspace{1mm}
    \end{minipage}
    \hfill
    \begin{minipage}[t]{\rsize}
        \renewcommand\arraystretch{1.2}
        \setlength{\tabcolsep}{2.2mm}
        \caption{{Results of our method w/o PDGC.} Superscript values mark the differences in metrics when PDGC is disabled.
        }
        \vspace{-1mm}
        \begin{tabularx}{\linewidth}{l|l|c}
            \toprule[1pt]
            Scene                    & \multicolumn{1}{c|}{init. \#G }                  & PSNR                       \\\midrule
            IUI3 Red Sea           & 21,907 $^{762\downarrow}$   & 30.176 $^{0.099\downarrow}$ \\
            Cura\c{c}ao            & 25,837 $^{453\downarrow}$   & 33.996 $^{0.124\downarrow}$ \\
            JapaneseGraden Red Sea & 21,140 $^{2,190\downarrow}$ & 24.947 $^{0.111\downarrow}$ \\
            Panama                 & 22,501 $^{90\downarrow}$    & 32.434 $^{0.001\downarrow}$ \\\bottomrule[1pt]
        \end{tabularx}
        \label{tab: pdgc}
        \vspace{1mm}
    \end{minipage}
    \begin{minipage}[t]{\rsize}
        \setlength{\tabcolsep}{1.33mm}
        \caption{{Effect of the PDGC.}}
        \vspace{-1mm}
        \begin{tabularx}{\linewidth}{l|ccc|cc}
            \toprule[1pt]
            Initialization & PSNR   & SSIM   & LPIPS  & FPS & Time   \\\midrule
            COLMAP         & 30.388 & 0.9207 & 0.1274 & 238 & 7.0min \\
            COLMAP + Rand. & 30.236 & 0.9218 & 0.1268 & 209 & 7.2min \\
            \rowcolor{gray!20}
            COLAMP + PDGC    & 30.472 & 0.9225 & 0.1276 & 249 & 7.0min \\\bottomrule[1pt]
        \end{tabularx}
        \label{tab: init}
    \end{minipage}
    \hfill
    \begin{minipage}[t]{\lsize}
        \caption{{Effect of the loss functions.}}
        \vspace{-1mm}
        \setlength{\tabcolsep}{1.3mm}
        \begin{tabularx}{\linewidth}{l|cc|cc}
            \toprule[1pt]
            Loss Configuration                                                                                 & PSNR   & SSIM   & FPS & Time   \\\midrule
            $\mathcal{L}_{\text{reg-}\mathcal{L}_1}\&\mathcal{L}_\text{reg-ssim}$                              & 30.307 & 0.9216 & 238 & 6.0min \\
            $\mathcal{L}_{\text{reg-}\mathcal{L}_1}\&\mathcal{L}_\text{reg-ssim}\&\mathcal{L}_\text{depth}$    & 30.389 & 0.9204 & 253 & 6.1min \\
            \rowcolor{gray!20}
            $\mathcal{L}_{\text{reg-}\mathcal{L}_1}\&\mathcal{L}_\text{reg-ms-ssim}\&\mathcal{L}_\text{depth}$ & 30.472 & 0.9225 & 249 & 7.0min \\
            \bottomrule[1pt]
        \end{tabularx}
        \label{tab: loss}
    \end{minipage}
    \vspace{-2mm}
\end{table}
\begin{figure}
    \centering
    \scriptsize
    \def\lsize{0.48\textwidth}
    \def\rsize{0.48\textwidth}
    \setlength{\tabcolsep}{0.5pt}
    \begin{minipage}{\lsize}
        \includegraphics[width=\linewidth]{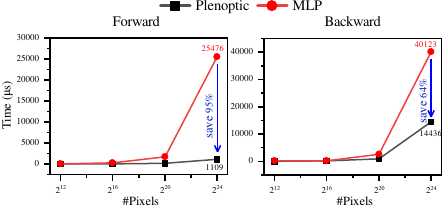}
        \vspace{-8pt}
        \caption{{Efficiency comparison of our plenoptic medium representation against MLP-based representation in forward and backward.}}
        \label{fig: speed}
    \end{minipage}
    \hfill
    \begin{minipage}{\rsize}
        \scriptsize
        \def\fs{0.32\linewidth} 
        \begin{tabular}{@{}cccc@{}}
                                                           &  step = 1000 &  step = 2000 &  step = 15000 \\
            \rotatebox{90}{\hspace{0pt} w/o PDGC}          &
            \includegraphics[width=\fs]{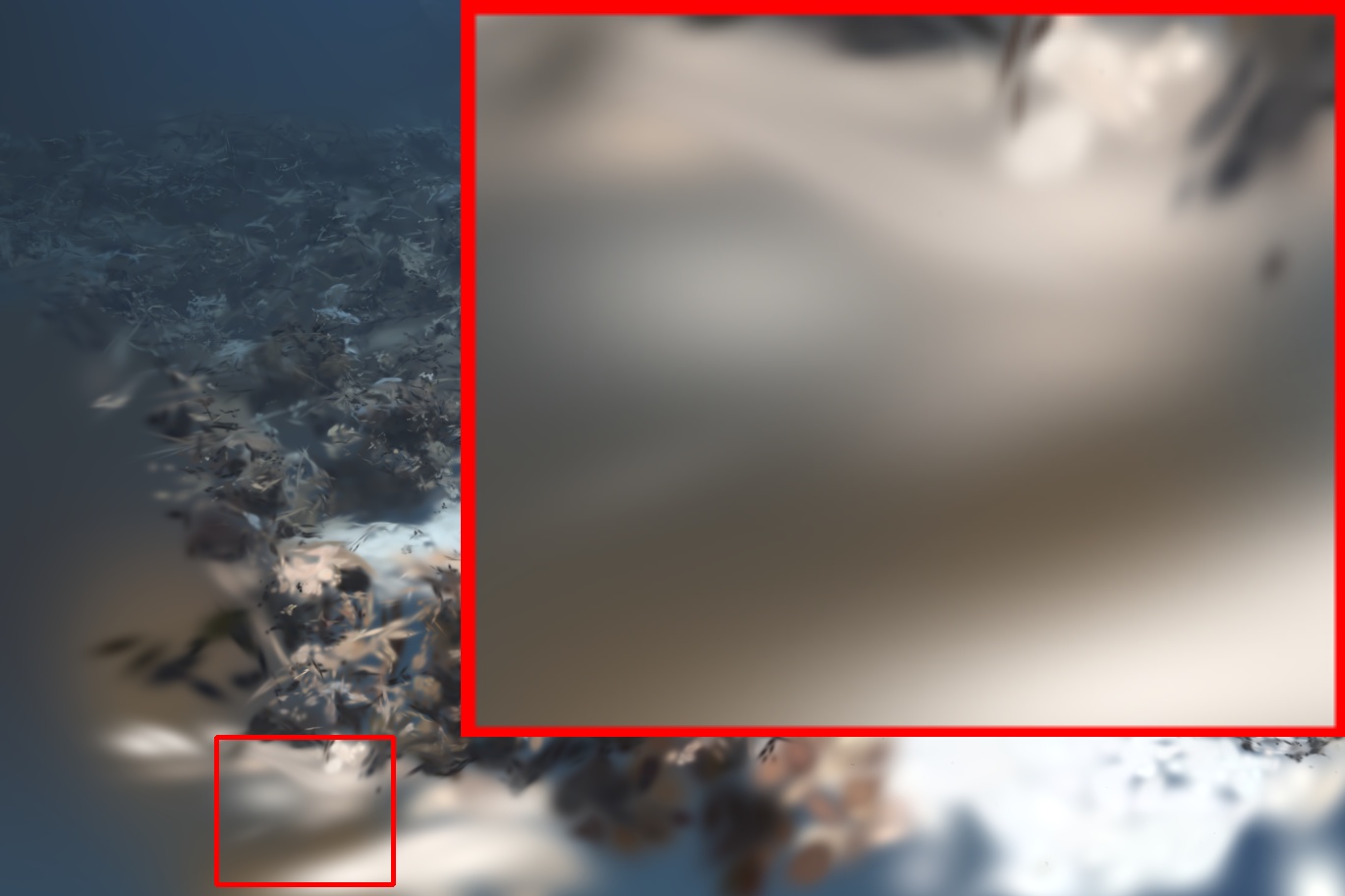} &
            \includegraphics[width=\fs]{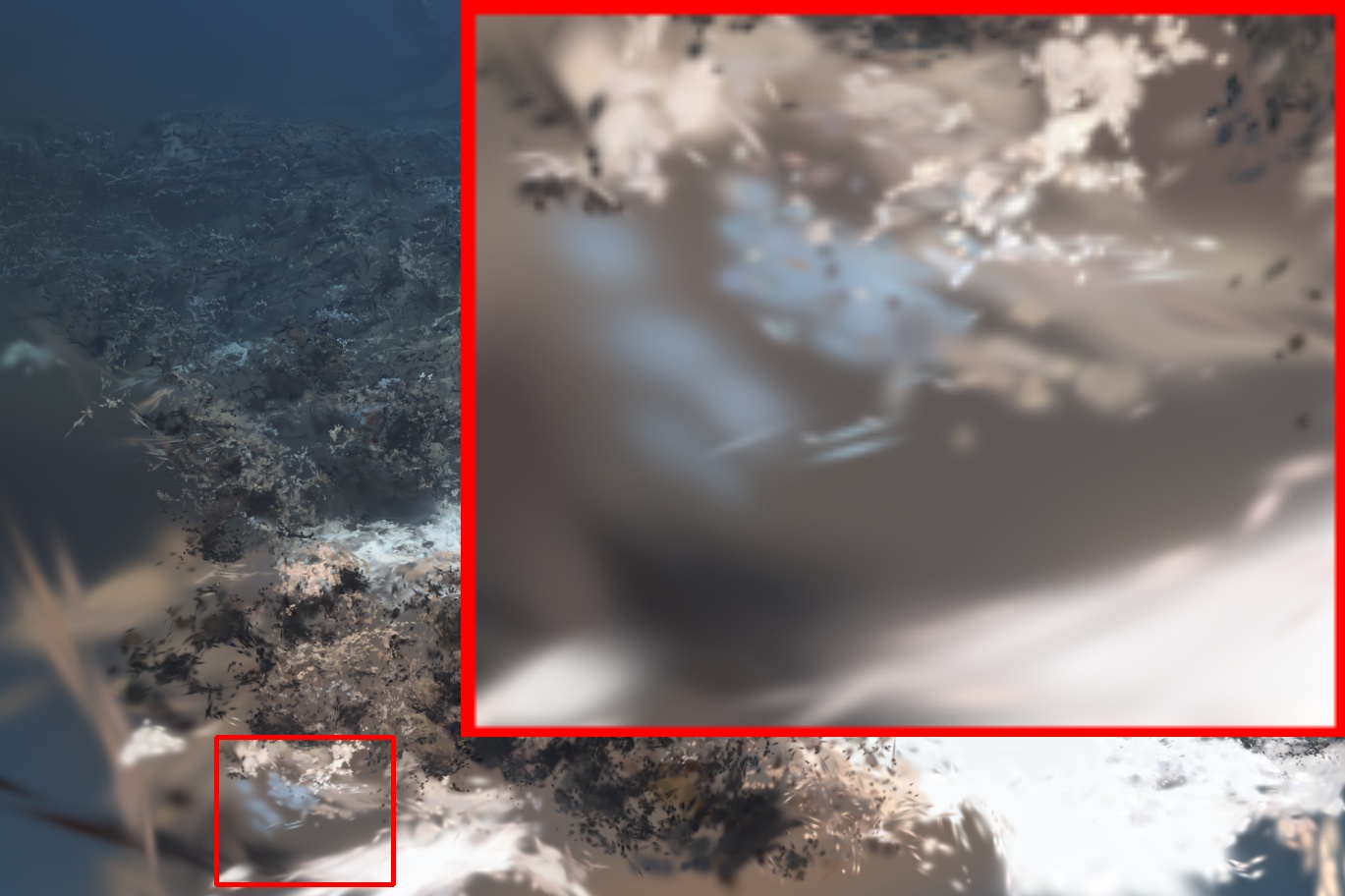} &
            \includegraphics[width=\fs]{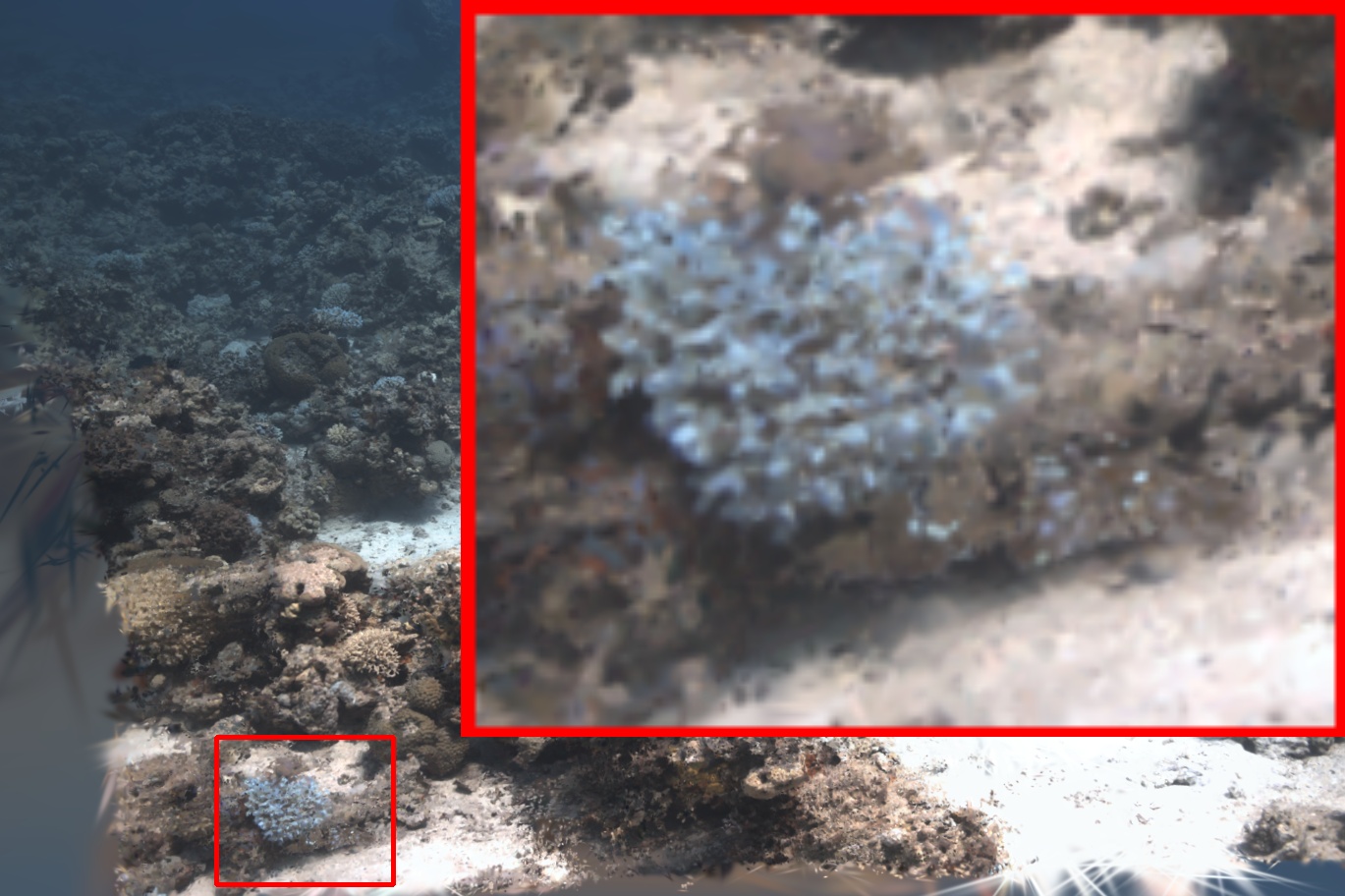}                                                                \\
            \rotatebox{90}{\hspace{0pt} w/ PDGC}           &
            \includegraphics[width=\fs]{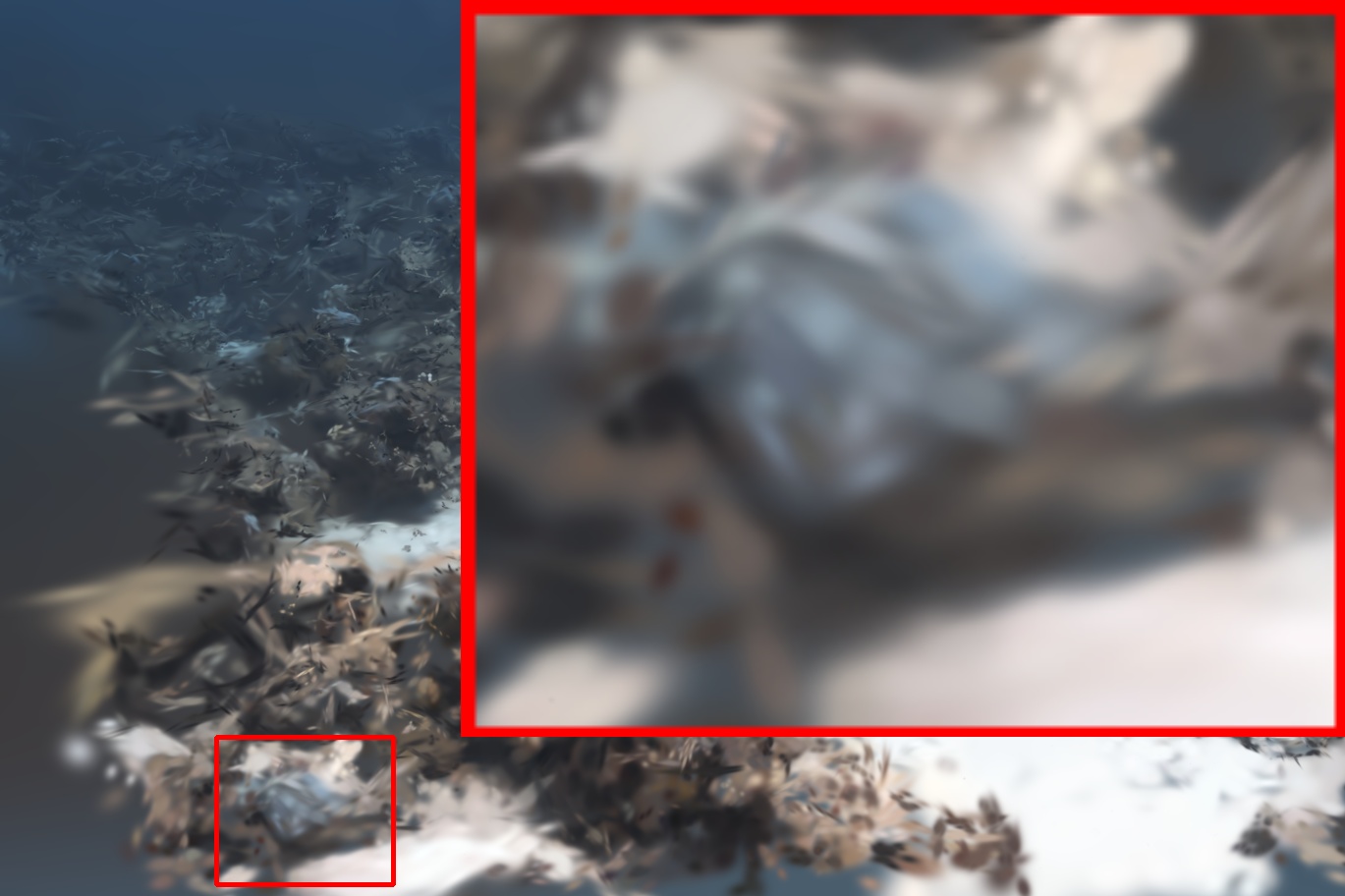}  &
            \includegraphics[width=\fs]{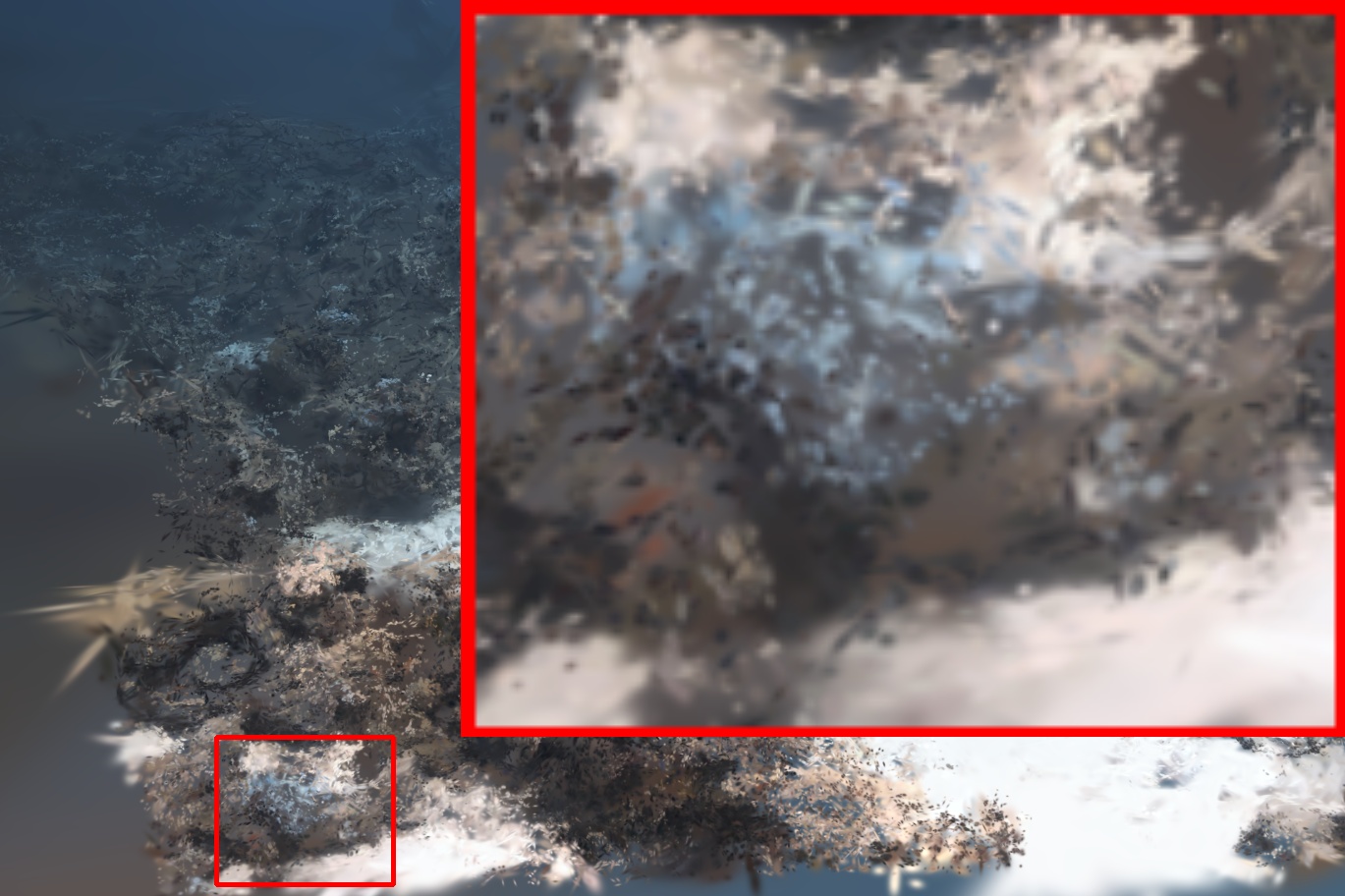}  &
            \includegraphics[width=\fs]{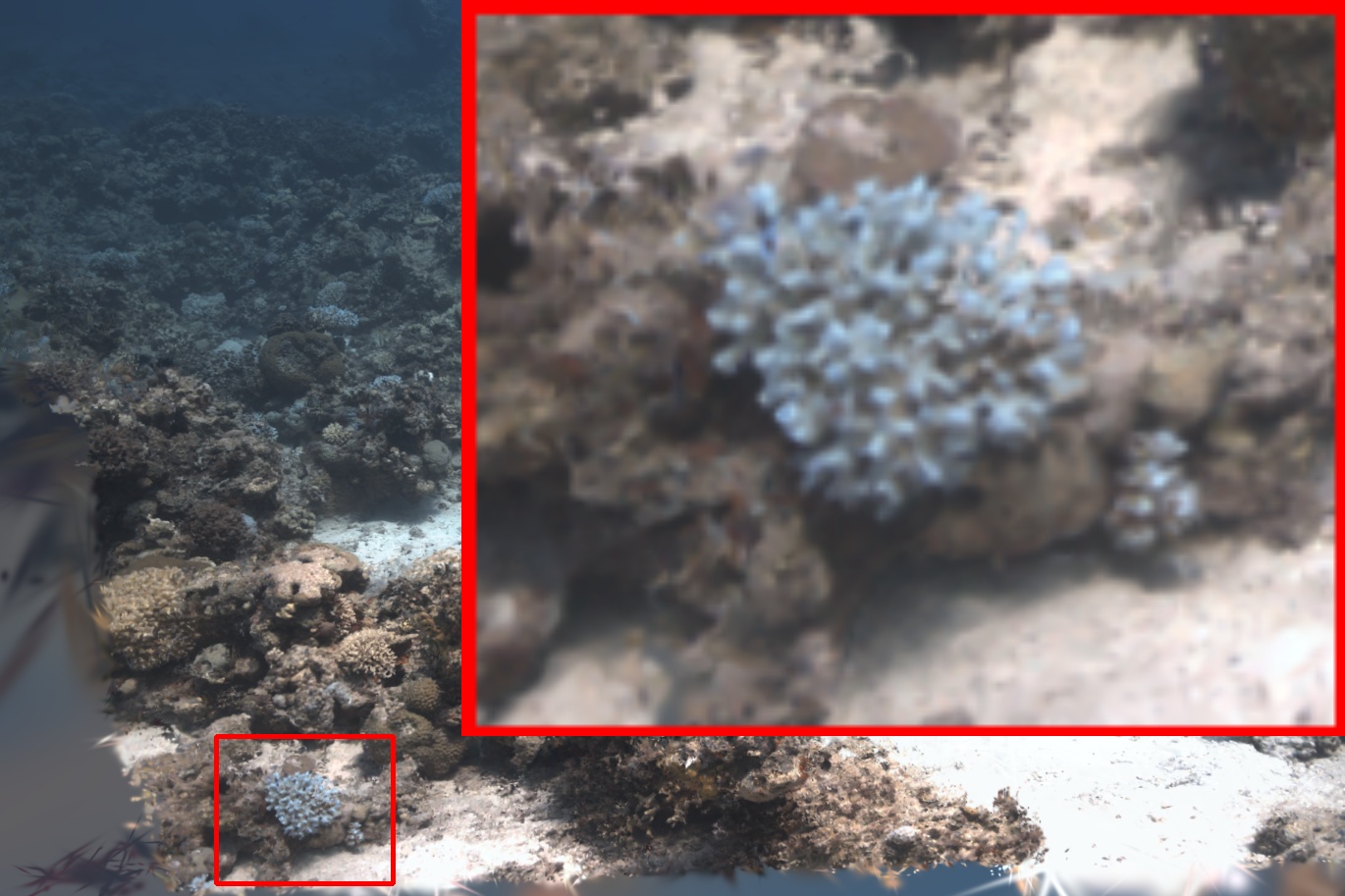}                                                                 \\
        \end{tabular}
        \caption{{Rendering image comparison between w/ PDGC and w/o PDGC when different steps (1000, 2000, 15000).}}
        \label{fig: step}
    \end{minipage}
    \vspace{-5mm}
\end{figure}

\noindent\textbf{Effect of the pseudo-depth Gaussian complementation.}
To demonstrate the effectiveness of our pseudo-depth Gaussian complementation (PDGC) method, we remove PDGC and train this baseline using the same settings as ours.
The quantitative results in~\tabref{tab: pdgc} show that our PDGC supplements 762, 453, 2190, and 90 Gaussian primitives for each test scene, improving the PSNR value by 0.099dB, 0.124dB, 0.111dB, and 0.001dB, respectively.
To further analyze the effect of the proposed PDGC, we compare with a baseline method that supplements Gaussians by randomly selecting positions rather than by our PDGC.
The comparison results in \tabref{tab: init} demonstrate the effectiveness of the proposed PDGC.
Furthermore, \figref{fig: step} visually validates that the PDGC effectively improves the reliability of the initialization for 3DGS in degraded scenarios, producing high-quality reconstructions with significantly enhanced clarity compared to the baseline without using PDGC.


\noindent\textbf{Effect of the loss function.}
To show the effect of our improved components in the loss function \eqnref{eq: loss}, we compare with two baseline methods that respectively replace our loss with $\lambda_{\mathcal{L}_1}\mathcal{L}_{\text{reg-}\mathcal{L}_1}+\lambda_{\text{ssim}}\mathcal{L}_\text{reg-ssim}$ (which contains a single-scale structural similarity loss and is used in prior work~\cite{watersplatting}) or $\lambda_{\mathcal{L}_1}\mathcal{L}_{\text{reg-}\mathcal{L}_1}+\lambda_{\text{ssim}}\mathcal{L}_\text{reg-ssim}+\lambda_{\text{depth}}\mathcal{L}_\text{depth}$.
\Tabref{tab: loss} reveals that both the improved multi-scale SSIM loss and the proposed depth ranking regularized loss yield significant improvements in terms of PSNR and SSIM while imposing minimal additional computational overhead during training.

\section{Conclusion}
\label{sec: conclusion}
In this paper, we propose \emph{Plenodium}, an efficient and robust framework for underwater 3D reconstruction.
Our innovative plenoptic medium representation effectively integrates positional information to enhance medium modeling accuracy. Utilizing spherical harmonics-based encoding, we achieve a 47\% speedup relative to WaterSplatting.
Using pseudo-depth Gaussian complementation, we significantly improve the robustness of the initialization process.
The proposed depth ranking regularized loss further improves the geometry by using depth order.
Extensive evaluations and comparisons in both real-world and simulated datasets with state-of-the-art methods demonstrate the effectiveness of our approach in reconstructing 3D underwater scenes and restoring underwater images.



{
\bibliographystyle{IEEEtran}
\bibliography{ref.bib}
}


\appendix

\end{document}